%% file: main.tex
\pdfoutput=1
\documentclass{article}

\usepackage{microtype}
\usepackage{graphicx}
\usepackage{subcaption}
\usepackage{booktabs}
\usepackage{multirow}
\usepackage{xcolor}
\usepackage{hyperref}

\usepackage[preprint]{icml2026}
\makeatletter\renewcommand{\ICML@preprint}{\textit{Preprint.}}\makeatother  
\hypersetup{pdfsubject={}}  

\usepackage{amsmath}
\usepackage{amssymb}
\usepackage{mathtools}
\usepackage{amsthm}
\usepackage[capitalize,noabbrev]{cleveref}
\usepackage{algorithm}
\usepackage{algorithmic}

\theoremstyle{plain}
\newtheorem{proposition}{Proposition}
\newtheorem{lemma}{Lemma}
\newtheorem{corollary}{Corollary}
\theoremstyle{definition}

\usepackage[textsize=tiny]{todonotes}
\usepackage{tikz}
\usepackage{enumitem}
\usepackage{placeins}
\usetikzlibrary{positioning,calc,arrows.meta}

\input{macros}

\icmltitlerunning{Multi-Future Drafting: Mixture Block Drafters for Tree-Verified Speculative Decoding}

\begin{document}

\twocolumn[
  \icmltitle{Multi-Future Drafting: Mixture Block Drafters\texorpdfstring{\\}{ }for Tree-Verified Speculative Decoding}

  \icmlsetsymbol{equal}{*}

  \begin{icmlauthorlist}
    \icmlauthor{Haoyu Huang}{ncee,canva}
    \icmlauthor{Jialiang Kang}{canva}
    \icmlauthor{Linlin Yang}{cuc}
    \icmlauthor{Xiao Yang}{canva}
    \icmlauthor{Baochang Zhang}{sai}
  \end{icmlauthorlist}

  \icmlaffiliation{ncee}{National College for Excellent Engineers, Beihang University}
  \icmlaffiliation{canva}{Canva Research}
  \icmlaffiliation{cuc}{Communication University of China}
  \icmlaffiliation{sai}{School of Artificial Intelligence, Beihang University}

  \icmlcorrespondingauthor{Xiao Yang}{xiaoy@canva.com}
  \icmlcorrespondingauthor{Baochang Zhang}{bczhang@buaa.edu.cn}

  \icmlkeywords{Speculative Decoding, Block Diffusion Drafters, Draft Trees, Mixture Models}

  \vskip 0.3in
]

\printAffiliationsAndNotice{Project page: \url{https://hhhhhhy5.github.io/multi-future-drafting/}. }

\input{sections/abstract}
\input{sections/intro}
\input{sections/related}
\input{sections/theory}
\input{sections/method}
\input{sections/experiments}
\input{sections/analysis}
\input{sections/conclusion}
\input{sections/impact}

\bibliography{refs}
\bibliographystyle{icml2026}

\newpage
\appendix
\onecolumn
\noindent\textbf{Overview.} \Cref{app:proofs} proves the results of \cref{sec:theory}. \Cref{app:rules} shows that the
merged heap is exact and compares tree rules, priors and assignment rules; \cref{app:hybrid} derives exact tree
verification for Gated-DeltaNet targets; \cref{app:pertask} breaks \cref{tab:main} down by task, and the hybrid
targets by budget. \Cref{app:modes} analyses what the modes learn and the alternatives of \cref{sec:alternatives};
\cref{app:kspeed} times the number of modes; \cref{app:chimeras} counts chimeras and locates the gain.
\Cref{app:details} gives training and evaluation details, including training longer (\cref{sec:epochs}), the timing
runs and DARTree at matched budgets; \cref{app:batch} studies larger batch sizes and ends with the limitations; and
\cref{sec:combine} puts modes on a drafter that already conditions. We will release our code for training and inference, including the SGLang engine.

\input{sections/proofs}
\input{sections/appendix}

\end{document}

%% file: macros.tex
\newcommand{\mfd}{MFD}
\newcommand{\ddtree}{DDTree}
\newcommand{\dflash}{DFlash}
\newcommand{\dpace}{D-PACE}
\newcommand{\K}{K}                         
\newcommand{\B}{B}                         
\newcommand{\Lb}{L}                        
\newcommand{\pimix}{\pi}                   
\newcommand{\qk}[1]{q_{#1}}                
\newcommand{\memb}{\mathbf{e}}             
\newcommand{\E}{\mathbb{E}}
\newcommand{\argmax}{\operatorname*{arg\,max}}

\newcommand{\TC}{\mathrm{TC}}              

%% file: sections/abstract.tex
\begin{abstract}
Block drafters such as \dflash{} draft a block of future tokens in one forward pass, and draft trees such as
\ddtree{} verify the prefixes ranked highest by the product of the drafter's per-position distributions. Under
tree verification, the expected number of accepted tokens is the probability mass the tree covers. We show that a
product misallocates it: when the continuation can follow several futures, it splices them into \emph{chimeras},
prefixes stitched together from different futures that the target never produces, and gives them as much budget as the
futures themselves. Barring ties, no product-ranked tree can hold two futures that disagree in $m$ positions without $2^{m-1}{-}1$
chimeras, however well the drafter is trained, and in practice chimeras make up 27--33\% of a 64-node tree. We
propose \emph{Multi-Future Drafting} (\mfd{}), which restores the coupling with a latent variable instead of the chain
rule: $\K$ learned mode embeddings at the drafter's input draft $\K$ coherent blocks in the same pass, and a
mixture \ddtree{} lets the verifier pick the future the target takes, losslessly and with $\K{+}1$ extra
vectors. On four Qwen targets from 4B to 27B parameters, at batch size 1, \mfd{} reaches up to 7.8$\times$ speedup, 5--8\% faster than the best single-mode tree trained identically and, on Qwen3-4B and Qwen3-8B, 9--16\% faster than the released DARTree trained on the same prompts.
\end{abstract}

%% file: sections/intro.tex
\section{Introduction}
\label{sec:intro}

Speculative decoding accelerates autoregressive generation by letting a cheap drafter propose tokens that the target
model verifies in parallel, without changing the output distribution
\citep{leviathan2023fast,chen2023accelerating}. The fastest drafters today draft a whole block in one forward pass:
\dflash{} \citep{chen2026dflash} proposes the next 15 tokens at once, conditioned on the target's hidden states, and
\ddtree{} \citep{ringel2026ddtree} turns its per-position distributions into a tree of $\B$ candidate prefixes that the
target verifies in a single pass.

\begin{figure*}[t]
  \centering
  \includegraphics[width=\textwidth]{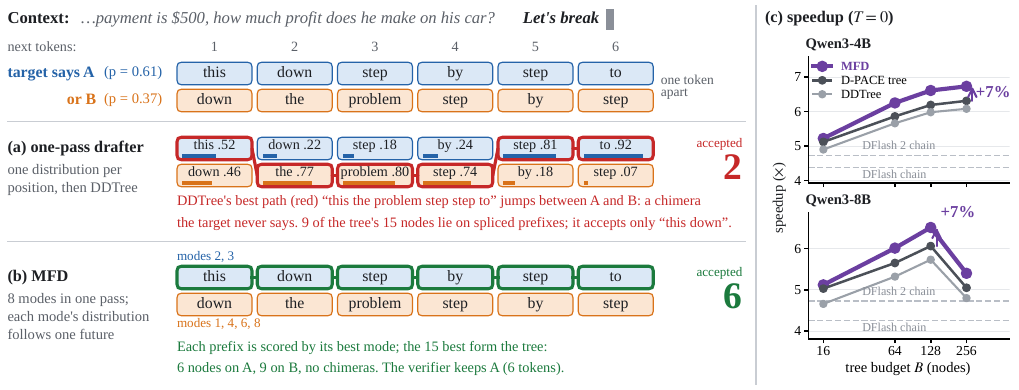}
  \caption{\textbf{Why a one-pass drafter needs several futures.} (a, b) A real GSM8K context for Qwen3-4B: after
    \emph{Let's break}, the target says future A ($p{=}0.61$) or future B ($p{=}0.37$). (a) A one-pass drafter
    (\dflash{} fine-tuned with \dpace{}) gives one distribution per position, each mixing A's token and B's; \ddtree{}'s best path jumps between
    the futures, a chimera. 9 of the tree's 15 nodes lie on prefixes that splice A and B, and it
    accepts 2 tokens. (b) \mfd{}'s 8 modes run in the same pass; six follow A or B, two neither. Every prefix is scored by its best mode and the 15 best form the tree: 6 nodes on A, 9 on B,
    none a chimera; the verifier keeps A, 6 tokens. (c) Speedup ($T{=}0$, 5-task mean) against the tree budget $\B$:
    \mfd{} leads the strongest \dpace{} tree under the same recipe at every budget, by about 2\% at $\B{=}16$ and
    6--7\% from $\B{=}64$ on. \ddtree{} uses the released drafter; dashed: chain drafters.}
  \label{fig:teaser}
\end{figure*}

\textbf{What should a draft tree contain?} Under tree verification, the expected number of accepted tokens is the
probability mass of the target's continuations that the tree covers, so the best tree with $\B$ nodes holds the $\B$
most probable prefixes (\cref{prop:coverage}), ranked by \emph{joint} probability. A one-pass drafter like
\dflash{} provides one distribution per position, and \ddtree{} ranks prefixes by the product of these distributions,
which for the target is guaranteed to be optimal only if each is the true marginal and the positions are independent
(\cref{eq:ptc}). The positions are not independent: a reasoning step can be phrased, ordered or formatted in several ways, and the futures
that result disagree at most positions after they split.

\textbf{A product cannot keep futures apart.} In \cref{fig:teaser}a, after \emph{Let's break} the target continues
with \emph{this down step by step} or with \emph{down the problem step by step}. The drafter's distribution at each
position splits between the two, and \ddtree{}'s best path takes the likeliest token everywhere: \emph{this the problem
step step to}. We call such a prefix a \emph{chimera}: like the creature of myth assembled from parts of different
animals, it is stitched together from tokens of different futures, and the target never says it. This is the common
case: over 628 prompts on Qwen3-4B and Qwen3-8B, 27--33\% of the nodes of a 64-node product tree are chimeras, and its
best path passes through one in 61--65\% of rounds (\cref{app:chimeras}). The failure is structural: the error of a
product is the total correlation between positions, which compounds with depth, so even a perfectly trained drafter of
this kind is capped by the coupling it cannot represent.

\textbf{Conditioning on drafted tokens.} The classical fix is the chain rule: condition each drafted position on
the tokens before it. Autoregressive drafters work this way, and concurrent works graft it onto block drafters
\citep{oda2026weaver,li2026dartree}. The drafter then learns which future a branch follows only from
the branch's own tokens, and the coupling is paid for with part of the sequential drafting that one-pass drafters
remove.

To address these limitations, we propose \emph{Multi-Future Drafting} (\mfd{}), which restores the coupling with the
other classical tool: a latent variable that names the future; given it, the positions are independent. \mfd{} adds
$\K$ learned mode embeddings to the input of a block drafter and runs the $\K$ modes side by side in the same forward
pass (\cref{fig:teaser}b). Each mode's distribution follows one future, so the tree no longer has to splice futures to
hold several of them; at the same budget, it covers 8--10\% more of the target's probability mass than the strongest
single-mode tree on Qwen3-4B (\cref{sec:budget}). The modes are not separate drafts: one tree keeps the $\B$ best prefixes over
all modes, each scored by its best mode, turning uncertainty about the future into breadth that the verifier checks in
parallel. Mixtures have drafted in parallel before, at the drafter's output with
a single verified branch \citep{dblast2026} or from random inputs \citep{ptp2025,flashmap2026}; \mfd{} learns its modes at the input and ranks all of them in one best-first tree. Our contributions are:
\begin{itemize}[leftmargin=1.2em, topsep=3pt, itemsep=2pt, parsep=0pt, partopsep=0pt]
  \item We show that a draft tree ranked by a product of per-position distributions cannot keep futures apart: to
  hold two futures that disagree in $m$ positions, it must also hold, barring ties, $2^{m-1}{-}1$ of their chimeras,
  and in practice chimeras make up 27--33\% of a 64-node tree.
  \item We propose \mfd{}, which adds $\K$ mode embeddings to the input of a block drafter, trains them
  winner-take-all, and keeps the $\B$ best prefixes over all modes in one exact tree, at a cost of $\K{+}1$ vectors of
  the drafter's width and no extra drafting pass.
  \item On four Qwen targets from 4B to 27B, two of them hybrid, \mfd{} is the fastest method on every task at both
  temperatures: 5--8\% faster than the strongest single-mode tree trained identically, 9--16\% faster than the released DARTree trained on the same prompts (Qwen3-4B and Qwen3-8B), and 45--46\% faster than the fastest released drafter in SGLang
  at batch size 1.
\end{itemize}

%% file: sections/related.tex
\section{Related Work}
\label{sec:related}

\textbf{Speculative decoding and draft trees.} Speculative sampling is lossless for any drafter
\citep{leviathan2023fast,chen2023accelerating}. Tree verification checks many candidates in one target pass
\citep{miao2024specinfer,sun2023spectr}, with tree shapes learned or optimized for a budget
\citep{cai2024medusa,li2024eagle2,li2025eagle3,chen2024sequoia,wang2025opttree}; \ddtree{}
\citep{ringel2026ddtree} builds best-first trees from a block drafter's per-position distributions.

\textbf{Block drafters.} \dflash{} \citep{chen2026dflash} drafts a block in one pass, conditioned on target features.
Follow-ups change its training objective \citep[\dpace{};][]{dpace2026}, make attention within the block causal
\citep[JetSpec;][]{hu2026jetspec}, add a selector over each position's top candidates
\citep[\dflash{}~2;][]{inco2026dflash2} or a recurrent correction of the logits \citep[Domino;][]{domino2026}, or draft
semi-autoregressively \citep{dspark2026}. Multi-token prediction
heads \citep{stern2018blockwise,gloeckle2024mtp,cai2024medusa} also give one distribution per position.

\textbf{Restoring dependencies by conditioning.} Concurrent works bring the chain rule of autoregressive drafters \citep{li2024eagle,ankner2024hydra} to block drafters \citep{rheinboldt2026treeflash}: Weaver's autoregressive adapter \citep{oda2026weaver}, DARTree's token-conditioned branch corrections
\citep{li2026dartree}, UBTree's bigram selector \citep{liang2026ubtree} and TreeSpark's parent-conditioned head
\citep{zhao2026treespark}. Weaver also estimates an acceptance ceiling for factorized chains; we
characterize the gap for trees and close it without sequential drafting.

\textbf{Multi-modality in parallel generation.} Per-position predictions also mix valid outputs into incoherent ones
in non-autoregressive translation \citep{gu2018nat,zhou2020understanding,huang2022dat} and masked diffusion models
\citep{benhamu2025entropy,liu2024copula,wu2025fastdllm}. Latent variables model this dependence in
translation \citep{kaiser2018fast,ma2019flowseq,shu2020latent}, and a teacher can rescore a few latent-conditioned
candidates \citep{gu2018nat}, but one output is committed, so modes are removed or chosen; under tree verification the
verifier checks all modes' prefixes at once, so the drafter should keep them.

\textbf{Mixtures and multiple hypotheses.} Mixtures of factorized distributions have been used for multi-token
prediction \citep{basharin2024faster,mtpc2025} and for block drafters with an output-side mixture head and a single
branch \citep{dblast2026}; latent inputs make parallel drafts dependent \citep{ptp2025}, drafts sampled with
independent noise can be merged into a trie \citep{flashmap2026}, as can those of several boost-tuned drafters
\citep{miao2024specinfer}. In translation, mixtures with a learned latent input,
trained by EM on the marginal likelihood \citep{he2018sequence} or by hard EM \citep{shen2019mixture}, generate diverse
outputs, and winner-take-all training is multiple-choice learning
\citep{guzmanrivera2012mcl,lee2016smcl,rupprecht2017uncertain}. Unlike output-side mixtures, \mfd{} injects the latent
at the drafter's input, which we show matters; unlike drafts from input noise or separate drafters, its modes share
one pass, are trained on acceptance and are searched by one exact mixture tree; and we tie the value of
a mixture to the verifier's ability to branch.

%% file: sections/theory.tex
\section{Why One-Pass Drafters Need Multiple Futures}
\label{sec:theory}

\subsection{Tree verification covers prefixes}

A block drafter such as \dflash{} \citep{chen2026dflash} drafts the $D=\Lb-1$ positions after the last verified token
($\Lb{=}16$) in one forward pass, each with its own distribution $q_i$ that does not see the tokens drafted before it,
and \ddtree{} \citep{ringel2026ddtree} keeps the $\B$ prefixes $v=(v_1,\dots,v_d)$ with the highest
$\sum_{i\le d}\log q_i(v_i)$, which form a tree that the target verifies in one forward pass. Proofs are in
\cref{app:proofs}.

Let $Y=(Y_1,\dots,Y_D)$ be the tokens the target will produce next (its greedy tokens at $T{=}0$, its samples at
$T{>}0$), and let $p$ be their law given the information $\mathcal{G}$ that the drafter's tree depends on: we draw
contexts from the data and set $p(v)=P(Y_{1:d}=v\mid\mathcal{G})$ for a prefix $v$ of depth $d$, so
\cref{prop:coverage} holds for every tree determined by $\mathcal{G}$. At $T{=}0$, $p$ would be a point mass if
$\mathcal{G}$ were the whole context; a small drafter reading a few target layers keeps coarser information, and $p$ is
the uncertainty that remains for it, typically split over the ways a step can be phrased, ordered or formatted.
Verification walks down the tree $\mathcal{T}$ following the target's own tokens and accepts
$\alpha_{\mathcal{T}}(Y)=\max\{d:Y_{1:d}\in\mathcal{T}\}$ drafted tokens plus one bonus token, which leaves the
target's output distribution unchanged \citep{miao2024specinfer,ringel2026ddtree}.

\begin{proposition}[Coverage]
  \label{prop:coverage}
  For every draft tree, $\E[\alpha_{\mathcal{T}}(Y)]=\sum_{v\in\mathcal{T}}p(v)$. Among trees with at most $\B$
  nodes, the expectation is maximized by the $\B$ prefixes with the largest $p(v)$, which form a tree when ties are
  broken toward shorter prefixes.
\end{proposition}

The identity holds for verification that follows the target's own tokens, unlike multi-draft sampling
\citep{sun2023spectr,khisti2025multidraft}; tree-search methods maximize the same sum under their own acceptance models
\citep{chen2024sequoia,wang2025opttree}, and \ddtree{} states it for the drafter's factorized distribution
\citep{ringel2026ddtree}. Stated for $p$, it says what drafting is
for: under a node budget, the drafter must rank prefixes by their \emph{joint} probability.

\subsection{A product cannot keep futures apart}

A tree built from one distribution per position ranks a prefix by a product $Q(v)=\prod_{i\le d}q_i(v_i)$. The
cross-entropy-optimal $q_i$ is the marginal $p_i=P(Y_i=\cdot\mid\mathcal{G})$, so the best such drafter ranks by
$\tilde p(v)=\prod_{i\le d}p_i(v_i)$, whose error is the dependence between positions: on average over the target's
continuations,
\begin{equation}
  \E\big[\log p(Y_{1:d})-\log\tilde p(Y_{1:d})\big]=\TC(Y_{1:d}),
  \label{eq:ptc}
\end{equation}
the total correlation $\TC(Y_{1:d})=\sum_{i\le d}H(Y_i)-H(Y_{1:d})$ \citep{watanabe1960information}, which grows by
$I(Y_{d+1};Y_{1:d})\ge0$ with every position. On average the product deflates coherent futures by a factor that
compounds with depth, while a \emph{chimera}, a prefix that splices tokens of different futures and has $p(v)\approx0$,
receives a score comparable to theirs. This holds for \emph{every} product $Q$, however it is trained: its error is
$\TC(Y_{1:d})+\sum_{i\le d}\mathrm{KL}(p_i\,\|\,q_i)$;
\citet{oda2026weaver} likewise trace the drop in acceptance of factorized drafters at larger budgets to their
independence assumption.

\begin{lemma}[Chimeras]
  \label{lem:chimera}
  Let $Q$ be a product distribution, and let $v,w$ be prefixes of depth $d$ that differ in $m$ positions. At least
  $2^{m-1}$ distinct prefixes $u$ of depth $d$ that take, at every position, the token of $v$ or of $w$ satisfy
  $Q(u)\ge\min\{Q(v),Q(w)\}$.
\end{lemma}

Barring ties, a tree ranked by a product can therefore hold two futures that disagree in $m$ places only if it also
pays for $2^{m-1}{-}1$ of their splices, all chimeras when the futures exclude each other. Futures shifted by one token, phrased differently or ordered
differently disagree almost everywhere after they split, so in the extreme a product needs a budget exponential in the
block length to match the optimal tree:

\begin{proposition}[Separation]
  \label{prop:separation}
  Let $p$ put mass $1/M$ on each of $M\ge2$ futures that differ at every position.
  (i) For $\B\ge MD$, the optimal tree accepts $D$ tokens in expectation.
  (ii) For every product $Q$ without ties, the top-$\B$ tree under $Q$ accepts at most $D/M+(1-1/M)(1+\log_2\B)$.
  (iii) If the futures are paired at random across positions, every tree computed from the marginals alone accepts at
  most $\log_M\big((M{-}1)\B{+}M\big)$ on average.
  (iv) Some mixture of $\K\ge M$ products equals $p$, so its $\B$ best prefixes form the optimal tree for every $\B$.
\end{proposition}

With $M{=}2$, $D{=}15$ and $\B{=}256$, the optimal tree accepts all 15 tokens, any product at most 12 and the true marginals, on average over how the futures pair up, at most $\log_2(\B{+}2)\approx8$: better marginals cannot close the gap.

\subsection{Restoring the coupling: condition on tokens or on a future}

Two classical tools make positions dependent again. The chain rule conditions each position on the tokens before it, $Q(v)=\prod_i q(v_i\mid v_{<i})$, as autoregressive drafters and the concurrent fixes of block drafters do.
The alternative is a latent variable $Z$ that names the future, with positions conditionally independent given it,
$\bar Q(v)=\sum_k\pimix_k\prod_i\qk{k,i}(v_i)$. Its benefit is measured by the same total correlation:

\begin{proposition}[A latent future explains the coupling]
  \label{prop:latent}
  For any $Z$ jointly distributed with $Y$, $\TC(Y\mid Z)=\sum_iH(Y_i\mid Z)-H(Y\mid Z)$ satisfies $\TC(Y)-\TC(Y\mid Z)=\sum_{i=1}^{D}I(Y_i;Z)-I(Y;Z)$
  \citep{versteeg2014corex}. The mixture of conditional marginals $\bar p=\sum_kP(Z{=}k)\prod_ip_i(\cdot\mid Z{=}k)$
  satisfies $\mathrm{KL}(p\,\|\,\bar p)\le\TC(Y\mid Z)$, whereas $\mathrm{KL}(p\,\|\,\prod_ip_i)=\TC(Y)$.
\end{proposition}

Because $I(Y;Z)\le\log\K$, a latent with $\K$ values removes at least $\sum_iI(Y_i;Z)-\log\K$ nats of total
correlation: when a uniform $Z$ identifies which of $\K$ futures, distinct at every position, is taken, the reduction
is $(D-1)\log\K$, a coupling that grows linearly with the block. The latent needs no drafted token, so the $\K$ blocks are drafted in one pass, and the verifier, not the
drafter, learns which future is realized.

\subsection{Hedging across futures needs a branching verifier}

A mixture drafter with modes $Q_k(v)=\prod_{i\le d}\qk{k,i}(v_i)$ and prior $\pimix$ can score a prefix by its best
mode,
\begin{equation}
  s(v)=\max_k\big[\log\pimix_k+\log Q_k(v)\big],
  \label{eq:score}
\end{equation}
and keep the $\B$ best prefixes. Let $\theta$ be the $\B$-th largest score and $\mathcal{T}_k(n)$ the top-$n$ tree of
mode $k$ alone.

\begin{proposition}[Union of mode trees]
  \label{prop:union}
  Without ties, the mixture tree equals $\bigcup_k\mathcal{T}_k(n_k)$, where $n_k=|\{v:\log\pimix_k+\log Q_k(v)\ge\theta\}|$ and
  $\sum_kn_k\ge\B$. For every continuation $y$, $\alpha_{\mathcal{T}}(y)=\max_k\alpha_{\mathcal{T}_k(n_k)}(y)$.
\end{proposition}

Ranking by the mixture probability $\bar Q$ would be optimal if the mixture were $p$ (\cref{prop:coverage}); the
best-mode score lower-bounds it, $s(v)\le\log\bar Q(v)\le s(v)+\log\K$, and is close to it when one mode dominates
$v$'s probability. We use the best mode because its tree is exactly this union, which one merged heap computes; the
mixture probability, a log-sum-exp over modes, would add only 0.2--0.8 points of gain at $\B{=}64$ and 256
(\cref{app:rules}). In the setting of \cref{prop:separation}, a product keeps up with point-mass modes only up to $\B=D+M-1$, and the
number of modes a budget can use grows with it (\cref{prop:pointmass} in \cref{app:proofs}).

Whether the hedge is worth its price depends on how predictable the right mode is. Let $A_k(y)$ be the number of
tokens of $y$ matched by the argmax chain of mode $k$. Given $\mathcal{G}$, a chain that follows one mode chosen
before verification obtains at most $\max_k\E[A_k(Y)]$, and a tree that holds all $\K$ chains obtains
$\E[\max_kA_k(Y)]$; the difference
\begin{equation}
  V_\K=\E\big[\max_kA_k(Y)\mid\mathcal{G}\big]-\max_k\E\big[A_k(Y)\mid\mathcal{G}\big]\ \ge0
  \label{eq:hedge}
\end{equation}
is the value of hedging. It vanishes when $\mathcal{G}$ determines the winning mode and is large when several futures are plausible and $\mathcal{G}$ cannot tell them apart.

\textbf{Predictions.} P1 and P3 follow from the results above, exactly so in the separation model
(\cref{prop:separation,prop:pointmass}); P2, P4 and P5 are hypotheses that they motivate.
\textbf{(P1)} The gain requires a branching verifier: a chain that follows one mode, even the best one, forfeits
$V_\K$.
\textbf{(P2)} The gain persists at $T{=}0$, where the uncertainty over futures is the drafter's.
\textbf{(P3)} Over the budgets we test, the gain grows with the budget, and more modes pay off only with larger
budgets.
\textbf{(P4)} The earlier the latent enters the drafter, the more it helps, since in a one-pass drafter positions
coordinate only through attention.
\textbf{(P5)} On the same data, more training saturates a single-product drafter first, while a mixture has a higher ceiling, so the mixture's lead grows.

%% file: sections/method.tex
\section{Multi-Future Drafting}
\label{sec:method}

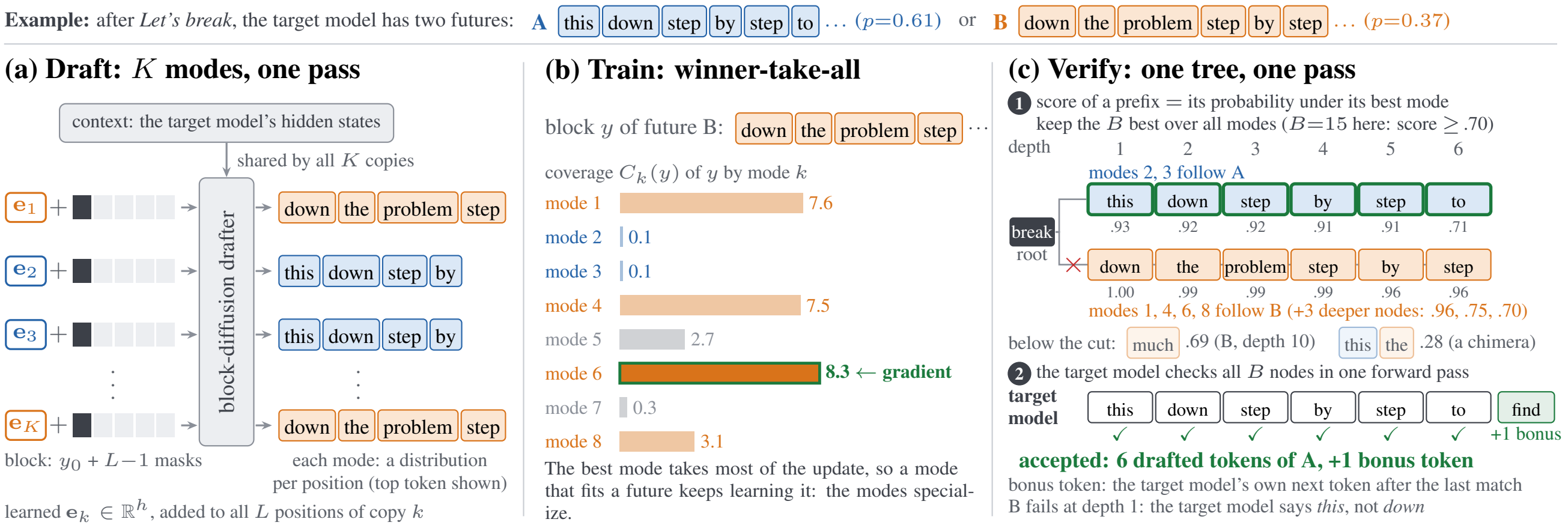
\begin{figure*}[t]
  \centering
  \input{figures/method_fig}
  \caption{\textbf{Multi-Future Drafting} on the example of \cref{fig:teaser}. (a) The block, the last verified token
    $y_0$ and $\Lb{-}1$ mask tokens, is copied $\K$ times, the
    mode embedding $\memb_k$ is added to every position of copy $k$, and all copies are drafted in one forward pass;
    each mode gives a distribution per position. (b) The mode that best covers the target's continuation
    (\cref{eq:softcov}) takes most of the update, so the modes specialize. (c) Every prefix is scored by its best mode
    and the $\B$ best form one tree, which keeps both futures and no chimera.}
  \label{fig:method}
\end{figure*}

\subsection{A block drafter with $\K$ modes}

\mfd{} (\cref{fig:method}) builds on the \dflash{} drafter (\cref{sec:theory}), of width $h$, and adds $\K$ learned
mode embeddings $\memb_1,\dots,\memb_\K\in\mathbb{R}^h$ and a gate. To draft mode $k$, $\memb_k$ is added to the input
embedding of every position of the block, so the latent reaches every position before the first attention layer (P4).
The $\K$ mode blocks are drafted side by side in the same forward pass: their $\K\cdot\Lb$ queries share the context
features, and each attends to the context and to its own block only. Mode $k$ yields its own product distribution
$Q_k(v)=\prod_{i\le d}\qk{k,i}(v_i)$, and the drafter defines the mixture
\begin{equation}
  \bar Q(v)=\sum_{k=1}^{\K}\pimix_k\,Q_k(v),
  \label{eq:mixture}
\end{equation}
where the prior $\pimix$ is uniform in our timed results (within 0.5 points of the gate's, or better;
\cref{app:rules}); the gate, a linear score of each mode's output at the block's first position, picks a single mode
for chain decoding and serves as the prior in the simulations of \cref{tab:hedge,tab:inject} and
\cref{app:rules}. The added parameters are $\K h$ for the embeddings plus $h$ for the gate, about 23K for $\K{=}8$ and
$h{=}2560$.

\subsection{Training: maximize the best mode's coverage}

For a block $y=(y_1,\dots,y_D)$ of a target response, let
\begin{equation}
  C_k(y)=\sum_{d=1}^{D}Q_k(y_{1:d})
  \label{eq:softcov}
\end{equation}
be the soft coverage of $y$ by mode $k$: by \cref{prop:coverage} with drafter and target exchanged, the expected number
of tokens of $y$ that a chain sampled from mode $k$ would match. The \dpace{} loss \citep{dpace2026} is a cross-entropy
whose weight at position $i$ is $w_i=\sum_{j\ge i}\prod_{l\le j}\big((1-\beta)q_l(y_l)+\beta\big)$ (smoothing $\beta{=}0.5$),
computed without gradient; with $\beta{=}0$, $w_i$ is the derivative of the single-mode coverage $C(y)$ with respect to
$\log q_i(y_i)$, so
minimizing the loss is gradient ascent on the soft coverage of the data. \mfd{} applies it inside each mode and weights
each block's loss by $1-\varepsilon$ for the mode with the largest coverage, $k^\star=\argmax_kC_k(y)$, and by
$\varepsilon/(\K-1)$ for every other mode ($\varepsilon{=}0.1$). With $\beta{=}0$ and $\varepsilon{=}0$,
each update is a stochastic subgradient of $\E[\max_kC_k(Y)]$, the soft form of the hedged acceptance $\E[\max_kA_k(Y)]$ of \cref{eq:hedge}. This is the oracle loss of
multiple-choice learning \citep{guzmanrivera2012mcl,lee2016smcl}, whose minimizers partition the continuations into one cell per hypothesis and whose $\varepsilon$ relaxation keeps every mode in use \citep{rupprecht2017uncertain}. Averaging the loss over modes would instead give every mode the same objective, whose optimum
is the single-mode drafter; any rule that trains each mode mainly on the blocks it fits works about as well, including
soft EM (\cref{app:rules}). $\E[\max_kC_k(Y)]$ itself is maximized by $\K$ deterministic chains, which would
leave a tree no alternatives to rank, so we use it as a surrogate, and the smoothing $\beta$ keeps the weights of uncertain
positions from vanishing. The gate is trained with a cross-entropy loss to predict $k^\star$. To keep the training
compute per step equal to the single-mode baseline, which drafts from 512 anchors (positions where a block
starts) per sequence, \mfd{} uses $\K{=}8$ modes $\times$ 64 anchors.

\subsection{Mixture \ddtree{}}

At inference, a prefix is scored by its best mode (\cref{eq:score}), and the tree is the set of $\B$ distinct prefixes
with the highest score. Every mode's score never increases along a path, so, with ties broken toward shorter prefixes,
the selection is prefix-closed and, by \cref{prop:union}, the union of the per-mode trees under a common threshold. It
is computed exactly (\cref{prop:heap}) by merging the $\K$ per-mode best-first searches into one heap: a prefix popped again from another
mode reuses its node and spends no budget, so modes that follow the same future share its nodes (\cref{fig:method}c: two branches of 6 and 9 nodes). With $\K{=}1$ the
procedure reduces to \ddtree{}. Drafting runs once per round with $\K\cdot\Lb$ instead of $\Lb$ queries. It is memory-bound at these sizes, so the dense
layers cost little more; one fused kernel attends over the context and all mode blocks, and the tree is built on the
host from fused top-$k$ statistics. Verification is identical to \ddtree{} at the same budget. On Qwen3-4B, the modes add about 0.6\,ms per round in eager execution (\cref{app:kspeed})
and 0.2\,ms in our SGLang engine (\cref{app:batch}).

%% file: figures/method_fig.tex
\definecolor{mfdBlue}{HTML}{2563A8}\definecolor{mfdBlueL}{HTML}{DBE8F6}
\definecolor{mfdOrange}{HTML}{D9731A}\definecolor{mfdOrangeL}{HTML}{FBE6D3}
\definecolor{mfdGreen}{HTML}{1B7A3D}\definecolor{mfdGray}{HTML}{8A8F98}\definecolor{mfdGrayL}{HTML}{ECEEF1}
\definecolor{mfdInk}{HTML}{3C4048}\definecolor{mfdMuted}{HTML}{5B6068}\definecolor{mfdRule}{HTML}{C9CDD3}\definecolor{mfdRed}{HTML}{C62828}
\begin{tikzpicture}[x=1cm, y=1cm,
    tok/.style={rounded corners=1.2pt, inner xsep=1.7pt, inner ysep=0pt, minimum height=0.34cm,
      text height=1.6ex, text depth=0.45ex, font=\fontsize{6.5}{7}\selectfont, line width=0.4pt},
    tA/.style={tok, fill=mfdBlueL, draw=mfdBlue},
    tB/.style={tok, fill=mfdOrangeL, draw=mfdOrange},
    col/.style={minimum width=0.74cm},
    acc/.style={draw=mfdGreen, line width=1pt},
    lab/.style={font=\scriptsize, text=mfdMuted, inner sep=0pt},
    small/.style={font=\fontsize{6}{7}\selectfont, inner sep=0pt},
    ttl/.style={font=\small\bfseries, anchor=west, inner sep=0pt},
    arr/.style={-{Stealth[length=3.2pt,width=2.6pt]}, mfdGray, line width=0.6pt},
    edge/.style={mfdGray, line width=0.5pt},
  ]
  \node[small, anchor=west, text=mfdInk, font=\fontsize{6.5}{7.5}\selectfont] (hx) at (0,5.3)
    {\textbf{Example:} after \emph{Let's break}, the target model has two futures:};
  \node[anchor=west, inner sep=0pt, text height=1.6ex, text depth=0.45ex, text=mfdBlue, font=\fontsize{7}{8}\selectfont\bfseries, right=5pt of hx] (hA) {A};
  \node[tA, right=3pt of hA] (hA1) {this};
  \node[tA, right=0.6pt of hA1] (hA2) {down};
  \node[tA, right=0.6pt of hA2] (hA3) {step};
  \node[tA, right=0.6pt of hA3] (hA4) {by};
  \node[tA, right=0.6pt of hA4] (hA5) {step};
  \node[tA, right=0.6pt of hA5] (hA6) {to};
  \node[small, text=mfdBlue, anchor=west, font=\fontsize{6.5}{7.5}\selectfont, right=1.5pt of hA6] (hpa)
    {\dots\ ($p{=}0.61$)};
  \node[small, text=mfdMuted, anchor=west, font=\fontsize{6.5}{7.5}\selectfont, right=5pt of hpa] (hor) {or};
  \node[anchor=west, inner sep=0pt, text height=1.6ex, text depth=0.45ex, text=mfdOrange, font=\fontsize{7}{8}\selectfont\bfseries, right=5pt of hor] (hB) {B};
  \node[tB, right=3pt of hB] (hB1) {down};
  \node[tB, right=0.6pt of hB1] (hB2) {the};
  \node[tB, right=0.6pt of hB2] (hB3) {problem};
  \node[tB, right=0.6pt of hB3] (hB4) {step};
  \node[tB, right=0.6pt of hB4] (hB5) {by};
  \node[tB, right=0.6pt of hB5] (hB6) {step};
  \node[small, text=mfdOrange, anchor=west, font=\fontsize{6.5}{7.5}\selectfont, right=1.5pt of hB6]
    {\dots\ ($p{=}0.37$)};
  \draw[mfdRule, line width=0.4pt] (0,5.05) -- (17.1,5.05);
  \draw[mfdRule, line width=0.4pt] (5.72,-0.15) -- (5.72,4.85);
  \draw[mfdRule, line width=0.4pt] (10.88,-0.15) -- (10.88,4.85);

  \node[ttl] at (0,4.75) {(a) Draft: $K$ modes, one pass};
  \node[draw=mfdGray, fill=mfdGrayL, rounded corners=2pt, minimum height=0.4cm, inner xsep=4pt, align=center,
    font=\fontsize{6}{7}\selectfont, text=mfdInk] (ctx) at (2.45,4.18)
    {context: the target model's hidden states};
  \node[draw=mfdGray, fill=mfdGrayL, rounded corners=2pt, minimum width=0.6cm, minimum height=2.95cm] (dr)
    at (2.45,2.1) {};
  \node[rotate=90, font=\scriptsize, text=mfdInk] at (dr.center) {block-diffusion drafter};
  \draw[arr] (ctx.south) -- (dr.north);
  \node[small, text=mfdMuted, anchor=west] at (2.56,3.75) {shared by all $\K$ copies};
  \foreach \y/\e/\clr in {3.25/1/mfdOrange, 2.55/2/mfdBlue, 1.85/3/mfdBlue, 0.85/K/mfdOrange} {
    \node[draw=\clr, fill=white, rounded corners=1.5pt, minimum width=0.44cm, minimum height=0.32cm,
      font=\scriptsize, inner sep=0pt, text=\clr, line width=0.6pt] at (0.24,\y) {$\memb_{\e}$};
    \node[font=\scriptsize, text=mfdMuted] at (0.6,\y) {$+$};
    \fill[mfdInk] (0.76,\y-0.13) rectangle ++(0.2,0.26);
    \foreach \i in {1,...,4} {\fill[mfdGrayL] (0.76+\i*0.23,\y-0.13) rectangle ++(0.2,0.26);}
    \draw[arr] (1.95,\y) -- (2.13,\y);
    \draw[arr] (2.77,\y) -- (2.95,\y);
  }
  \node[font=\scriptsize, text=mfdMuted] at (1.2,1.4) {$\vdots$};
  \node[font=\scriptsize, text=mfdMuted] at (3.9,1.4) {$\vdots$};
  \node[small, text=mfdMuted, anchor=north west] at (0,0.55) {block: $y_0$ + $\Lb{-}1$ masks};
  \node[small, text=mfdMuted, anchor=north west] at (0,0.05)
    {learned $\memb_k\in\mathbb{R}^h$, added to all $\Lb$ positions of copy $k$};
  \node[tB, anchor=west] (a1) at (3.02,3.25) {down};
  \node[tB, right=0.6pt of a1] (a2) {the};
  \node[tB, right=0.6pt of a2] (a3) {problem};
  \node[tB, right=0.6pt of a3] (a4) {step};
  \node[tA, anchor=west] (b1) at (3.02,2.55) {this};
  \node[tA, right=0.6pt of b1] (b2) {down};
  \node[tA, right=0.6pt of b2] (b3) {step};
  \node[tA, right=0.6pt of b3] (b4) {by};
  \node[tA, anchor=west] (c1) at (3.02,1.85) {this};
  \node[tA, right=0.6pt of c1] (c2) {down};
  \node[tA, right=0.6pt of c2] (c3) {step};
  \node[tA, right=0.6pt of c3] (c4) {by};
  \node[tB, anchor=west] (k1) at (3.02,0.85) {down};
  \node[tB, right=0.6pt of k1] (k2) {the};
  \node[tB, right=0.6pt of k2] (k3) {problem};
  \node[tB, right=0.6pt of k3] (k4) {step};
  \node[small, text=mfdMuted, anchor=north, align=center] at (4.25,0.55)
    {each mode: a distribution\\per position (top token shown)};

  \node[ttl] at (5.95,4.75) {(b) Train: winner-take-all};
  \node[lab, anchor=west] at (5.95,4.12) {block $y$ of future B:};
  \node[tB, anchor=west] (y1) at (8.05,4.12) {down};
  \node[tB, right=0.6pt of y1] (y2) {the};
  \node[tB, right=0.6pt of y2] (y3) {problem};
  \node[tB, right=0.6pt of y3] (y4) {step};
  \node[small, text=mfdInk, right=1.5pt of y4] {\dots};
  \node[small, text=mfdMuted, anchor=west] at (5.95,3.62) {coverage $C_k(y)$ of $y$ by mode $k$};
  \foreach \k/\c/\clr in {1/7.6/mfdOrange, 2/0.1/mfdBlue, 3/0.1/mfdBlue, 4/7.5/mfdOrange, 5/2.7/mfdGray,
                          6/8.3/mfdOrange, 7/0.3/mfdGray, 8/3.1/mfdOrange} {
    \pgfmathsetmacro{\yy}{3.3-(\k-1)*0.375}
    \pgfmathsetmacro{\len}{max(0.03,\c/8.3*2.2)}
    \node[small, text=\clr, anchor=west] at (5.95,\yy) {mode \k};
    \ifnum\k=6
      \fill[\clr] (6.78,\yy-0.11) rectangle ++(\len,0.22);
      \draw[mfdGreen, line width=0.9pt] (6.78,\yy-0.11) rectangle ++(\len,0.22);
      \node[small, text=mfdGreen, anchor=west, font=\fontsize{6}{7}\selectfont\bfseries] at (6.78+\len+0.06,\yy)
        {\c\ $\leftarrow$ gradient};
    \else
      \fill[\clr, opacity=0.4] (6.78,\yy-0.11) rectangle ++(\len,0.22);
      \node[small, text=\clr, anchor=west] at (6.78+\len+0.06,\yy) {\c};
    \fi
  }
  \node[small, anchor=south west, text width=4.75cm, align=left, text=mfdInk] at (5.95,-0.19)
    {The best mode takes most of the update, so a mode that fits a future keeps learning it: the modes specialize.};

  \node[ttl] at (11.05,4.75) {(c) Verify: one tree, one pass};
  \node[circle, fill=mfdInk, text=white, inner sep=0.4pt, minimum size=0.26cm, font=\fontsize{5.5}{6}\selectfont\bfseries]
    at (11.18,4.4) {1};
  \node[small, anchor=west, text=mfdInk] at (11.36,4.4) {score of a prefix $=$ its probability under its best mode};
  \node[small, anchor=west, text=mfdInk] at (11.36,4.16)
    {keep the $B$ best over all modes ($B{=}15$ here: score $\ge$\,.70)};
  \node[small, text=mfdMuted, anchor=west] at (11.05,3.9) {depth};
  \foreach \d in {1,...,6} {\node[small, text=mfdMuted] at (12.29+\d*0.745-0.745,3.9) {\d};}
  \node[rounded corners=1.2pt, fill=mfdInk, text=white, minimum height=0.32cm, minimum width=0.5cm,
    font=\fontsize{5.8}{7}\selectfont, inner sep=0pt] (root) at (11.32,2.98) {break};
  \node[small, text=mfdMuted, anchor=north] at (11.32,2.8) {root};
  \foreach \d/\w/\s in {1/this/.93, 2/down/.92, 3/step/.92, 4/by/.91, 5/step/.91, 6/to/.71} {
    \node[tA, acc, minimum width=0.71cm, inner xsep=0.5pt, font=\fontsize{5.8}{7}\selectfont] (A\d)
      at (12.29+\d*0.745-0.745,3.34) {\w};
    \node[text=mfdMuted, font=\fontsize{5}{6}\selectfont, inner sep=0pt] at (12.29+\d*0.745-0.745,3.04) {\s};
  }
  \foreach \d/\w/\s in {1/down/1.00, 2/the/.99, 3/problem/.99, 4/step/.99, 5/by/.96, 6/step/.96} {
    \node[tB, minimum width=0.71cm, inner xsep=0.5pt, font=\fontsize{5.8}{7}\selectfont] (B\d)
      at (12.29+\d*0.745-0.745,2.62) {\w};
    \node[text=mfdMuted, font=\fontsize{5}{6}\selectfont, inner sep=0pt] at (12.29+\d*0.745-0.745,2.32) {\s};
  }
  \draw[edge] (root.east) -- ++(0.04,0) |- (A1.west);
  \draw[edge] (root.east) -- ++(0.04,0) |- (B1.west);
  \foreach \d [evaluate=\d as \n using int(\d+1)] in {1,...,5} {
    \draw[edge] (A\d.east) -- (A\n.west);
    \draw[edge] (B\d.east) -- (B\n.west);
  }
  \node[small, text=mfdBlue, anchor=south west] at (11.935,3.53) {modes 2, 3 follow A};
  \node[small, text=mfdOrange, anchor=north west] at (11.935,2.2)
    {modes 1, 4, 6, 8 follow B (+3 deeper nodes: .96, .75, .70)};
  \node[small, text=mfdMuted, anchor=west] (lo) at (11.05,1.76) {below the cut:};
  \node[tok, draw=mfdOrange!45, fill=mfdOrangeL!45, text=mfdMuted, font=\fontsize{5.8}{7}\selectfont,
    right=3pt of lo] (lo1) {much};
  \node[small, text=mfdMuted, right=1.5pt of lo1] (lo1s) {.69 (B, depth 10)};
  \node[tok, draw=mfdBlue!45, fill=mfdBlueL!45, text=mfdMuted, font=\fontsize{5.8}{7}\selectfont,
    right=7pt of lo1s] (lo2) {this};
  \node[tok, draw=mfdOrange!45, fill=mfdOrangeL!45, text=mfdMuted, font=\fontsize{5.8}{7}\selectfont,
    right=0.6pt of lo2] (lo3) {the};
  \node[small, text=mfdMuted, right=1.5pt of lo3] {.28 (a chimera)};
  \node[circle, fill=mfdInk, text=white, inner sep=0.4pt, minimum size=0.26cm, font=\fontsize{5.5}{6}\selectfont\bfseries]
    at (11.18,1.42) {2};
  \node[small, anchor=west, text=mfdInk] at (11.36,1.42) {the target model checks all $B$ nodes in one forward pass};
  \node[small, anchor=west, text=mfdInk, font=\fontsize{6}{7}\selectfont\bfseries, align=left] at (11.05,1.05)
    {target\\model};
  \foreach \d/\w in {1/this, 2/down, 3/step, 4/by, 5/step, 6/to} {
    \node[tok, fill=white, draw=mfdInk, minimum width=0.71cm, inner xsep=0.5pt, font=\fontsize{5.8}{7}\selectfont] (T\d)
      at (12.29+\d*0.745-0.745,1.05) {\w};
    \node[small, text=mfdGreen] at (12.29+\d*0.745-0.745,0.76) {$\checkmark$};
  }
  \node[tok, fill=mfdGreen!12, draw=mfdGreen, minimum width=0.62cm, inner xsep=0.5pt, font=\fontsize{5.8}{7}\selectfont]
    (T7) at (12.29+6*0.745,1.05) {find};
  \node[small, text=mfdGreen] at (12.29+6*0.745,0.76) {+1 bonus};
  \node[text=mfdRed, font=\fontsize{9}{9}\selectfont\bfseries, inner sep=0pt] at (11.775,2.62) {$\times$};
  \node[font=\scriptsize\bfseries, text=mfdGreen, anchor=west] at (11.05,0.45)
    {accepted: 6 drafted tokens of A, +1 bonus token};
  \node[small, text=mfdMuted, anchor=west] at (11.05,0.18)
    {bonus token: the target model's own next token after the last match};
  \node[small, text=mfdMuted, anchor=west] at (11.05,-0.06)
    {B fails at depth 1: the target model says \emph{this}, not \emph{down}};
\end{tikzpicture}

%% file: sections/experiments.tex
\section{Experiments}
\label{sec:experiments}

\subsection{Setup}

\textbf{Targets and data.} Qwen3-4B and Qwen3-8B (non-thinking) and the hybrid Gated-DeltaNet models Qwen3.5-4B and
Qwen3.6-27B (thinking). All fine-tuned drafters start from the released \dflash{} checkpoints and share one recipe: one
or three epochs on 100K \dflash{} training prompts with responses regenerated by each target (released
ones on Qwen3-4B), and, for the comparison with DARTree, on the 1.42M prompts of DARTree's drafter
(\cref{app:details}).

\textbf{Benchmarks and protocol.} GSM8K, MATH-500, HumanEval, MBPP and MT-Bench at $T{=}0$ and $T{=}1$, from \citet{ringel2026ddtree}, timed with Hugging Face Transformers at batch size 1 on one RTX PRO 6000 GPU; each method
decodes back to back with an autoregressive baseline on each prompt (\cref{app:details}). $\tau$ counts the tokens
committed per round, the bonus token included, and a budget $\B$ counts drafted nodes, as in \ddtree{} and DARTree
\citep{li2026dartree}; hybrid targets use exact tree kernels (\cref{app:hybrid}).

\textbf{Baselines.} EAGLE-3 \citep{li2025eagle3} with released heads; the released \dflash{} drafter as a chain and
under \ddtree{}; \dflash{} fine-tuned with \dpace{} (identical data, schedule and compute);
\dflash{}~2 fine-tuned likewise, as a selector chain and as a tree; and DARTree and JetSpec as
released.

\subsection{Main results}
\label{sec:main}

\begin{table*}[t]
  \centering
  \caption{Speedup over autoregressive decoding and acceptance length $\tau$ (5-task means). Each tree method except
  DARTree is at its best budget, $\B\in\{16,64,128,256\}$ on Qwen3-4B and Qwen3-8B and $\{16,64,256\}$ on the
  thinking-mode targets (\cref{app:details}). \dflash{}~2 + \ddtree{}:
  \dflash{}~2's distributions in the same tree; its selector chain at $T{=}1$ is the faster of its greedy and sampled
  chains. FT: fine-tuned under our recipe for 1 or 3 epochs (on Qwen3.6-27B, the 1-epoch rows are the first-epoch
  checkpoints of the 3-epoch runs). Bottom block: fine-tuned on the 1.42M target-regenerated prompts of Domino
  \citep{domino2026}, the data of DARTree's drafter; DARTree verifies 64 nodes with its released drafter and code.
  Bold: best within each block.}
  \label{tab:main}
  \small
  \setlength{\tabcolsep}{3pt}
  \resizebox{\textwidth}{!}{%
  \begin{tabular}{l l cc cc cc cc}
    \toprule
    & & \multicolumn{2}{c}{Qwen3-4B} & \multicolumn{2}{c}{Qwen3-8B} & \multicolumn{2}{c}{Qwen3.5-4B (thinking)}
      & \multicolumn{2}{c}{Qwen3.6-27B (thinking)} \\
    \cmidrule(lr){3-4}\cmidrule(lr){5-6}\cmidrule(lr){7-8}\cmidrule(lr){9-10}
    Method & Training & $T{=}0$ & $T{=}1$ & $T{=}0$ & $T{=}1$ & $T{=}0$ & $T{=}1$ & $T{=}0$ & $T{=}1$ \\
    \midrule
    EAGLE-3 & released & 4.21$\times$ (6.46) & 3.87$\times$ (6.01) & 3.75$\times$ (6.07) & 3.42$\times$ (5.58) & -- & -- & -- & -- \\
    \dflash{} chain & released & 4.38$\times$ (5.98) & 3.73$\times$ (5.08) & 4.26$\times$ (5.99) & 3.48$\times$ (4.88) & 4.85$\times$ (6.25) & 3.29$\times$ (4.23) & 4.66$\times$ (5.81) & 4.04$\times$ (5.04) \\
    \dflash{}~2 selector chain & FT 1 ep & 4.73$\times$ (6.63) & 4.12$\times$ (5.79) & 4.73$\times$ (6.74) & 4.00$\times$ (5.75) & 5.38$\times$ (7.01) & 3.97$\times$ (5.18) & 4.89$\times$ (6.09) & 4.35$\times$ (5.43) \\
    \ddtree{} & released & 6.08$\times$ (8.59) & 5.44$\times$ (7.70) & 5.73$\times$ (8.18) & 5.00$\times$ (7.13) & 6.53$\times$ (8.78) & 4.77$\times$ (6.51) & 5.89$\times$ (7.62) & 5.27$\times$ (6.86) \\
    \dflash{}~2 + \ddtree{} & FT 1 ep & 6.13$\times$ (8.84) & 5.39$\times$ (7.78) & 5.80$\times$ (8.60) & 4.88$\times$ (7.28) & 6.80$\times$ (9.29) & 4.62$\times$ (6.40) & 5.73$\times$ (7.40) & 5.07$\times$ (6.57) \\
    \dpace{} + \ddtree{} & FT 1 ep & 6.31$\times$ (8.91) & 5.54$\times$ (7.84) & 6.06$\times$ (8.62) & 5.10$\times$ (7.26) & 6.96$\times$ (9.37) & 4.67$\times$ (6.37) & 6.43$\times$ (8.31) & 5.52$\times$ (7.19) \\
    \dpace{} + \ddtree{} & FT 3 ep & 6.22$\times$ (8.78) & 5.39$\times$ (7.63) & 6.04$\times$ (8.60) & 5.03$\times$ (7.15) & 6.97$\times$ (9.38) & 4.66$\times$ (6.35) & 6.46$\times$ (8.35) & 5.50$\times$ (7.17) \\
    \midrule
    \mfd{} ($\K{=}8$) & FT 1 ep & 6.67$\times$ (9.63) & 5.90$\times$ (8.54) & 6.41$\times$ (9.35) & 5.41$\times$ (7.88) & 7.17$\times$ (9.88) & 5.03$\times$ (7.02) & 6.56$\times$ (8.64) & 5.82$\times$ (7.71) \\
    \mfd{} ($\K{=}8$) & FT 3 ep & \textbf{6.73$\times$} (9.71) & \textbf{5.96$\times$} (8.63) & \textbf{6.51$\times$} (9.47) & \textbf{5.43$\times$} (7.89) & \textbf{7.31$\times$} (10.07) & \textbf{5.04$\times$} (7.03) & \textbf{6.77$\times$} (8.91) & \textbf{5.94$\times$} (7.87) \\
    \midrule
    \multicolumn{10}{l}{\emph{Full data: 1.42M target-regenerated prompts}} \\
    DARTree & released & 6.53$\times$ (9.97) & 5.49$\times$ (8.34) & 6.42$\times$ (10.09) & 5.23$\times$ (8.11) & -- & -- & -- & -- \\
    \dpace{} + \ddtree{} & FT 3 ep & 6.66$\times$ (9.34) & 5.66$\times$ (7.96) & 6.43$\times$ (9.18) & 5.26$\times$ (7.48) & 7.31$\times$ (9.82) & 4.74$\times$ (6.46) & 6.74$\times$ (8.72) & 5.69$\times$ (7.40) \\
    \mfd{} ($\K{=}8$) & FT 3 ep & \textbf{7.32$\times$} (10.49) & \textbf{6.35$\times$} (9.13) & \textbf{6.98$\times$} (10.18) & \textbf{5.88$\times$} (8.52) & \textbf{7.78$\times$} (10.72) & \textbf{5.18$\times$} (7.23) & \textbf{7.16$\times$} (9.42) & \textbf{6.16$\times$} (8.16) \\
    \bottomrule
  \end{tabular}}
\end{table*}

\mfd{} is the fastest method for every target, temperature and task (\cref{tab:main} and
\cref{app:pertask}): the three-epoch \mfd{} is 4.8--7.4\% faster than the strongest baseline at $T{=}0$ and
5.7--7.6\% at $T{=}1$; the strongest baseline is \dpace{}+\ddtree{}, except at $T{=}1$ on Qwen3.5-4B, where it is the
released \ddtree{}. The released EAGLE-3 heads reach 3.4--4.2$\times$; their sequential draft takes about 1.7 times as long as the
one-pass \dpace{} draft. The tree of \dflash{}~2 trails \dpace{}+\ddtree{} timed in the same run by 0.6--2.5\% (9--11\% on
Qwen3.6-27B), and the one-epoch \mfd{} is 27--43\% faster than \dflash{}~2's selector chain and 5--15\% faster than its tree. Fine-tuned on the full
1.42M prompts (bottom block), both trees improve, and the mixture's lead grows to 6.2--12.2\%.

\textbf{Concurrent tree drafters.} We time DARTree \citep{li2026dartree} in its released configuration (code, Domino drafters
\citep{domino2026} and 64-node pruned tree) in the same process and on the same prompts as \mfd{} (\cref{tab:main}). Fine-tuned on the same prompts, \mfd{} is
faster on every task, by 9--16\% in the five-task mean. At the same 64 nodes DARTree accepts more tokens per round, but its
per-depth correction takes 2.6--2.7\,ms per round to build the tree, against about 0.35\,ms for \mfd{}
(\cref{app:details}); even with 14 times less fine-tuning data, \mfd{} at its best budget is 1.4--8.7\% faster than
DARTree's 64-node tree. JetSpec's released head trails the released \dflash{} drafter (\cref{app:details}).

\subsection{In a serving engine}
\label{sec:engine}

\begin{table}[t]
  \centering
  \caption{Batch size 1 in SGLang 0.5.22 on one RTX PRO 6000 GPU: decode speedup over SGLang's autoregressive
  decoding (prefill excluded), with $\tau$ in parentheses, 5-task means. Top: released drafters as distributed, each
  at its best setting on SGLang's own engine paths (\cref{app:batch}). Bottom: our tree engine: \ddtree{} with the released \dflash{} drafter, and \dpace{}, \mfd{} and \mfd{}-D (with a single-mode default path) fine-tuned from it on the full data of \cref{tab:main}, at $\B{=}63$. Bold: fastest per
  column.}
  \label{tab:sglang}
  \small
  \setlength{\tabcolsep}{2.5pt}
  \resizebox{\columnwidth}{!}{%
  \begin{tabular}{l cc cc}
    \toprule
    & \multicolumn{2}{c}{Qwen3-4B} & \multicolumn{2}{c}{Qwen3-8B} \\
    \cmidrule(lr){2-3}\cmidrule(lr){4-5}
    & $T{=}0$ & $T{=}1$ & $T{=}0$ & $T{=}1$ \\
    \midrule
    \multicolumn{5}{l}{\emph{Released drafters on SGLang's own engine paths}} \\
    EAGLE-3 & 2.75$\times$ (5.21) & 2.58$\times$ (4.91) & 3.12$\times$ (5.69) & 2.88$\times$ (4.91) \\
    \dflash{} chain & 4.06$\times$ (5.97) & 3.46$\times$ (5.10) & 4.39$\times$ (5.90) & 3.62$\times$ (4.88) \\
    \dflash{}~2 selector chain & 3.37$\times$ (5.20) & 3.07$\times$ (4.72) & 3.57$\times$ (4.94) & 3.26$\times$ (4.52) \\
    \midrule
    \multicolumn{5}{l}{\emph{Tree verification in our SGLang engine}} \\
    \ddtree{} (released drafter) & 4.97$\times$ (7.81) & 4.44$\times$ (6.99) & 5.27$\times$ (7.79) & 4.56$\times$ (6.74) \\
    \dpace{}+\ddtree{} (full data) & 5.52$\times$ (8.72) & 4.72$\times$ (7.48) & 5.99$\times$ (8.84) & 4.82$\times$ (7.15) \\
    \mfd{} (ours) & \textbf{5.93$\times$} (9.62) & 5.06$\times$ (8.23) & \textbf{6.43$\times$} (9.72) & \textbf{5.26$\times$} (7.96) \\
    \mfd{}-D (ours) & 5.89$\times$ (9.59) & \textbf{5.11$\times$} (8.27) & 6.41$\times$ (9.69) & 5.24$\times$ (7.95) \\
    \bottomrule
  \end{tabular}}
\end{table}

We implemented \ddtree{} and the mixture tree in SGLang 0.5.22 \citep{zheng2024sglang}, verifying trees inside the
target's CUDA graphs; both draft over the 64K most frequent tokens of target responses and verify over the full
vocabulary, so decoding stays lossless (\cref{app:batch}). At batch size 1
(\cref{tab:sglang}), \mfd{} is 45--46\% faster than the fastest released drafter, the \dflash{} chain on SGLang's own engine paths, and 7--9\% faster than \dpace{} fine-tuned on the same data in our engine, which shares every optimization. Its lead is a small-batch one: with as many requests in flight as the batch size and the drafters' linear layers in FP8 (\cref{app:batch}), it is
10--18\% faster than the released options and \ddtree{} with the released drafter at batch 1--2 and 3--11\% faster at batch 4, but falls behind that tree from
batch 8 and SGLang's own paths from batch 16, because its draft cost grows with the batch while its acceptance length does not. \mfd{}-D, which switches to a single-mode default path from three requests, is the fastest
configuration we timed at every batch size.

%% file: sections/analysis.tex
\section{Analysis: Testing the Predictions}
\label{sec:analysis}

We first measure where the budget of a product tree goes (\cref{sec:budget}), then test the predictions of
\cref{sec:theory}: the gain needs a branching verifier, persists at $T{=}0$ and grows with the budget (P1--P3;
\cref{sec:hedge,sec:alternatives}), is larger the earlier the modes enter (P4; \cref{sec:inject}), and grows with training as a single-mode drafter saturates (P5; \cref{sec:epochs}). \Cref{sec:hedge,sec:inject} use paired simulations on Qwen3-4B with quarter-epoch drafters and report the change in $\tau$ relative to a single mode trained the same way (\cref{app:details}); the other analyses use the drafters of
\cref{tab:main} unless noted. $\tau$ gains run about two points above speed gains, which also pay for drafting the modes (\cref{app:kspeed}).

\subsection{Where the budget of a product tree goes}
\label{sec:budget}

\begin{figure}[t]
  \centering
  \includegraphics[width=0.85\columnwidth]{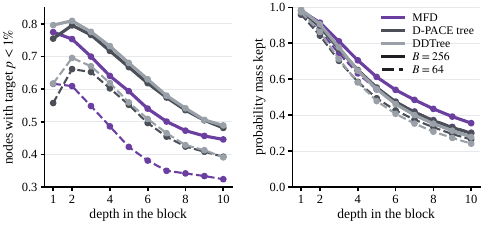}
  \caption{Where the budget goes (Qwen3-4B, 24 prompts per task, $T{=}0$ contexts, $T{=}1$ probabilities from the
    verify pass; the three-epoch \mfd{}, the strongest \dpace{} (one epoch) of \cref{tab:main} and the released
    \ddtree{}). Left: share of the nodes at each depth whose last step the target takes with probability below 1\%
    (chimeras and dead steps, \cref{app:chimeras}). Right: probability mass kept at each depth. From depth 2 on, where chimeras
    first form, the product trees waste more nodes and keep less mass.}
  \label{fig:budget}
\end{figure}

Verification gives the exact $p(v)$ of every drafted prefix at $T{=}1$, with $\mathcal{G}$ the whole context. Over 24
prompts per task (Qwen3-4B), \mfd{} trees cover 8.3\% more probability mass than \dpace{}'s at $\B{=}64$ and
10.3\% at $\B{=}256$ (6.78 vs.\ 6.26 and 7.58 vs.\ 6.87 accepted drafted tokens per round): 64 \mfd{} nodes cover nearly as much as 256 \dpace{} nodes. Both put about as many nodes at each depth, but at depths 2--8, 4--13 points more of the product tree's nodes are steps with target probability below 1\% (\cref{fig:budget}); at depth 1, before any chimera can form, it is no worse. Where the target forks (a runner-up token of at least 10\%; 60--71\% of anchors),
\mfd{} accepts 10--12\% more tokens than \dpace{}, and 2--4\% more elsewhere ($\B{=}64$). At equal difficulty the
gain is about as large where the target is confident: it tracks where a single-mode tree stops early, mostly at forks (\cref{app:chimeras}). Mixture trees still hold chimeras (22--24\% of nodes at $\B{=}64$ vs.\ 27--33\%), but forks add 54--57\% fewer.

\subsection{Hedging needs a branching verifier (P1--P3)}
\label{sec:hedge}

\begin{table}[t]
  \centering
  \caption{One drafter, several ways to decode it ($T{=}0$; change in $\tau$ relative to single-mode \dpace{} decoded
  the same way, as a chain or as a tree with the same $\B$). Gate: the mode the gate picks before verification.
  Oracle: each block's best mode, known only after verification. Mixture: the tree of \cref{eq:score}, with the gate's
  prior. The multi-mode drafters train on 1024 draft blocks per sequence against 512 for \dpace{}; at a matched 512,
  $\K{=}4$ gains 5.0\% / 5.6\% at $\B{=}64$ ($T{=}0$ / $T{=}1$). Bold: the best $\K$ at each budget.}
  \label{tab:hedge}
  \scriptsize
  \setlength{\tabcolsep}{3.2pt}
  \begin{tabular*}{\columnwidth}{@{\extracolsep{\fill}} r cc c ccc c}
    \toprule
    & \multicolumn{2}{c}{Chain} & \multicolumn{5}{c}{Tree} \\
    \cmidrule(lr){2-3}\cmidrule(lr){4-8}
    & Gate & Oracle & Gate & \multicolumn{3}{c}{Mixture} & Oracle \\
    $\K$ & & & $\B{=}64$ & $\B{=}16$ & $\B{=}64$ & $\B{=}512$ & $\B{=}64$ \\
    \midrule
    4  & $-3.1$ & $+20.2$ & $-4.9$  & \textbf{+2.5} & $+5.4$ & $+6.2$ & $+13.0$ \\
    8  & $-4.6$ & $+27.4$ & $-7.3$  & $+2.3$ & $+6.4$ & $+7.8$ & $+17.7$ \\
    16 & $-7.2$ & $+34.7$ & $-10.4$ & $+1.0$ & \textbf{+6.6} & $+8.4$ & $+21.5$ \\
    32 & $-8.8$ & $+40.2$ & $-12.1$ & $+0.0$ & $+6.1$ & \textbf{+8.7} & $+24.4$ \\ \bottomrule \end{tabular*}
\end{table}

\Cref{tab:hedge} decodes \mfd{} drafters with $\K\in\{4,8,16,32\}$ in several ways. As a chain following the gate's choice, every mixture is \emph{worse} than the single-mode drafter, increasingly so with $\K$: each mode is
narrower than the averaged drafter, and the gate picks the winning mode for only 38\% of training blocks at
$\K{=}4$ (chance: 25\%) and 10\% at $\K{=}32$ (\cref{app:rules}). Yet the modes contain the right futures: if each block followed its best mode, the chain would be 20--40\% longer. This gap is the hedging value $V_\K$ of \cref{eq:hedge} plus the gate's
selection error, and a learned prior as in \citet{dblast2026} gains nothing from committing to one mode
either (\cref{app:rules}), so most of it is hedging value that no chain can collect. A tree collects part of it without
knowing the mode: spending the whole budget on the gate's mode loses 5--12\%, while the mixture tree gains up to 8.7\% (its drafters train on twice \dpace{}'s draft blocks).
The gain grows with the budget, and larger $\K$ pays off only at larger budgets: $\K{=}4$ leads $\K{=}32$ at $\B{=}16$
and trails it at $\B{=}512$, since with a small budget each mode's share is too small to reach useful depth
(\cref{prop:union}). Under the log-sum-exp rule, many modes lose less at
small budgets, but the best $\K$ still grows from 8 at $\B{=}16$ to 32 at $\B{=}256$
(\cref{app:rules}). Drafting time caps $\K$ at 8--16 (\cref{app:kspeed}); we use 8.

The gain persists at $T{=}0$ (P2): \mfd{} leads at both temperatures (\cref{sec:main}), and the simulated mixture
gains of \cref{tab:hedge} are only 0.5--1.4 points larger at $T{=}1$ (\cref{tab:rules}). At $T{=}0$ the continuation is deterministic, but the drafter's uncertainty about
it is not, and this uncertainty is what the modes represent; a tree hedges over it at any temperature, while a mixture
that decodes a single sampled chain \citep{dblast2026} can use it only through the choice of mode.

\subsection{Why the tree must be a mixture}
\label{sec:alternatives}

Simpler uses of the same modes fall short (\cref{tab:alt} in \cref{app:modes}). Committing to the gate's mode makes a
worse tree than \dpace{} (7.52 vs.\ 8.06 at 64 nodes on Qwen3-4B). Verifying the eight modes' argmax chains, with
up to 120 nodes, only matches \dpace{}'s 64-node tree, since chains repeat shared prefixes; one tree of $\B/8$ nodes per mode stays below the mixture tree even with twice its nodes. The gain lies in ranking whole
prefixes: ranked as a product, the same drafter's mean marginals fall 2--3\% below \dpace{} at 64
nodes, and the mixture tree is 11\% above them. Where their chains first disagree, one mode favours \emph{Therefore}, another \emph{This}, another a
closing \texttt{\$.} (\cref{tab:forks}). The modes must differ, and the drafter must learn to use them: with every
embedding set to the table's mean, the mixture falls 2\% below \dpace{}, and random offsets with the trained norms keep
60--80\% of the gain on the \mfd{} drafter but 12--24\% on \dpace{}'s, trained without modes: most of the
gain lives in the drafter's weights, not in the particular embeddings.

\subsection{Modes help most when they enter early (P4)}
\label{sec:inject}

\begin{table}[t]
  \centering
  \caption{Where the modes enter ($\K{=}4$, $T{=}0$; change in $\tau$ relative to \dpace{}). Top: mode embeddings added at
  different depths of the five-layer drafter, or a low-rank per-mode map of the output. Bottom: input embeddings
  against a full-rank per-mode output expansion with a learned prior \citep{dblast2026}, both trained with the mixture
  NLL, relative to a single mode trained with plain NLL. Params: parameters added, including the gate or prior head.
  Oracle chain: each block's best mode.}
  \label{tab:inject}
  \scriptsize
  \setlength{\tabcolsep}{3pt}
  \begin{tabular*}{\columnwidth}{@{\extracolsep{\fill}} l r cc c}
    \toprule
    Injection & Params & $\B{=}64$ & $\B{=}512$ & Oracle chain \\
    \midrule
    Input (layer 0)              & 13K  & \textbf{+5.0} & \textbf{+5.9} & $+19.1$ \\
    Layer 2                      & 13K  & $+4.1$ & $+5.0$ & $+18.0$ \\
    Layer 4 (last)               & 13K  & $+1.5$ & $+1.9$ & $+14.8$ \\
    Output, rank 128             & 2.6M & $-0.6$ & $-0.2$ & $+12.9$ \\
    \midrule
    Input embeddings             & 13K  & \textbf{+7.0} & \textbf{+7.4} & -- \\
    Output, full rank            & 26M & $+2.5$ & $+2.6$ & -- \\ \bottomrule \end{tabular*}
\end{table}

\Cref{tab:inject} moves the modes through the drafter. The gain falls monotonically as the modes enter later, and a
per-mode map of the output, the design of output-side mixture heads, gives nothing; under the same mixture objective,
input embeddings give about three times the gain of a full-rank output expansion with a learned prior
\citep{dblast2026}, with three orders of magnitude fewer parameters. In a one-pass drafter, positions coordinate only
through attention, so a mode that enters after the last attention layer can no longer coordinate them unless the
shared layers already line up the alternatives, which our output-side drafters do not learn. After their argmax chains split, input-injected modes still differ at 61\% of the remaining
positions, against 43--46\% for the last layer and the output map, whose local swaps a single-mode tree covers cheaply (\cref{lem:chimera}; \cref{tab:disagree}).

\subsection{Single-mode drafters saturate before mixtures (P5)}
\label{sec:epochs}

Fine-tuning for three epochs instead of one leaves \dpace{} unchanged or slightly worse on every
target, while \mfd{} improves (\cref{tab:main}); in simulation at $\B{=}256$, \mfd{}'s lead over \dpace{} grows from
+7.7\% / +8.9\% after the first of three epochs to +9.7\% / +10.8\% after the third on Qwen3-4B
($T{=}0$ / $T{=}1$; \cref{tab:epochs}). Trained from scratch with \dflash{}'s schedule, \mfd{} starts 6.0\% behind and leads from epoch four, by 3.9\% / 5.0\% at epoch six,
after the single-mode drafter peaks, as P5 predicts. The slower approach may partly come from the data each mode sees: 64 anchors per sequence against 512 for \dpace{}. Fine-tuned on 14 times more data, both improve and the lead grows to 6--12\%.

%% file: sections/conclusion.tex
\section{Conclusion}
\label{sec:conclusion}

A product of per-position distributions cannot keep a block's futures apart; a latent future can. \mfd{} adds $\K$
learned modes to a one-pass drafter and ranks all their futures' prefixes in one tree. The gains need a
branching verifier, grow with budget and training, and extend to hybrid targets from a few dozen nodes on (\cref{tab:hybrid-budget}). Mixture drafting and conditioning on drafted tokens close the same gap: on
a drafter that already conditions, modes add little (\cref{sec:combine}), and trained on the same prompts, \mfd{}
outpaces DARTree in one parallel pass; \cref{app:batch} covers its large-batch limits.

%% file: sections/impact.tex
\section*{Impact Statement}

This paper presents work whose goal is to advance the field of Machine Learning. \mfd{} speeds up decoding
losslessly: the outputs follow the target model's own distribution, so it adds no capability and no risk beyond the
target's, while it lowers the latency and the compute spent per generated token. There are many potential societal
consequences of our work, none of which we feel must be specifically highlighted here.

%% file: sections/proofs.tex
\section{Proofs}
\label{app:proofs}

\textbf{Proof of \cref{prop:coverage}.} Because $\mathcal{T}$ is prefix-closed, $Y_{1:d}\in\mathcal{T}$ implies
$Y_{1:d'}\in\mathcal{T}$ for every $d'<d$, so
$\alpha_{\mathcal{T}}(Y)=\sum_{d=1}^{D}\mathbb{I}[Y_{1:d}\in\mathcal{T}]=\sum_{v\in\mathcal{T}}\mathbb{I}[Y_{1:|v|}=v]$,
and taking expectations gives the identity. For the second claim, $p(v)\ge p(vw)$ for every extension $vw$ of $v$,
because $\{Y_{1:|vw|}=vw\}\subseteq\{Y_{1:|v|}=v\}$. Order the prefixes by decreasing $p$, breaking ties in favour of
shorter prefixes, and let $S$ be the first $\B$. If $u\in S$ and $v$ is a proper prefix of $u$, then $p(v)\ge p(u)$ and
$v$ is shorter, so $v$ precedes $u$ and $v\in S$: $S$ is a tree. Since $\sum_{v\in\mathcal{T}}p(v)$ is a sum of
nonnegative weights, $S$ maximizes it over all sets of at most $\B$ prefixes, and in particular over trees.\qed

\textbf{Proof of \cref{eq:ptc}.} Since $\tilde p(v)=\prod_{i\le d}p_i(v_i)$,
$\E[\log p(Y_{1:d})-\sum_{i\le d}\log p_i(Y_i)]=\sum_{i\le d}H(Y_i)-H(Y_{1:d})=\TC(Y_{1:d})$, and
$\TC(Y_{1:d+1})-\TC(Y_{1:d})=H(Y_{d+1})-H(Y_{1:d+1})+H(Y_{1:d})=H(Y_{d+1})-H(Y_{d+1}\mid Y_{1:d})
=I(Y_{d+1};Y_{1:d})$.\qed

\textbf{Proof of \cref{lem:chimera}.} Let $P\subseteq\{1,\dots,d\}$ be the $m\ge1$ positions where $v$ and $w$
differ. For $S\subseteq P$, let $u^S$ take the token of $v$ at the positions in $S$ and the token of $w$ at all other
positions (outside $P$ the two agree). The $2^m$ prefixes $u^S$ are distinct, $u^P=v$ and $u^\emptyset=w$, and because
$Q$ is a product,
\[
  Q(u^S)\,Q(u^{P\setminus S})=\prod_{i\in P}q_i(v_i)\,q_i(w_i)\prod_{i\notin P}q_i(v_i)^2=Q(v)\,Q(w).
\]
Hence $\max\{Q(u^S),Q(u^{P\setminus S})\}\ge\sqrt{Q(v)Q(w)}\ge\min\{Q(v),Q(w)\}$. The $2^m$ subsets form $2^{m-1}$
complementary pairs $\{S,P\setminus S\}$, and each pair contributes a distinct prefix with this property.\qed

\textbf{Proof of \cref{prop:separation}.} Let $f^1,\dots,f^M$ be the futures. Because they differ at the first
position, every prefix of $f^j$ has probability $1/M$, and every other prefix has probability $0$. For any tree,
$\sum_{v\in\mathcal{T}}p(v)=\frac1M\sum_jd_j$, where $d_j$ is the depth to which $f^j$ is contained in $\mathcal{T}$.

(i) For $\B\ge MD$, the $M$ chains $f^j_{1:D}$ form a tree with $MD$ nodes, so every $d_j=D$.

(ii) Without ties, the top-$\B$ tree under $Q$ is $\mathcal{T}=\{u:Q(u)\ge\vartheta\}$, where $\vartheta$ is the $\B$-th
largest value of $Q$. If $d_j\ge d$ and $d_l\ge d$ for some $j\ne l$, \cref{lem:chimera} with $m=d$ yields $2^{d-1}$
prefixes of depth $d$ with $Q(u)\ge\min\{Q(f^j_{1:d}),Q(f^l_{1:d})\}\ge\vartheta$, all of which are in $\mathcal{T}$, so $2^{d-1}\le\B$. At most
one future therefore has $d_j>1+\log_2\B$, and $\frac1M\sum_jd_j\le\frac1M\big(D+(M-1)(1+\log_2\B)\big)$.

(iii) At position $i$, let the futures use the tokens $a_{i,1},\dots,a_{i,M}$, assigned to them by independent,
uniformly random permutations. Every pairing has the same marginals, uniform on $\{a_{i,1},\dots,a_{i,M}\}$, so a tree
computed from the marginals is the same tree $\mathcal{T}$ for every pairing. A prefix $v$ of depth $d$ whose tokens are
among the $a_{i,\cdot}$ is the prefix of a given future with probability $M^{-d}$ over the pairing, so
$\E[p(v)]=\frac1M\cdot M\cdot M^{-d}=M^{-d}$; every other prefix has probability $0$. The average acceptance of
$\mathcal{T}$ is thus $\sum_{v\in\mathcal{T}}M^{-|v|}$, summed over prefixes of the first kind. A tree holds at most
$M^d$ of them at depth $d$, each worth $M^{-d}$, so the sum is largest when depths are filled in increasing order.
Filling depths $1,\dots,k$ takes $N_k=(M^{k+1}-M)/(M-1)$ nodes and yields $k$. If $N_k\le\B<N_{k+1}$, the remaining
$\B-N_k<M^{k+1}$ nodes at depth $k+1$ yield less than $1$, so the average is below
$k+1\le\log_M\big((M-1)\B+M\big)\le1+\log_M\B$ for $\B\ge M$ (for $\B<M$ it is below $1$).

(iv) Let the first $M$ components be the point masses on $f^1,\dots,f^M$, which are products, with weight $1/M$, and
give any further components weight $0$. The mixture equals $p$, and by \cref{prop:coverage} its $\B$ best prefixes form
the optimal tree.\qed

\textbf{Proof of \cref{prop:latent}.} By definition $\TC(Y)=\sum_iH(Y_i)-H(Y)$ and
$\TC(Y\mid Z)=\sum_iH(Y_i\mid Z)-H(Y\mid Z)$, so the difference is
$\sum_i[H(Y_i)-H(Y_i\mid Z)]-[H(Y)-H(Y\mid Z)]=\sum_iI(Y_i;Z)-I(Y;Z)$. Writing $p=\sum_kP(Z{=}k)\,p(\cdot\mid Z{=}k)$,
the joint convexity of the KL divergence gives
\[
  \mathrm{KL}(p\,\|\,\bar p)\le\sum_kP(Z{=}k)\,
  \mathrm{KL}\Big(p(\cdot\mid Z{=}k)\,\Big\|\,\prod_ip(\cdot\mid Z{=}k)\Big)
  =\sum_kP(Z{=}k)\,\TC(Y\mid Z{=}k)=\TC(Y\mid Z),
\]
where the $i$-th factor of the product is the conditional marginal of $Y_i$. Finally,
$\mathrm{KL}(p\,\|\,\prod_ip_i)=\sum_iH(Y_i)-H(Y)=\TC(Y)$.\qed

\textbf{Other priors.} The same argument applied to the joint law of $(Z,Y)$ gives, for any prior $\pimix$ over
the values of $Z$, $\mathrm{KL}\big(p\,\|\,\sum_k\pimix_k\prod_ip_i(\cdot\mid Z{=}k)\big)\le\mathrm{KL}(P_Z\,\|\,\pimix)
+\TC(Y\mid Z)$, which is $\log\K-H(Z)+\TC(Y\mid Z)$ for the uniform prior of the timed trees.

\textbf{From likelihood to acceptance.} \Cref{eq:ptc} and \cref{prop:latent} bound likelihood ratios, while
acceptance depends only on how a tree ranks prefixes. For any law $Q$ of the block, the $\B$ prefixes with the
largest $Q(v)$ form a tree $\mathcal{T}_Q$, since $Q$ never increases along a path, and this tree maximizes
$\sum_{v\in\mathcal{T}}Q(v)$ by the argument of \cref{prop:coverage}. Let $\mathcal{T}^\star$ be the optimal tree and
$p_d$, $Q_d$ the laws of the depth-$d$ prefix. The prefixes in exactly one of the two trees form disjoint sets at
each depth, so by Pinsker's inequality
\[
  \E[\alpha_{\mathcal{T}^\star}(Y)]-\E[\alpha_{\mathcal{T}_Q}(Y)]
  \le\sum_{v\in\mathcal{T}^\star\setminus\mathcal{T}_Q}\big(p(v)-Q(v)\big)+\sum_{v\in\mathcal{T}_Q\setminus\mathcal{T}^\star}\big(Q(v)-p(v)\big)
  \le\sum_{d=1}^{D}\|p_d-Q_d\|_1\le\sum_{d=1}^{D}\sqrt{2\,\mathrm{KL}(p_d\,\|\,Q_d)}.
\]
For the product of the true marginals, the divergence at depth $d$ is $\TC(Y_{1:d})$ (\cref{eq:ptc}); for the
mixture of conditional marginals it is at most $\TC(Y_{1:d}\mid Z)$ (\cref{prop:latent} applied to $Y_{1:d}$). A latent
that explains the coupling therefore also bounds the acceptance lost to ranking. The bound covers trees that rank prefixes by $Q$ itself, for the mixture the log-sum-exp rule of \cref{eq:lse} rather than the max rule of the timed trees, and it is informative only while the divergences are small: in the separation model, $\TC(Y_{1:d})=(d-1)\ln M$, so the product's term exceeds 1 from depth 2 on (\cref{prop:separation} bounds that gap directly), while the mixture's vanishes.

\textbf{Proof of \cref{prop:union}.} Let $S_k=\{v:\log\pimix_k+\log Q_k(v)\ge\theta\}$. Since $Q_k$ never increases along
paths, $S_k$ is prefix-closed and consists of the $n_k=|S_k|$ prefixes with the largest $Q_k$, that is,
$S_k=\mathcal{T}_k(n_k)$. A prefix has $s(v)\ge\theta$ exactly when it belongs to some $S_k$, so, without ties, the
mixture tree $\{v:s(v)\ge\theta\}$ equals $\bigcup_kS_k$, and $\B=|\bigcup_kS_k|\le\sum_kn_k$. For a continuation $y$,
let $d=\alpha_{\mathcal{T}}(y)$. Then $y_{1:d}\in S_k$ for some $k$, and because $S_k$ is prefix-closed,
$\alpha_{S_k}(y)\ge d$; conversely, $S_k\subseteq\mathcal{T}$ gives $\alpha_{S_k}(y)\le\alpha_{\mathcal{T}}(y)$ for every
$k$.\qed

\textbf{Further results in the separation model.} Two further results hold in the setting
of \cref{prop:separation}; \cref{sec:theory} uses the second.

\begin{corollary}[One deep future per mode]
  \label{cor:kmodes}
  In the setting of \cref{prop:separation}(ii), the tree of a mixture of $\K\le M$ products without ties, ranked by
  \cref{eq:score}, accepts at most $\frac1M\big(\K D+(M-\K)(1+\log_2\B)\big)$ in expectation.
\end{corollary}

\begin{proposition}[Point-mass modes]
  \label{prop:pointmass}
  In the setting of \cref{prop:separation}, no tree with $\B$ nodes accepts more than $\min(\B,MD)/M$ in expectation.
  A mixture of $\K\le M$ modes, each a point mass on a different future, with a uniform prior, ties broken toward
  shorter prefixes and only prefixes of positive probability kept, accepts $\min(\B,\K D)/M$, which is optimal for
  $\B\le\K D$. Some product accepts as much for $\B\le D+M-1$; for $D+M-1<\B\le\K D$, every product without ties
  accepts less.
\end{proposition}

A further mode therefore pays off only once the others cannot use the budget ($\B>(\K-1)D$), so the number of modes
the budget can use, $\min(M,\lceil\B/D\rceil)$, grows with $\B$. A product keeps up only while the budget is small:
beyond $D+M-1$ nodes it pays for chimeras, and \cref{prop:separation}(ii) caps it at $D/M+(1-1/M)(1+\log_2\B)$, while
the mixture gains $1/M$ per node up to $\K D$.

\textbf{Unequal futures.} The separation model gives every future the same mass, the case a uniform prior fits. If
future $f^j$ has mass $\mu_j$, point-mass modes with prior $\mu$ score every prefix by its log-probability and still form
the optimal tree, but a uniform prior splits the budget evenly across the futures and can lose to a product: with
$M{=}\K{=}2$, $\mu=(0.9,0.1)$ and $\B{=}D{=}15$, it accepts at most 7.9 tokens in expectation, while a product
concentrated on $f^1$ accepts 13.5. The timed trees use a uniform prior, which is within 0.5 points of the trained
gate's prior or better (\cref{app:rules}).

\textbf{Proof of \cref{cor:kmodes}.} By \cref{prop:union} the tree is $\bigcup_k\mathcal{T}_k(n_k)$, and
$n_k\le\B$ because each $\mathcal{T}_k(n_k)$ lies in the tree. Each $\mathcal{T}_k(n_k)$ is the top-$n_k$ tree of the
product $Q_k$, so by the argument of \cref{prop:separation}(ii) above it contains at most one future beyond depth
$1+\log_2n_k\le1+\log_2\B$. A future's depth in the union is its largest depth in any $\mathcal{T}_k(n_k)$, so at most
$\K$ futures are contained beyond depth $1+\log_2\B$, and $\frac1M\sum_jd_j\le\frac1M\big(\K D+(M-\K)(1+\log_2\B)\big)$.
\qed

\textbf{Proof of \cref{prop:pointmass}.} As in the proof of \cref{prop:separation}, a tree accepts
$\frac1M\sum_jd_j$, which is $1/M$ times the number of its nodes that lie on a future; the futures differ at every
position, so no node lies on two of them, and this is at most $\min(\B,MD)/M$. Let mode $k$ be a point mass on future
$f^{j_k}$. Every prefix of $f^{j_k}$ scores $-\log\K$ and every other prefix $-\infty$, so the tree takes the $\K$
prefixes of depth 1, then those of depth 2, and so on, until it holds $\B$ nodes or all $\K D$ prefixes of positive
probability. All its nodes lie on futures, so it accepts $\min(\B,\K D)/M$.

For the product, let every $q_i$ put $1-\epsilon$ on $f^1_i$ and $\epsilon/(M-1)$ on the token of each other future,
with $(1-\epsilon)^D>\epsilon/(M-1)$. The prefixes of $f^1$ then score highest, the first tokens of the other futures
next, and every other prefix at most $(1-\epsilon)\epsilon/(M-1)$, since it contains a token of another future and a
further factor of at most $1-\epsilon$. For $\B\le D+M-1$ the top-$\B$ tree is therefore $f^1$ to depth $\min(\B,D)$
and, if $\B>D$, $\B-D$ first tokens of other futures; all its nodes lie on futures, so it accepts $\B/M$. Conversely,
let $\B>D+M-1$ and let $Q$ be a product without ties. If its tree holds at most one future beyond depth 1, at most
$D+M-1<\B$ of its nodes lie on futures. Otherwise it holds the depth-2 prefixes of two futures, which differ at both
positions, so by \cref{lem:chimera} it also holds one of their two chimeras of depth 2, which has probability 0. Either
way it accepts at most $(\B-1)/M$, less than the $\B/M$ of the mixture for $\B\le\K D$.\qed

\textbf{The training update.} For one block, let $r_k$ be the assignment weights and $w_{k,i}$ the \dpace{}
weights, both computed without gradient; the loss is $\sum_kr_k\sum_iw_{k,i}\,(-\log\qk{k,i}(y_i))$. Without
smoothing, $w_{k,i}=\sum_{j\ge i}Q_k(y_{1:j})=\partial C_k(y)/\partial\log\qk{k,i}(y_i)$, so the gradient of the loss is
$-\sum_kr_k\nabla C_k(y)$. With $\varepsilon=0$, $r$ is the indicator of $k^\star=\argmax_kC_k(y)$, and the gradient is
$-\nabla C_{k^\star}(y)$, a Clarke subgradient of $-\max_kC_k(y)$. Blocks are drawn from target responses, so each update
is a stochastic subgradient step on $\E[\max_kC_k(Y)]$ (\cref{sec:method}).

%% file: sections/appendix.tex
\section{Tree rules for the mixture}
\label{app:rules}

The mixture tree keeps the $\B$ distinct prefixes with the largest best-mode score $s(v)$ of \cref{eq:score}. This
section shows that the merged heap of \cref{sec:method} computes it exactly, compares the max rule with scoring
prefixes by their mixture probability, and measures what committing to one mode costs.

\textbf{Merged heap.} \ddtree{} enumerates the prefixes of one product $Q_k$ best first. Each position keeps its $\B$
most likely tokens, and a heap holds candidate prefixes keyed by score; popping a prefix whose last token has rank $r$
pushes its next sibling, which replaces that token by the one of rank $r{+}1$, and its first child, which appends the
most likely token of the next position. Every prefix is pushed once, by its predecessor (its previous sibling, or its
parent if it is a first child), which scores at least as high. The merged heap holds entries $(k,v)$ with key
$s_k(v)=\log\pimix_k+\log Q_k(v)$ and starts from each mode's most likely first token. Popping $(k,v)$ adds a node for
$v$ unless the tree already holds $v$, and pushes the successors of $v$ under mode $k$ either way; the search stops
when the tree has $\B$ nodes.

\begin{proposition}[Merged heap]
  \label{prop:heap}
  Up to ties, the merged heap returns the $\B$ distinct prefixes with the largest $s(v)$ and pops at most $\K\B$
  entries. With $\K{=}1$ it performs the pops and pushes of \ddtree{} in the same order and returns its tree.
\end{proposition}

\textbf{Proof.} Let $\sigma$ be the key of some pop and $e$ an entry with a larger key. Following predecessors from $e$
leads to an initial entry through entries whose keys are at least that of $e$, hence above $\sigma$. If $e$ had not
been popped before, the first unpopped entry on this chain would be in the heap, because its predecessor was popped or
it is initial, and the heap would hold a key above $\sigma$, contradicting the heap order. Entries are therefore popped
in nonincreasing key order, and each prefix $v$ is first popped at the largest key $s_k(v)$ among the modes that
enumerate it. Mode $k$ enumerates every prefix whose tokens are among its $\B$ most likely at each position. If the
best mode $k$ of $v$ ranks some $v_i$ below its top $\B$, replacing $v_i$ by each of those $\B$ tokens gives $\B$ other
prefixes $w$ with $s(w)\ge s_k(w)\ge s_k(v)=s(v)$, so $v$ is not among the $\B$ largest. Every prefix among them is
thus enumerated by its best mode and first popped at key $s(v)$, and since nodes are created in the order of first
pops, the $\B$ nodes are the $\B$ largest prefixes under $s$. Every popped entry belongs to a prefix in the tree, and
each mode enumerates a prefix once, so at most $\K\B$ entries are popped. With $\K{=}1$, $\pimix_1=1$, every prefix has
one entry and no node is reused, so the pops and pushes are those of \ddtree{}.\qed

The implementation agrees: on 900 random instances with up to 8 modes, 5 positions, 10 tokens per position and budgets
up to 40, under uniform and random priors, it returned exactly the $\B$ largest prefixes under $s$, and with $\K{=}1$
the tree of \ddtree{}'s builder, node for node.

\textbf{Max or log-sum-exp.} A prefix can instead be scored by its mixture probability (\cref{eq:mixture}),
\begin{equation}
  s_{\mathrm{lse}}(v)=\log\sum_{k=1}^{\K}\pimix_k\,Q_k(v)=\log\bar Q(v),
  \label{eq:lse}
\end{equation}
which also never increases along paths; its $\B$ best prefixes form a tree, the best one by \cref{prop:coverage} if the
target's continuation followed $\bar Q$. The two scores satisfy $s(v)\le s_{\mathrm{lse}}(v)\le s(v)+\log\K$: the max rule
credits a prefix shared by several modes through its best mode only, the log-sum-exp adds the modes up. Its tree is
not a merge of the per-mode searches, because the rank of a prefix depends on every mode, including those that rank it
low, so we evaluate it in simulation only: the simulator ranks by $\bar Q$ the distinct prefixes of the modes' per-depth
top-$\B$ lists, a candidate set that contains the max rule's tree and approximates the log-sum-exp one.
\Cref{tab:rules} extends \cref{tab:hedge} with this rule. The log-sum-exp tree is better in every setting, most for
many modes and small budgets: by 0.5--3.3 points at $\B{=}16$ and by 0.2--1.0 points at $\B\ge256$; at $\B{=}16$ the
best $\K$ moves from 4 to 8, and $\K{=}32$ keeps a gain of 3.3\% instead of none ($T{=}0$). On the three-epoch
drafters of \cref{tab:main} ($\K{=}8$, uniform prior), it would add 0.5--0.8 points at $\B{=}64$ and 0.2--0.6 at
$\B{=}256$ (Qwen3-4B, Qwen3-8B and Qwen3.5-4B). We use the max rule: its exact tree is the union of the per-mode trees and needs only each mode's top
candidates, and our timed trees use it.

\begin{table}[h]
  \centering
  \caption{Max and log-sum-exp scoring of the mixture tree: change in $\tau$ relative to single-mode \dpace{} (5-task
    means; the quarter-epoch drafters of \cref{tab:hedge}, scored with the gate's prior as there). Max: the rule of
    \cref{eq:score}, whose $T{=}0$ columns at $\B\in\{16,64,512\}$ are those of \cref{tab:hedge}; LSE: log-sum-exp
    (\cref{eq:lse}).}
  \label{tab:rules}
  \small
  \setlength{\tabcolsep}{4pt}
  \begin{tabular}{r cc cc cc cc}
    \toprule
    & \multicolumn{2}{c}{$\B{=}16$} & \multicolumn{2}{c}{$\B{=}64$} & \multicolumn{2}{c}{$\B{=}256$}
      & \multicolumn{2}{c}{$\B{=}512$} \\
    \cmidrule(lr){2-3}\cmidrule(lr){4-5}\cmidrule(lr){6-7}\cmidrule(lr){8-9}
    $\K$ & Max & LSE & Max & LSE & Max & LSE & Max & LSE \\
    \midrule
    \multicolumn{9}{l}{\emph{$T{=}0$}} \\
    4  & $+2.5$ & $+3.0$ & $+5.4$ & $+5.7$ & $+6.2$ & $+6.4$ & $+6.2$ & $+6.4$ \\
    8  & $+2.3$ & $+3.7$ & $+6.4$ & $+7.1$ & $+7.7$ & $+8.1$ & $+7.8$ & $+8.1$ \\
    16 & $+1.0$ & $+3.3$ & $+6.6$ & $+7.7$ & $+8.2$ & $+8.8$ & $+8.4$ & $+9.0$ \\
    32 & $+0.0$ & $+3.3$ & $+6.1$ & $+7.5$ & $+8.4$ & $+9.0$ & $+8.7$ & $+9.2$ \\
    \midrule
    \multicolumn{9}{l}{\emph{$T{=}1$}} \\
    4  & $+3.1$ & $+3.9$ & $+6.0$ & $+6.5$ & $+6.9$ & $+7.1$ & $+6.8$ & $+7.1$ \\
    8  & $+3.0$ & $+4.5$ & $+7.1$ & $+7.8$ & $+8.2$ & $+8.7$ & $+8.5$ & $+8.9$ \\
    16 & $+1.7$ & $+4.2$ & $+7.4$ & $+8.5$ & $+9.3$ & $+9.9$ & $+9.8$ & $+10.3$ \\
    32 & $+0.5$ & $+3.8$ & $+6.9$ & $+8.4$ & $+9.2$ & $+10.2$ & $+9.9$ & $+10.7$ \\
    \bottomrule
  \end{tabular}
\end{table}

\textbf{Committing to one mode.} A drafter can instead choose one mode before verification and spend the whole budget
on that mode's tree. For the $\K{=}4$ \mfd{} drafter of \cref{tab:hedge}, the gate picks the best mode in 38\% of the blocks of its last logged training batch (chance: 25\%; 25, 18 and 10\% at $\K{=}8$, 16 and 32, against 12.5, 6.3 and 3.1\%); its tree loses 4.9\% against \dpace{} at $\B{=}64$ ($T{=}0$; 5.6\% at $T{=}1$), while
the mixture tree gains 5.4\% (6.0\%). The output-side drafter of \cref{tab:inject}, DBLAST's category expander and prior head trained with the mixture 
likelihood, as in its DFlash+DS variant rather than with its acceptance-length objective \citep{dblast2026}, picks the best mode more often (45\%), but its
modes are used less evenly: the most frequent winner takes 51\% of the training blocks, against 28\% for \mfd{}.
Committing to its prior's choice loses 1.1\% (2.1\%) against the single-mode reference of \cref{tab:inject}, while its
mixture tree gains 2.5\% (1.8\%). Spending the budget on each block's best mode, which is known only after
verification, would gain 13.0\% (14.7\%) and 8.9\% (9.0\%). This bounds what a better prior could add; the priors
above, right for fewer than half of the blocks, lose against a single mode.

\textbf{Assignment rule.} How blocks are assigned to modes matters little for the tree. With $\K{=}4$ in the setting
of \cref{tab:hedge}, the mixture tree gains 5.4\% / 6.0\% at $\B{=}64$ ($T{=}0$ / $T{=}1$) with the relaxed
winner-take-all of \cref{sec:method} ($\varepsilon{=}0.1$), 4.9\% / 5.1\% with $\varepsilon{=}0.3$, 5.5\% / 5.2\% when
each block's loss is spread over the modes by their posterior probabilities (soft EM), and 5.5\% / 6.2\% without the
gate's loss. The gate matters only for chain decoding: its chain falls 3.1\% below the single-mode chain with
winner-take-all and 12.0\% without the gate's loss ($T{=}0$).

\textbf{Prior.} \Cref{tab:hedge,tab:rules} and the mixture trees above weight the modes by each drafter's own prior
(the gate for \mfd{}). With a uniform prior, as in the timed results, the entries of \cref{tab:rules} change by at
most 0.5 points, except that the max rule at $T{=}1$ gains up to 1.3 points, most at $\B{=}16$.

\section{Exact tree verification for Gated-DeltaNet targets}
\label{app:hybrid}

The hybrid targets interleave attention layers with Gated-DeltaNet (GDN) layers \citep{yang2025gated}. Attention
layers verify a tree with the ancestor mask of \ddtree{}. A GDN layer carries a recurrent state that every token
rewrites, so a tree flattened into one sequence would pass each node's update on to the nodes after it, siblings
included, and a rejected branch cannot be undone by truncating a cache. We therefore evaluate every node from the
committed state without writing to it, and commit the accepted path in closed form. STree first brought tree-based speculative decoding to hybrid 
state-space models \citep{wu2025stree}. Weaver verifies GDN layers in the
same way, read-only with an ancestor-masked triangular solve, and commits only the accepted path
\citep{oda2026weaver}; Twig derives a closed form for tree verification of linear-attention layers \citep{bole2026},
and TreeWY a tree-shaped WY transform that solves all nodes at once and rebuilds only the accepted state
\citep{treewy2026}. We state the construction in our notation.

\textbf{Gated delta rule.} Per head, a GDN layer keeps a state $\mathbf{S}\in\mathbb{R}^{d_v\times d_k}$. Token $t$
brings a query $\mathbf{q}_t$, a unit-norm key $\mathbf{k}_t$ and a value $\mathbf{v}_t$ (computed through a short
causal convolution over the layer's inputs), a decay $\gamma_t\in(0,1]$ and a step size $\beta_t\in(0,1)$ (not the smoothing $\beta$ of \cref{sec:method}), and
\begin{equation}
  \mathbf{S}_t=\gamma_t\,\mathbf{S}_{t-1}\big(\mathbf{I}-\beta_t\mathbf{k}_t\mathbf{k}_t^{\top}\big)
  +\beta_t\,\mathbf{v}_t\mathbf{k}_t^{\top},\qquad \mathbf{o}_t=\mathbf{S}_t\mathbf{q}_t.
  \label{eq:gdn}
\end{equation}

\textbf{Verification.} A round verifies the root and the $\B$ drafted nodes, $n=\B{+}1$ tokens ordered so that parents
precede children. Node $i$ has the parent $\mathrm{pa}(i)$ and the root path $P(i)$, made of its ancestors and itself;
the root's parent state is the committed state $\mathbf{S}_0$. Node $i$ must see the state of \cref{eq:gdn} along its
root path, $\mathbf{S}_i=\gamma_i\mathbf{S}_{\mathrm{pa}(i)}(\mathbf{I}-\beta_i\mathbf{k}_i\mathbf{k}_i^{\top})
+\beta_i\mathbf{v}_i\mathbf{k}_i^{\top}$. Let $G_i=\sum_{j\in P(i)}\log\gamma_j$ and $\Gamma_{ij}=e^{G_i-G_j}$ for
$j\in P(i)$.

\begin{lemma}[Tree form of the gated delta rule]
  \label{lem:gdntree}
  Let $\mathbf{u}_1,\dots,\mathbf{u}_n\in\mathbb{R}^{d_v}$ solve
  \begin{equation}
    \mathbf{u}_i+\beta_i\sum_{j\in P(i)\setminus\{i\}}\Gamma_{ij}\,(\mathbf{k}_i^{\top}\mathbf{k}_j)\,\mathbf{u}_j
    =\beta_i\big(\mathbf{v}_i-e^{G_i}\mathbf{S}_0\mathbf{k}_i\big),\qquad i=1,\dots,n.
    \label{eq:gdnsolve}
  \end{equation}
  The system is unit lower triangular, so its solution is unique, and for every node $i$
  \begin{equation}
    \mathbf{S}_i=e^{G_i}\mathbf{S}_0+\sum_{j\in P(i)}\Gamma_{ij}\,\mathbf{u}_j\mathbf{k}_j^{\top},\qquad
    \mathbf{o}_i=e^{G_i}\mathbf{S}_0\mathbf{q}_i+\sum_{j\in P(i)}\Gamma_{ij}\,(\mathbf{q}_i^{\top}\mathbf{k}_j)\,
    \mathbf{u}_j.
    \label{eq:gdnstate}
  \end{equation}
\end{lemma}

\textbf{Proof.} Row $i$ of \cref{eq:gdnsolve} involves only $\mathbf{u}_i$ and the $\mathbf{u}_j$ of strict
ancestors, which precede $i$, so the system is unit lower triangular. Define $\mathbf{u}_i=\beta_i(\mathbf{v}_i-
\gamma_i\mathbf{S}_{\mathrm{pa}(i)}\mathbf{k}_i)$ from the true states; then \cref{eq:gdn} reads
$\mathbf{S}_i=\gamma_i\mathbf{S}_{\mathrm{pa}(i)}+\mathbf{u}_i\mathbf{k}_i^{\top}$. Suppose the state formula of
\cref{eq:gdnstate} holds for $\mathrm{pa}(i)$, or that $i$ is the root (its parent state $\mathbf{S}_0$ has $G=0$ and
an empty sum). Since $G_i=G_{\mathrm{pa}(i)}+\log\gamma_i$ and $P(i)=P(\mathrm{pa}(i))\cup\{i\}$, multiplying it by
$\gamma_i$ and adding $\mathbf{u}_i\mathbf{k}_i^{\top}$ gives the formula for $i$, and inserting it into the definition
of $\mathbf{u}_i$ gives row $i$ of \cref{eq:gdnsolve}. By induction in topological order, these $\mathbf{u}_i$ solve
\cref{eq:gdnsolve} and are therefore its unique solution, and $\mathbf{o}_i=\mathbf{S}_i\mathbf{q}_i$ gives the output
formula.\qed

Restricted to the rows of $P(i)$, \cref{eq:gdnsolve} is the chunked form of the gated delta rule for the chain $P(i)$
\citep{yang2025gated}: the tree replaces its causal mask by the ancestor mask and sums the decays along parent
pointers. Each node thus sees exactly its own root path, siblings never interact, and no per-node state is formed: a
node keeps one vector $\mathbf{u}_i$ per head. Each node's short convolution reads its window through parent
pointers: the inputs of its nearest ancestors, completed above the root from the cached convolution state. Nothing is
written to the layer's state during verification.

\textbf{Commit.} When the target accepts the path that ends at node $a$, each GDN layer sets its state to
$\mathbf{S}_a$ by \cref{eq:gdnstate}, from the stored $\mathbf{u}_j$, $\mathbf{k}_j$ and $G_j$ of the accepted nodes:
one update of rank $|P(a)|$ per head, batched over layers, with no replay of the path. The convolution cache takes the
inputs of the last positions of $P(a)$, read through parent pointers, and attention layers keep the keys and values of
the accepted nodes, as in \ddtree{}. A chain is a tree with one path and goes through the same code, so on the hybrid
targets every method, chains included, verifies with these kernels; the autoregressive baseline keeps the model's
native path.

\textbf{Cost.} Per layer and head, the kernels form the masked products $\Gamma_{ij}\mathbf{k}_i^{\top}\mathbf{k}_j$
and $\Gamma_{ij}\mathbf{q}_i^{\top}\mathbf{k}_j$ of ancestor pairs, invert the diagonal blocks of \cref{eq:gdnsolve}
in blocks of 64 nodes, and sweep the blocks in order. This takes $O(n^2(d_k{+}d_v)+nd_kd_v)$ operations and
$O(n^2+nd_v)$ memory, instead of the $O(nd_kd_v)$ memory of per-node states; the commit takes
$O(|P(a)|\,d_kd_v)$. In the harness of \cref{tab:main} on Qwen3.5-4B at $T{=}0$, the verification pass of a round
takes 26.1\,ms for a 16-token chain and 26.1, 26.2 and 26.7\,ms for trees of 16, 64 and 256 nodes, against 24.5\,ms
for an autoregressive step, and the step after it, which samples the target's tokens, walks the tree, commits the
states and compacts the attention cache, takes 0.9\,ms for a chain and 1.2\,ms for a 256-node tree. The launch-bound
harness hides most of the tree's compute: summed over its kernels, the verification of a 256-node tree takes
17.6\,ms against 12.2\,ms for a chain (2K-token context).

\textbf{Exactness.} \Cref{lem:gdntree} is exact in real arithmetic. In float64, on random trees of 40 nodes, the
outputs of \cref{eq:gdnstate} and the committed state match node-by-node recurrence to within $4\times10^{-15}$. The
fused bf16 kernels match a float32 evaluation of the same equations to a relative error of $2$--$4\times10^{-4}$ in
$\mathbf{u}$ and about $2\times10^{-3}$ in the outputs. On Qwen3.5-4B, after a 600-token context, the logits of every
node of random 48-node trees differ from those of the native forward pass over the node's root path by at most 0.31,
about as much as two native implementations of the layer differ from each other, and the step after a commit agrees
with the native path. End to end (MATH-500, thinking mode, $T{=}0$), greedy decoding through these kernels reproduces
autoregressive decoding token for token until a near-tie: wherever the two outputs split, the target's logits for the
two tokens differ by at most 0.25, one or two bf16 rounding steps, and the native implementations split at such ties
as well. Verification is therefore lossless up to bf16 ties, as for every tree method on a bf16 target.

\section{Per-task results}
\label{app:pertask}
\input{sections/pertask_tables}

\Cref{tab:hybrid-budget} adds the hybrid targets at every budget of their main runs. At $\B{=}16$, \mfd{} trails
\dpace{} by 1.7--3.7\% at $T{=}0$ and leads it by 1--2\% at $T{=}1$; from $\B{=}64$ on, it leads by 3--9\%.

\begin{table}[h]
  \centering
  \caption{Hybrid targets by tree budget: speedup over autoregressive decoding and $\tau$ (in parentheses), 5-task
  means, from the main runs of \cref{tab:main} (\ddtree{} with the released drafter; \dpace{} and \mfd{} fine-tuned
  for three epochs).}
  \label{tab:hybrid-budget}
  \small
  \setlength{\tabcolsep}{4pt}
  \begin{tabular}{r l cc cc}
    \toprule
    & & \multicolumn{2}{c}{Qwen3.5-4B (thinking)} & \multicolumn{2}{c}{Qwen3.6-27B (thinking)} \\
    \cmidrule(lr){3-4}\cmidrule(lr){5-6}
    $\B$ & Method & $T{=}0$ & $T{=}1$ & $T{=}0$ & $T{=}1$ \\
    \midrule
    16  & \ddtree{}         & 5.38$\times$ (7.04) & 3.88$\times$ (5.06) & 5.28$\times$ (6.61) & 4.72$\times$ (5.92) \\
        & \dpace{}+\ddtree{} & 5.91$\times$ (7.74) & 3.90$\times$ (5.08) & 5.88$\times$ (7.35) & 5.00$\times$ (6.27) \\
        & \mfd{}            & 5.69$\times$ (7.63) & 3.97$\times$ (5.30) & 5.78$\times$ (7.36) & 5.06$\times$ (6.47) \\
    \midrule
    64  & \ddtree{}         & 6.18$\times$ (8.14) & 4.47$\times$ (5.86) & 5.89$\times$ (7.62) & 5.27$\times$ (6.86) \\
        & \dpace{}+\ddtree{} & 6.65$\times$ (8.75) & 4.42$\times$ (5.79) & 6.46$\times$ (8.35) & 5.50$\times$ (7.17) \\
        & \mfd{}            & 6.85$\times$ (9.23) & 4.77$\times$ (6.40) & 6.77$\times$ (8.91) & 5.94$\times$ (7.87) \\
    \midrule
    256 & \ddtree{}         & 6.53$\times$ (8.78) & 4.77$\times$ (6.51) & 4.48$\times$ (8.33) & 4.00$\times$ (7.53) \\
        & \dpace{}+\ddtree{} & 6.97$\times$ (9.38) & 4.66$\times$ (6.35) & 4.82$\times$ (8.97) & 4.13$\times$ (7.79) \\
        & \mfd{}            & 7.31$\times$ (10.07) & 5.04$\times$ (7.03) & 5.16$\times$ (9.72) & 4.52$\times$ (8.62) \\
    \bottomrule
  \end{tabular}
\end{table}

\section{Mode analyses}
\label{app:modes}

The analyses of \cref{sec:alternatives} run each drafter teacher-forced on the target's responses in the
simulations (128 prompts per task, every anchor; the target's greedy continuation at $T{=}0$ and the sampled response
at $T{=}1$) and score every decoding on the same blocks. At $T{=}1$ (\cref{tab:alt_t1}) the ordering of
\cref{tab:alt} holds: the gate's mode loses about 8\% against \dpace{}'s tree, the modes' chains stay below that tree, the
per-mode trees with 128 nodes stay below the mixture tree with 64, the random table keeps 53--73\% of the gain, the product of the mean marginals matches \dpace{}'s tree, with the mixture tree 10--11\% above it, with each position's modes shuffled the mixture tree falls 5--6\% below \dpace{}'s (7--8\% at $T{=}0$), and the product of the per-position maxima falls 8--9\% below \dpace{}'s at both temperatures. Both fall below \dpace{}'s tree on every task, at 64, 128 and 256 nodes and at both temperatures, the shuffled modes also when ranked by log-sum-exp; since each position keeps its eight distributions, the modes help only through their coupling across positions.

\begin{table}[h]
  \centering
  \caption{Ways to decode the same drafters: simulated $\tau$ ($T{=}0$, 5-task means over 128 prompts per
  task, 80 for MT-Bench) of the three-epoch drafters of \cref{tab:main} (uniform prior), with the nodes verified per round. Chains are verified side by
  side; best-first: the argmax chain and the seven single-token deviations with the largest expected gain. Per-mode
  trees: one tree of $\B/8$ nodes per mode. Product of mean marginals: the \ddtree{} tree of the average of the
  eight modes' per-position distributions, the same drafter without coupling. Product of per-position maxima: the \ddtree{} tree of $\prod_i\max_k\qk{k,i}(v_i)$, each position scored by its best mode. Modes shuffled: the mixture tree after permuting the eight modes' distributions independently at each position, which keeps each position's distributions but not their coupling (mean of two seeds, which differ by at most 0.02). Mean / random table: every mode embedding replaced by the table's mean, or
  by the mean plus a random direction with that mode's norm about it. $T{=}1$ in \cref{tab:alt_t1}.}
  \label{tab:alt}
  \small
  \setlength{\tabcolsep}{1.8pt}
  \begin{tabular}{l c cc}
    \toprule
    Decoding & Nodes & Qwen3-4B & Qwen3-8B \\
    \midrule
    \multicolumn{4}{l}{\emph{One mode: the \dpace{} drafter}} \\
    Chain & 15 & 6.18 & 6.26 \\
    8 sampled chains & $\le$120 & 7.05 & 7.14 \\
    8 chains, best-first & $\le$120 & 7.63 & 7.70 \\
    Tree & 64 / 128 & 8.06 / 8.42 & 8.14 / 8.49 \\
    \midrule
    \multicolumn{4}{l}{\emph{Eight modes: the \mfd{} drafter}} \\
    The gate's mode, chain & 15 & 6.01 & 6.03 \\
    The gate's mode, tree & 64 / 128 & 7.52 / 7.84 & 7.54 / 7.85 \\
    The 8 modes' chains & $\le$120 & 8.05 & 8.14 \\
    8 per-mode trees & $\le$64 / $\le$128 & 6.57 / 8.69 & 6.60 / 8.75 \\
    Product of mean marginals & 64 / 128 & 7.90 / 8.29 & 7.91 / 8.33 \\
    Product of per-position maxima & 64 / 128 & 7.37 / 7.68 & 7.37 / 7.70 \\
    Mixture tree, modes shuffled & 64 / 128 & 7.48 / 7.85 & 7.48 / 7.86 \\
    Mixture tree, mean table & 64 / 128 & 7.90 / 8.26 & 7.98 / 8.33 \\
    Mixture tree, random table & 64 / 128 & 8.63 / 9.04 & 8.57 / 8.96 \\
    Mixture tree (\mfd{}) & 64 / 128 & \textbf{8.77} / \textbf{9.20} & \textbf{8.81} / \textbf{9.27} \\
    \bottomrule
  \end{tabular}
\end{table}

\begin{table}[h]
  \centering
  \caption{\Cref{tab:alt} at $T{=}1$.}
  \label{tab:alt_t1}
  \small
  \setlength{\tabcolsep}{2pt}
  \begin{tabular}{l c cc}
    \toprule
    Decoding & Nodes & Qwen3-4B & Qwen3-8B \\
    \midrule
    \multicolumn{4}{l}{\emph{One mode: the \dpace{} drafter}} \\
    Chain & 15 & 5.23 & 5.10 \\
    8 sampled chains & $\le$120 & 6.02 & 5.87 \\
    8 chains, best-first & $\le$120 & 6.55 & 6.38 \\
    Tree & 64 / 128 & 6.97 / 7.30 & 6.77 / 7.10 \\
    \midrule
    \multicolumn{4}{l}{\emph{Eight modes: the \mfd{} drafter}} \\
    The gate's mode, chain & 15 & 5.10 & 4.98 \\
    The gate's mode, tree & 64 / 128 & 6.44 / 6.74 & 6.22 / 6.50 \\
    The 8 modes' chains & $\le$120 & 6.94 & 6.75 \\
    8 per-mode trees & $\le$64 / $\le$128 & 6.00 / 7.64 & 5.89 / 7.44 \\
    Product of mean marginals & 64 / 128 & 6.93 / 7.31 & 6.77 / 7.15 \\
    Product of per-position maxima & 64 / 128 & 6.40 / 6.70 & 6.19 / 6.48 \\
    Mixture tree, modes shuffled & 64 / 128 & 6.56 / 6.90 & 6.41 / 6.74 \\
    Mixture tree, mean table & 64 / 128 & 6.82 / 7.14 & 6.63 / 6.95 \\
    Mixture tree, random table & 64 / 128 & 7.48 / 7.85 & 7.16 / 7.51 \\
    Mixture tree (\mfd{}) & 64 / 128 & \textbf{7.67} / \textbf{8.07} & \textbf{7.46} / \textbf{7.87} \\
    \bottomrule
  \end{tabular}
\end{table}

\begin{figure}[h]
  \centering
  \includegraphics[width=0.55\linewidth]{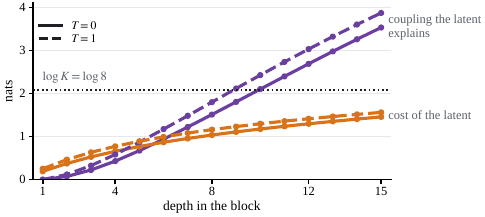}
  \caption{Coupling the latent explains on the target's continuations, and its cost against the best mode, by depth in
    the block (Qwen3-4B, three-epoch \mfd{} drafter, $\K{=}8$, 5 tasks). The explained coupling grows with depth and
    passes $\log\K$, the largest possible cost, at depth 9--10.}
  \label{fig:coupling}
\end{figure}

\textbf{Coupling.} For a block whose first $d$ gold tokens are $y_{1:d}$, the latent explains
$\log \tfrac{1}{\K}\sum_k \prod_{i\le d} \qk{k,i}(y_i) - \sum_{i\le d} \log \tfrac{1}{\K}\sum_k \qk{k,i}(y_i)$ nats: the
uniform mixture against the product of its own per-position marginals. It costs $\max_k \sum_{i\le d}\log \qk{k,i}(y_i)$
minus the mixture's log-likelihood, at most $\log\K$. \Cref{fig:coupling} averages both over the blocks of the five
tasks; the Qwen3-8B drafter is within 0.03 nats of the Qwen3-4B one at every depth (0.04 at $T{=}1$).

\textbf{Joint likelihood of the target's blocks.} \Cref{fig:coupling} compares the mixture with its own marginals.
Against \dpace{} (FT 1 ep, the strongest single-mode drafter), the uniform mixture's log-likelihood of the gold block,
$\log\frac1\K\sum_kQ_k(y_{1:d})$ against $\sum_{i\le d}\log q_i(y_i)$, averaged over all anchors whose block lies
within the response, is no higher at depth 1 ($-0.28$ against $-0.27$ nats on Qwen3-4B, $T{=}0$) and higher by 0.6
nats at depth 4, 2.5 at depth 8 and 6.7 at depth 15 (6.6--8.9 nats at depth 15 on both targets and temperatures).
About half of the gap at depth 15 is the coupling of \cref{fig:coupling} (3.5 nats); the rest is the mixture's better
deep marginals. The gap lies mostly where the target forks (\cref{app:chimeras}): 8.9--11.5 nats at depth 15 there,
against 1.9--3.1 where the target is confident. A higher likelihood does not by itself make a better tree: the product
of the mixture's own marginals scores higher than \dpace{} here but builds a worse tree (\cref{tab:alt}).

\textbf{Mode use and geometry.} The effective number of modes is 7.9 / 8.0 (Qwen3-4B / Qwen3-8B, $T{=}0$) for the
likeliest mode of each gold block, 7.9 / 8.0 for the mode that scores the accepted path of the $\B{=}64$ mixture tree
best, and 7.3 / 7.7 for the gate's top mode; no mode is the likeliest for more than 16\% of the blocks. Leaving any one mode out of the mixture tree lowers $\tau$
by 0.3--1.2\% at $\B{=}64$ and 256 (0.4--1.4\% at $T{=}1$). The embeddings share an offset
(norm 0.27, against 0.41--0.45 for each mode on Qwen3-4B), and about their mean they are nearly equiangular: pairwise cosines lie between $-0.20$ and $-0.09$ on Qwen3-4B and
between $-0.35$ and $-0.04$ on Qwen3-8B, around the $-1/7$ of a regular simplex, which centered vectors of equal norm
average to. Trained from
scratch on the full data, the Qwen3-4B modes spread less evenly ($-0.47$ to $0.22$).

\textbf{Predicting the winning mode.} Call the winning mode of a gold block the mode with the largest coverage
$C_k(y)$, the target the gate is trained on. The gate's top 1, 2 and 4 modes contain it in 28, 47 and 72\% of the
blocks on Qwen3-4B ($T{=}0$; 24--28, 41--47 and 66--72\% on both targets and temperatures), against 12.5, 25 and 50\%
by chance and 17, 33 and 58\% for a fixed ranking by how often each mode wins. Several modes often fit a block almost
equally well (2.8--2.9 modes within 5\% of the best coverage on average), but this does not explain the low accuracy:
in the 47--49\% of blocks with a single winner, the gate's top mode is the winner in 22--27\%, and counting any mode
within 5\% of the best as a hit raises its top-1 accuracy only to 48--51\% against 35--36\% by chance. The gate's
mode keeps 84--85\% of the best mode's coverage, a random mode 76--77\%.

\textbf{Forks.} The modes' argmax chains disagree somewhere in 88\% of the blocks at $T{=}0$ (91\% at $T{=}1$). The
first disagreement falls at the first drafted position in 29\% of these blocks, at the second in 16\%, and less often deeper. \Cref{tab:forks} lists, for each mode, the tokens it drafts at that position most often relative to the
average mode.

\begin{table}[h]
  \centering
  \caption{Tokens each mode drafts at the first position where the modes' argmax chains disagree, most often relative
  to the average mode (lift in parentheses; Qwen3-4B, $T{=}0$; tokens with at least 300 forks over all modes, variants that differ only in surrounding whitespace merged, byte fragments omitted).}
  \label{tab:forks}
  \small
  \setlength{\tabcolsep}{4pt}
  \begin{tabular}{c l}
    \toprule
    Mode & Distinctive fork tokens \\
    \midrule
    1 & Our (3.9), But (3.7), implementation (3.2), To (2.6), However (2.5) \\
    2 & \texttt{Rightarrow} (2.8), \texttt{**,} (2.6), \texttt{\$,} (2.5), \texttt{\textasciigrave,} (2.3), Thus (2.2) \\
    3 & This (3.4), These (3.1), which (3.1), Key (2.6), Since (2.6) \\
    4 & Therefore (5.0), also (3.0), Test (2.7), correctly (2.6), Then (2.6) \\
    5 & without (2.8), We (2.8), may (2.7), They (2.6), we (2.5) \\
    6 & \texttt{Analy} (2.5), \texttt{-{}-{}-} (2.2), in (2.1), The (2.1), denote (2.1) \\
    7 & passes (4.2), Solve (2.4), end of turn (1.9), Find (1.9), \texttt{\#\#\#\#} (1.8) \\
    8 & \texttt{\$.} (3.1), So (2.7), indentation (2.5), let (2.5), abc (2.3) \\
    \bottomrule
  \end{tabular}
\end{table}

\textbf{Diverging futures or local swaps.} \Cref{tab:disagree} asks how the argmax chains of the injection variants of
\cref{sec:inject} differ once they part. All four part at about the same position; after the split, the input-injected
modes still differ at 61\% of the remaining positions and the last-layer and output-side modes at 43--46\%. The former
follow different continuations, which a product tree can hold only with exponentially many chimeras
(\cref{lem:chimera}); the latter swap a token or two and agree again, which a single-mode tree covers cheaply. The order
matches the $\tau$ gains of \cref{tab:inject}.

\begin{table}[h]
  \centering
  \caption{Where the modes' argmax chains part (Qwen3-4B; the $\K{=}4$ quarter-epoch drafters of \cref{tab:inject}; 8
  held-out conversations per task, every fourth anchor, 4330 anchors, all six pairs of modes, 15-token argmax chains).
  Identical: share of pairs whose chains agree at every position. First split: mean position of the first differing
  token among the other pairs (1 is the first drafted token). After the split: mean number of differing positions from
  there on, and their share of the remaining positions. Gain: the mixture tree's change in $\tau$ at $\B{=}64$ relative
  to \dpace{} (\%, \cref{tab:inject}).}
  \label{tab:disagree}
  \small
  \setlength{\tabcolsep}{5pt}
  \begin{tabular}{l ccccr}
    \toprule
    Injection & Identical & First split & \multicolumn{2}{c}{After the split} & Gain \\
    \cmidrule(lr){4-5}
    & & & positions & share & \\
    \midrule
    Input (layer 0)  & 0.23 & 5.5 & 6.2 & 0.61 & $+5.0$ \\
    Layer 2          & 0.23 & 5.6 & 5.8 & 0.59 & $+4.1$ \\
    Layer 4 (last)   & 0.25 & 5.8 & 4.3 & 0.46 & $+1.5$ \\
    Output, rank 128 & 0.28 & 5.7 & 4.1 & 0.43 & $-0.6$ \\
    \bottomrule
  \end{tabular}
\end{table}

\textbf{Offsets on a single-mode drafter.} To separate what the offsets do from what training does,
\cref{tab:ctrl} loads the \dpace{} drafter, fine-tuned identically but without modes, into the eight-mode
architecture and offsets its eight copies by \mfd{}'s trained table, by the random table of \cref{sec:alternatives}
(the table's mean plus random directions with each mode's norm about it), or by the random directions alone at one,
two and four times that norm. With zero offsets the copies are identical and the tree is \dpace{}'s own, which the first row reproduces up to 
0.01, the difference between separate simulation runs (\cref{tab:alt_t1,tab:epochs}). The offsets make the copies disagree: with \mfd{}'s table their argmax chains differ in 76--81\%
of the blocks, against 88--91\% for \mfd{}. Few of these disagreements are futures of the target. \mfd{}'s table gains
2.7--3.7\% over \dpace{}'s tree and random directions 1.0--2.5\%, while four times the norm loses up to 15\%; the same
random table gains three to five times as much on the \mfd{} drafter. \mfd{}'s trained table transferred to \dpace{}
recovers about a third of \mfd{}'s gain (26--39\%), more than random directions (10--27\%), so the table carries some
information of its own; the rest of the gain is in the drafter's weights.

\begin{table}[h]
  \centering
  \caption{Mixture trees over eight offset copies of the \dpace{} drafter, which was not trained with modes, against
  the \mfd{} drafter (simulated $\tau$ at $\B{=}64$ / 256, 5-task means, three-epoch drafters of \cref{tab:main},
  uniform prior). Random table: the trained table's mean plus random directions with each mode's norm about it, as in
  \cref{tab:alt}; random directions: the same directions without the mean, at 1, 2 and 4 times that norm.}
  \label{tab:ctrl}
  \small
  \setlength{\tabcolsep}{4pt}
  \begin{tabular}{l cc cc}
    \toprule
    & \multicolumn{2}{c}{Qwen3-4B} & \multicolumn{2}{c}{Qwen3-8B} \\
    \cmidrule(lr){2-3}\cmidrule(lr){4-5}
    Drafter and offsets & $T{=}0$ & $T{=}1$ & $T{=}0$ & $T{=}1$ \\
    \midrule
    \dpace{}, identical copies (its \ddtree{}) & 8.06 / 8.71 & 6.96 / 7.60 & 8.14 / 8.78 & 6.77 / 7.38 \\
    \dpace{} + \mfd{}'s trained table & 8.30 / 9.03 & 7.18 / 7.87 & 8.37 / 9.05 & 6.95 / 7.62 \\
    \dpace{} + \mfd{}'s random table & 8.21 / 8.91 & 7.08 / 7.75 & 8.24 / 8.90 & 6.85 / 7.49 \\
    \dpace{} + random directions & 8.24 / 8.93 & 7.12 / 7.79 & 8.22 / 8.89 & 6.84 / 7.47 \\
    \dpace{} + random directions $\times 2$ & 8.06 / 8.80 & 6.91 / 7.65 & 8.19 / 8.89 & 6.80 / 7.44 \\
    \dpace{} + random directions $\times 4$ & 6.96 / 7.81 & 5.94 / 6.72 & 7.83 / 8.59 & 6.41 / 7.12 \\
    \midrule
    \mfd{} + random table & 8.63 / 9.38 & 7.48 / 8.19 & 8.57 / 9.30 & 7.16 / 7.82 \\
    \mfd{} (trained table) & \textbf{8.77} / \textbf{9.54} & \textbf{7.67} / \textbf{8.43} & \textbf{8.81} / \textbf{9.61}
      & \textbf{7.46} / \textbf{8.23} \\
    \bottomrule
  \end{tabular}
\end{table}

\FloatBarrier
\section{The number of modes in wall-clock time}
\label{app:kspeed}

\Cref{tab:kspeed} times the quarter-epoch drafters of \cref{tab:hedge} and one with $\K{=}2$ ($\K\in\{2,4,8,16,32\}$, with the same number
of draft blocks per training step, and the single-mode \dpace{} drafter, with 512 instead of 1024 draft blocks per sequence, as $\K{=}1$) in the harness
of \cref{tab:main} on Qwen3-4B, as mixture trees with the max rule. Two reference drafters in the same run, \dpace{} and \mfd{}
fine-tuned for one epoch, reproduce the $T{=}0$ $\tau$ of \cref{tab:main} exactly and its speedups within 0.6\%. At $T{=}0$, the measured $\tau$ gains over $\K{=}1$ match or exceed those of the simulations (\cref{tab:rules}), except at $\K{=}32$ and $\B{=}16$ ($-1.0\%$ against $0.0$) and at $\K{=}4$ and $\B{=}256$ (6.1\% against 6.2\%); at $T{=}1$ they fall 0.2--0.9 points short in three of the 16 cells ($\K{=}4$ at $\B{=}16$ and 256, $\K{=}16$ at $\B{=}64$). Drafting the modes
costs 0.5--0.6\,ms more per round than a single mode up to $\K{=}8$, 0.7\,ms at $\K{=}16$ and 1.1\,ms at $\K{=}32$,
whose tree building also takes 0.65\,ms longer; at $\B{=}256$ the draft cost takes about 2 points of the $\tau$ gain
for $\K\le8$, 3 for $\K{=}16$ and 6.5 of 8.6 for $\K{=}32$. At $\B{=}16$ no $\K$ gains more than 1.6\%, and $\B{=}512$ is slower
than $\B{=}256$ for every $\K$ (verification grows from 20.5 to 30.7\,ms per round), so the larger best $\K$ that the
simulations find at large budgets does not turn into speed.

\begin{table}[h]
  \centering
  \caption{Speedup and $\tau$ (in parentheses) of the quarter-epoch drafters with $\K$ modes on Qwen3-4B (5-task
    means, the harness of \cref{tab:main}); $\K{=}1$ is the single-mode \dpace{} drafter
    (512 draft blocks per sequence). Draft and tree: drafting and tree-building time per round at $\B{=}256$ (ms; drafting time
    hardly changes with $\B$). Bold: fastest at each temperature.}
  \label{tab:kspeed}
  \small
  \setlength{\tabcolsep}{4pt}
  \begin{tabular}{r cccc cc}
    \toprule
    $\K$ & $\B{=}16$ & $\B{=}64$ & $\B{=}256$ & $\B{=}512$ & Draft & Tree \\
    \midrule
    \multicolumn{7}{l}{\emph{$T{=}0$}} \\
    1 & 5.08$\times$ (7.10) & 5.80$\times$ (8.08) & 6.23$\times$ (8.77) & 4.63$\times$ (9.01) & 3.70 & 0.37 \\
    2 & 5.04$\times$ (7.19) & 5.91$\times$ (8.38) & 6.33$\times$ (9.08) & 4.76$\times$ (9.39) & 4.22 & 0.34 \\
    4 & 5.11$\times$ (7.29) & 6.05$\times$ (8.58) & 6.49$\times$ (9.31) & 4.90$\times$ (9.66) & 4.24 & 0.36 \\
    8 & 5.10$\times$ (7.29) & 6.11$\times$ (8.70) & \textbf{6.60$\times$} (9.49) & 4.96$\times$ (9.81) & 4.28 & 0.44 \\
    16 & 5.02$\times$ (7.22) & 6.07$\times$ (8.72) & \textbf{6.60$\times$} (9.55) & 5.00$\times$ (9.97) & 4.38 & 0.58 \\
    32 & 4.74$\times$ (7.03) & 5.81$\times$ (8.65) & 6.36$\times$ (9.52) & 4.85$\times$ (9.89) & 4.81 & 1.02 \\
    \midrule
    \multicolumn{7}{l}{\emph{$T{=}1$}} \\
    1 & 4.45$\times$ (6.19) & 5.12$\times$ (7.11) & 5.45$\times$ (7.71) & 4.00$\times$ (8.01) & 3.70 & 0.37 \\
    2 & 4.49$\times$ (6.37) & 5.26$\times$ (7.45) & 5.64$\times$ (8.11) & 4.12$\times$ (8.34) & 4.23 & 0.34 \\
    4 & 4.48$\times$ (6.37) & 5.35$\times$ (7.58) & 5.70$\times$ (8.22) & 4.27$\times$ (8.65) & 4.24 & 0.36 \\
    8 & 4.52$\times$ (6.45) & 5.37$\times$ (7.62) & \textbf{5.84$\times$} (8.43) & 4.32$\times$ (8.79) & 4.29 & 0.43 \\
    16 & 4.44$\times$ (6.38) & 5.28$\times$ (7.57) & 5.82$\times$ (8.47) & 4.32$\times$ (8.85) & 4.39 & 0.58 \\
    32 & 4.24$\times$ (6.27) & 5.17$\times$ (7.66) & 5.64$\times$ (8.48) & 4.25$\times$ (8.90) & 4.82 & 1.01 \\
    \bottomrule
  \end{tabular}
\end{table}

\FloatBarrier
\section{Chimeras in product trees}
\label{app:chimeras}

We count chimeras directly with the protocol of \cref{fig:budget} ($T{=}0$ contexts, the target's $T{=}1$
conditionals from the verification pass, at most 1024 new tokens, first turns) on all 628 prompts of the five tasks;
on the first 24 prompts per task, the set of \cref{fig:budget}, the counts reproduce that figure exactly. A chimera is a prefix that splices tokens of different
futures (\cref{sec:theory}); to detect them at scale, we count a node $v=(v_1,\dots,v_d)$ with $d\ge2$ as a
\emph{chimera} if the target gives its last step probability below 1\% after its
own prefix but at least 10\% after another prefix of depth $d{-}1$ in the same tree: the step belongs to a future that
another branch holds. Other steps below 1\% are \emph{dead}, the rest \emph{live}. The target's continuation
\emph{forks} at an anchor if, along its greedy path within the block, a runner-up token has probability at least
10\% and the greedy continuation from it differs from that path in at least two of the remaining positions.

\begin{table}[h]
  \centering
  \caption{Chimeras in product and mixture trees (628 prompts, $T{=}0$ contexts, $T{=}1$ target probabilities): share
    of a tree's nodes that are chimeras, dead or live, chimeras per tree, and share of rounds whose best path passes
    through a chimera. FT: fine-tuned under our recipe.}
  \label{tab:chimeras}
  \small
  \setlength{\tabcolsep}{3pt}
  \begin{tabular}{l c ccc c c}
    \toprule
    Drafter & $\B$ & Chimera & Dead & Live & Chimeras & Best path \\
     & & (\%) & (\%) & (\%) & per tree & (\%) \\
    \midrule
    \multicolumn{7}{l}{\emph{Qwen3-4B}} \\
    \ddtree{} (released) & 64 & 30.7 & 22.1 & 47.2 & 19.6 & 64.6 \\
    \dpace{} FT 1 ep & 64 & 29.0 & 20.9 & 50.0 & 18.6 & 63.0 \\
    \dpace{} FT 3 ep & 64 & 27.6 & 22.2 & 50.2 & 17.7 & 64.6 \\
    \mfd{} FT 1 ep & 64 & 23.8 & 19.4 & 56.7 & 15.2 & 44.6 \\
    \mfd{} FT 3 ep & 64 & 22.9 & 19.6 & 57.5 & 14.7 & 45.2 \\
    \ddtree{} (released) & 256 & 41.7 & 19.5 & 38.8 & 106.9 & 76.8 \\
    \dpace{} FT 1 ep & 256 & 39.9 & 18.4 & 41.7 & 102.3 & 73.7 \\
    \dpace{} FT 3 ep & 256 & 38.7 & 19.6 & 41.7 & 99.0 & 75.1 \\
    \mfd{} FT 1 ep & 256 & 37.8 & 17.0 & 45.2 & 96.7 & 64.1 \\
    \mfd{} FT 3 ep & 256 & 36.9 & 17.1 & 46.0 & 94.4 & 64.1 \\
    \midrule
    \multicolumn{7}{l}{\emph{Qwen3-8B}} \\
    \ddtree{} (released) & 64 & 32.5 & 20.1 & 47.4 & 20.8 & 61.6 \\
    \dpace{} FT 1 ep & 64 & 28.3 & 19.0 & 52.7 & 18.1 & 61.4 \\
    \dpace{} FT 3 ep & 64 & 26.8 & 20.4 & 52.7 & 17.2 & 63.0 \\
    \mfd{} FT 1 ep & 64 & 23.0 & 17.6 & 59.3 & 14.7 & 42.7 \\
    \mfd{} FT 3 ep & 64 & 22.1 & 17.7 & 60.2 & 14.1 & 42.4 \\
    \ddtree{} (released) & 128 & 38.7 & 19.3 & 42.0 & 49.5 & 69.1 \\
    \dpace{} FT 1 ep & 128 & 34.1 & 18.2 & 47.6 & 43.7 & 68.3 \\
    \dpace{} FT 3 ep & 128 & 32.6 & 19.5 & 47.9 & 41.7 & 69.8 \\
    \mfd{} FT 1 ep & 128 & 30.4 & 16.7 & 52.9 & 38.9 & 54.1 \\
    \mfd{} FT 3 ep & 128 & 29.4 & 17.0 & 53.7 & 37.6 & 53.9 \\
    \bottomrule
  \end{tabular}
\end{table}

Chimeras are the common case (\cref{tab:chimeras}). At $\B{=}64$ they are 27--33\% of a product tree's nodes, 17--21
per tree; at least one appears in 99--100\% of rounds, and the best path passes through one in 61--65\% of rounds. They
are most frequent at depths 3--4 and grow with the budget (39--42\% at $\B{=}256$ on Qwen3-4B). In the budget analysis of \cref{sec:budget}, the \mfd{} and \dpace{} trees hold
nearly the same number of nodes at every depth: per round, they differ by at most 0.9 nodes at $\B{=}64$ and 1.4 at
$\B{=}256$. The target forks at
64\% of anchors on Qwen3-4B and 68\% on Qwen3-8B, and after a fork its two continuations differ in 9.6--9.8 of the
10.3--10.6 remaining positions, the regime in which \cref{lem:chimera} makes chimeras exponential. The mixture tree
holds fewer (22--24\% of its nodes at $\B{=}64$), and its best path passes through one in 42--45\% of rounds. The
difference lies almost entirely in the blocks where the target forks, 74--78\% of decoding rounds (against 64--68\% of anchors with a full block): 20.5 against 15.5 chimeras per tree on Qwen3-4B
and 20.0 against 14.9 on Qwen3-8B (\dpace{} FT 1 ep against \mfd{} FT 3 ep, $\B{=}64$), against 12.6 and 11.9, and 11.0 and 11.0, where it does not. Counted above each tree's level where the target does
not fork, forks add 7.9 and 9.0 chimeras per tree to \dpace{}'s tree and 3.6 and 3.9 to the mixture tree, which
thus removes 54--57\% of them. The ordering is the same with a 5\% threshold for the other prefix, when only live
prefixes may support a chimera (16.0\% against 12.7\% of nodes on Qwen3-4B), and when any prefix off the node's path may (36.6\% against 30.1\%). The share of live nodes, which unlike the chimera count does not depend on the tree's other prefixes, separates the trees as well: 57--60\% of the mixture tree's nodes are live at $\B{=}64$ against 50--53\% of \dpace{}'s, and 45--46\% against 42\% at $\B{=}256$ on Qwen3-4B. In the example of \cref{fig:teaser}, the product tree has three chimeras by this
definition and two more dead steps that append future A's fifth token, which no prefix of depth 4 supports because the
tree holds A only to depth 2; nine of its 15 nodes lie on spliced prefixes, and the mixture tree has none.

\textbf{Where the gain comes from.} On the simulation prompts of \cref{tab:alt}, we evaluate \mfd{} (FT 3 ep) and \dpace{} (FT 1
ep) at every anchor of the same continuations and count the tokens each tree accepts there; at $T{=}1$ the fork test rolls the target out
greedily from the sampled tokens. Anchors where the target forks are 60--71\% of those whose block lies within the response and carry 75--88\% of \mfd{}'s gain: \mfd{} accepts 9.7--12.1\% more there and 2.4--3.8\% more elsewhere at $\B{=}64$ (10.6--12.8\% and 2.6--4.1\% at $\B{=}256$), and more at forks than elsewhere on every task and with fork thresholds of 5--30\% and two or four differing positions. Forks are also
where single-mode trees stop early: \dpace{} accepts 7.4 tokens per anchor there against 13.2 elsewhere (Qwen3-4B,
$T{=}0$, $\B{=}64$). Compared at equal difficulty, binning anchors by the two drafters' mean acceptance in half-token bins and
reweighting the other anchors to the forks' distribution over the bins, \mfd{} gains
0.8 tokens per anchor at forks, 0.8--1.0 where the target is confident and 0.5--0.6 where its alternatives rejoin:
the gain follows the anchors where a single mode's tree stops early, and most of these are forks.

\FloatBarrier
\section{Training and evaluation details}
\label{app:details}

\textbf{Training.} The fine-tuned drafters start from the released \dflash{} checkpoints, which have five layers of the
target's width (six on Qwen3.5-4B) and read five of its layers (eight on Qwen3.5-4B), and are trained with AdamW
without weight decay: learning rate $10^{-4}$ with linear warmup over the first 4\% of steps and cosine decay to zero
over the whole run, gradient clipping at norm 1, sequences of at most 3072 tokens, and 12 sequences per step on the
100K data (36 on the full data). The one-epoch drafters are separate runs whose schedule ends after one epoch, except
on Qwen3.6-27B (\cref{tab:main}). The fine-tuned drafters of \cref{tab:main} use the \dpace{} loss (smoothing 0.5) on
512 anchor blocks per sequence. \mfd{} drafts $\K{=}8$ modes on 64 anchors, weights each mode's loss by
$1{-}\varepsilon$ if it is the block's winner and by $\varepsilon/(\K{-}1)$ otherwise ($\varepsilon{=}0.1$), and adds
the gate's cross-entropy with weight 0.1; its mode embeddings start i.i.d.\ Gaussian with 0.3 times the standard
deviation of the target's token embeddings, and the gate at zero, a uniform prior. \mfd{}-D adds the drafter without a
mode embedding as a default path, trained with the single-mode \dpace{} loss on 512 further anchors per sequence with
weight 1.
\dflash{}~2 adds its selector's cross-entropy with weight 1 to the \dpace{} loss. The from-scratch runs of
\cref{sec:epochs} start both drafters from one random initialization and train for six epochs at a learning rate of
$6\times10^{-4}$, otherwise with the same recipe. A further \mfd{} drafter, used only for the embedding geometry of
\cref{app:modes}, is trained from scratch on the full data of Qwen3-4B for three epochs at the same learning rate (24
sequences per step).

\textbf{Training longer.} \Cref{tab:epochs} gives the simulated acceptance lengths behind \cref{sec:epochs}: the
three-epoch fine-tuning runs after each epoch, and both drafters trained from scratch for six epochs. \dpace{} stops
improving after one or two epochs of fine-tuning and after five from scratch, while the lead of \mfd{} grows with every
epoch.

\begin{table}[h]
  \centering
  \caption{Training longer (Qwen3-4B; simulated $\tau$ at $\B{=}256$ with a uniform prior, 5-task means; in
  parentheses, the change of \mfd{} relative to \dpace{} in \%). Top: the three-epoch fine-tuning runs of
  \cref{tab:main} after each epoch. Bottom: both drafters trained from scratch with \dflash{}'s schedule and the \dpace{} loss.}
  \label{tab:epochs}
  \small
  \begin{tabular}{c cc cc}
    \toprule
    & \multicolumn{2}{c}{$T{=}0$} & \multicolumn{2}{c}{$T{=}1$} \\
    \cmidrule(lr){2-3}\cmidrule(lr){4-5}
    Epoch & \dpace{} & \mfd{} & \dpace{} & \mfd{} \\
    \midrule
    \multicolumn{5}{l}{\emph{Fine-tuned from the released drafter}} \\
    1 & 8.76 & 9.43 ($+7.7$) & 7.67 & 8.35 ($+8.9$) \\
    2 & 8.77 & 9.55 ($+8.8$) & 7.65 & 8.43 ($+10.2$) \\
    3 & 8.70 & 9.54 ($+9.7$) & 7.60 & 8.42 ($+10.8$) \\
    \midrule
    \multicolumn{5}{l}{\emph{Trained from scratch}} \\
    1 & 6.25 & 5.87 ($-6.0$) & 5.69 & 5.41 ($-4.9$) \\
    2 & 6.83 & 6.60 ($-3.4$) & 6.16 & 6.03 ($-2.0$) \\
    3 & 7.15 & 7.09 ($-1.0$) & 6.44 & 6.42 ($-0.3$) \\
    4 & 7.35 & 7.42 ($+1.0$) & 6.58 & 6.71 ($+2.0$) \\
    5 & 7.43 & 7.63 ($+2.7$) & 6.62 & 6.88 ($+4.0$) \\
    6 & 7.36 & 7.65 ($+3.9$) & 6.57 & 6.90 ($+5.0$) \\
    \bottomrule
  \end{tabular}
\end{table}

\textbf{Data.} The 100K data are the 99,987 Qwen3-4B training samples that the \dflash{} authors released on Hugging Face (\nolinkurl{z-lab/qwen3-4b-instruct-100k}, no longer public; we use its public mirror \nolinkurl{TY233/qwen3-4b-instruct-100k-backup}), with prompts from Nemotron post-training data, OpenCodeInstruct, OpenR1-Math and Evol-CodeAlpaca; the released drafters were trained on about 800K samples \citep{chen2026dflash}. Qwen3-4B uses the released responses; the
other targets regenerate them at $T{=}0.7$, top-$p$ 0.8 and top-$k$ 20: Qwen3-8B without thinking and with at most
2048 new tokens, Qwen3.5-4B and Qwen3.6-27B in thinking mode with at most 3072, keeping responses cut at that length.
The full data are the 1.42M prompts of the open-perfectblend set on which Domino trains \citep{domino2026}, with every
assistant turn regenerated by each target in the same way (Qwen3-4B like Qwen3-8B).

\textbf{Ablations.} The quarter-epoch drafters of \cref{sec:analysis} are fine-tuned on Qwen3-4B from the released
drafter with the data and schedule of its one-epoch runs, and stopped after 2,066 of the 8,263 steps, at about 89\% of
the peak learning rate; the single-mode \dpace{} reference is the one-epoch \dpace{} run at the same step. Each
sequence drafts 512 blocks for the single-mode drafters and for the injection study at the top of \cref{tab:inject}
(4 modes $\times$ 128 anchors), and 1024 for the $\K$ sweep of \cref{tab:hedge} ($\K$ modes $\times$ $1024/\K$
anchors) and the four-mode drafters at the bottom of \cref{tab:inject}. The paired simulations run the target over its responses to 128
prompts per task (a seed-42 subset of each benchmark; all 80 for MT-Bench), generated greedily for $T{=}0$ and by pure
sampling for $T{=}1$, with the token limits of the timing protocol below. From each response's first token, a round accepts the
deepest prefix of the response's next tokens that the drafted tree or chains contain, and the next round starts after
it, as decoding would: at $T{=}0$ this replays greedy speculative decoding exactly, and at $T{=}1$ it is a Monte Carlo
estimate of sampled verification, with the same samples for every drafter. Mixture trees use the max rule with the
drafter's own prior unless noted (\cref{app:rules}).

\textbf{Hardware.} Speed is measured on one NVIDIA RTX PRO 6000 Blackwell GPU (96\,GB) per process. Training runs on
8-GPU nodes, where some GPUs serve the target's hidden states to trainers on the others: one H200 node for each
three-epoch or from-scratch run on the 100K data, three RTX PRO 6000 nodes for each full-data run, and four RTX PRO
6000 GPUs (one serving the target, three training) for the quarter- and one-epoch runs.

\textbf{Timing protocol.} Timing uses Hugging Face Transformers (bf16, SDPA, eager execution, batch size 1) on the
prompts that \ddtree{}'s evaluation code selects from five of its ten benchmarks, the first ones of a seed-0 shuffle of each (the simulations use seed-42 subsets): 128 each for GSM8K,
MATH-500 and MBPP, all 164 for HumanEval and all 80 two-turn conversations of MT-Bench. Decoding stops after 2048 new
tokens on Qwen3-4B and Qwen3-8B (without thinking) and after 4096 on the thinking-mode targets. For each
target, every drafter is loaded into one process, and for each prompt the autoregressive baseline and every (drafter,
decoding) pair decode back to back, in an order rotated by the prompt index; multi-turn MT-Bench prompts continue from
the baseline's answer, so that every method sees the same conversation. Decoding time excludes the prefill. At $T{=}1$, trees and chains
verify by sampling the target's token at each step and accepting the drafted token that equals it, as in \ddtree{};
the sampled selector chain of \dflash{}~2 uses speculative sampling, and EAGLE-3 its own rule. For each
task, the speedup is the baseline's decoding time per output token divided by the method's, both pooled over the
task's prompts, and the tables report the mean over the five tasks. Long tasks are split into shards of consecutive
prompts that continue the rotation, each run as a separate process with its own baseline on one GPU of the same type.

\textbf{Timing runs.} \Cref{tab:main} combines several runs of this protocol, each with its own autoregressive
baseline: for each target, a main run of the released drafter and the three-epoch drafters (and of the one-epoch
drafters, except on Qwen3.6-27B) at $\B\in\{16,64,256\}$; re-timings of the Qwen3 trees at $\B{=}128$, the best budget
on Qwen3-8B; a run of \dflash{}~2 with its fused draft path; the EAGLE-3 runs; the full-data runs; and on Qwen3.6-27B,
a run of the one-epoch checkpoints. Each \mfd{} drafter is timed in the same run as the \dpace{} drafter trained for as
long, and every run also times a \dpace{} drafter: the same drafter at the same budget reproduces its speedup across
runs within 1.9\% on three targets and within 3.3\% on Qwen3-8B, whose re-timing ran fastest. In the run of
\dflash{}~2, its tree trails \dpace{} by 0.6--2.5\% (8.8--10.9\% on Qwen3.6-27B). In \cref{tab:main}, EAGLE-3 comes from its runs, the \dflash{} chain from the main runs and the \dflash{}~2 rows
from their run; the released \ddtree{}, \dpace{} and \mfd{} rows come from the main runs, except on Qwen3-8B, where
they come from the re-timing at $\B{=}128$, and the one-epoch rows on Qwen3.6-27B, which come from the checkpoint
run; the bottom block comes from the full-data runs. In \cref{fig:teaser}c, $\B{=}128$ comes from the re-timings
and the chain of \dflash{}~2 from its own run.

\textbf{Harness optimizations.} We sped up the harness of \ddtree{} for all methods alike and checked every change
against the original: trees are built in C++ (heap, ancestor mask and path walk) and match the original builder node
for node; the top-$k$ candidates are taken on the bf16 logits and the log-normalizers come from one Triton pass, equal
to the float32 computation up to summation order; and the KV cache is compacted to the accepted path in C++. Two
kernels serve one drafter type each: the mode blocks of \mfd{} attend through a Triton kernel that splits the context
in flash-decoding style instead of SDPA with a dense block-diagonal mask, equal up to summation order, and the draft
path of \dflash{}~2 (grouped convolutions and selector) runs as fused CUDA graphs, as in SGLang, which leaves its
greedy outputs unchanged token for token. On the hybrid targets, every method verifies through
the kernels of \cref{app:hybrid}.

\textbf{Tree budgets.} In \cref{tab:main}, the best budget, chosen on the timed prompts by the five-task mean
speedup, is $\B{=}256$ on Qwen3-4B and Qwen3.5-4B, 128 on Qwen3-8B and 64 on Qwen3.6-27B, the same for every tree method and in both blocks, except that at $T{=}1$ the full-data \dpace{} tree on Qwen3-4B is 0.1\% faster at $\B{=}128$; the full-data trees on Qwen3-4B and Qwen3-8B are timed at
$\B\in\{64,128,256\}$ in the same harness as DARTree's released code. A tree with budget $\B$ verifies $\B{+}1$ tokens, so the budgets of \ddtree{} and DARTree
($\B{=}64$, 256) put one row past a power of two into every GEMM and attention call of the verification. In our
harness this costs little: on Qwen3-4B ($T{=}0$, 32 prompts each of GSM8K and HumanEval, the interleaved protocol of
\cref{sec:experiments}), $\B{=}63$ shortens a round by 0.1--0.4\% against $\B{=}64$ and $\B{=}255$ by 1.0--1.7\%
against $\B{=}256$, for \dpace{} trees and \mfd{} alike, and acceptance lengths move by less than 1\% (apart from one
HumanEval slice whose greedy output changed). On Qwen3-8B, verifying 257 tokens is markedly slower: in the run of
\dflash{}~2, a round's verification takes 20.9, 21.2 and 26.5\,ms at $\B{=}64$, 128 and 256, against 20.5, 20.3 and
20.8\,ms on Qwen3-4B, which is why every tree slows down at $\B{=}256$ on Qwen3-8B (\cref{fig:teaser}c). A round on Qwen3-4B is dominated by fixed costs: an autoregressive step takes
21\,ms, a 16-token chain round 29.1\,ms and a 64-token tree round 29.6\,ms. In SGLang, where CUDA graphs remove the
launch overhead, the extra tile shows: at batch 32, $\B{=}16$ makes a round 34\% longer than $\B{=}15$ (25.0 against
18.6\,ms, 544 against 512 verified rows), and at batch 1 the \dpace{} tree is 3.7\% faster at $\B{=}63$ than at
$\B{=}64$ (Qwen3-4B, $T{=}0$; within 2\% on Qwen3-8B). The engine results of \cref{sec:engine} therefore use
$\B \in \{7, 15, 31, 63\}$.

\textbf{Serving engine.} Every SGLang configuration runs with CUDA graphs, the FlashInfer attention backend, bf16
weights, the radix cache off and a static memory fraction of 0.6 on one RTX PRO 6000 GPU.

\textbf{EAGLE-3.} We use the released heads \texttt{WhiFlash/Qwen3-4B-EAGLE3} and \texttt{WhiFlash/Qwen3-8B-EAGLE3},
the best of the released heads we screened for these targets, and port EAGLE's Hugging Face inference into our
harness, so EAGLE-3 runs in the interleaved protocol of \cref{sec:experiments}, with its own autoregressive baseline
and a \dpace{} reference run in the same process. Its dynamic tree has three settings (draft steps, top-$k$ per step and draft tokens,
in SGLang's terms). On 16 prompts per task we tried eleven settings on Qwen3-4B and seven on Qwen3-8B, from
(7, 8, 64) and EAGLE's default (8, 10, 60) up to (14, 16, 320) and (12, 12, 256), and kept the best mean of the
$T{=}0$ and $T{=}1$ speedups: (12, 12, 256) on Qwen3-4B and (10, 10, 128) on Qwen3-8B, where larger trees are slower.
At $T{=}1$ it uses EAGLE's own acceptance rule. With target and heads in fp32, its greedy outputs equal
autoregressive decoding token for token; in bf16 they split at near-ties, as every tree method's do. Its sequential
draft takes about 1.7 times as long as the one-pass \dpace{} draft in the same runs (7.0--7.5 against
4.1--4.5\,ms).

\textbf{DARTree.} We run DARTree's released configuration, a pruned tree of 64 nodes with its released code and
drafters. At the same 64 nodes, the full-data \mfd{} is 3.7\% / 7.5\% faster on Qwen3-4B and 3.3\% / 5.5\% faster on
Qwen3-8B ($T{=}0$ / $T{=}1$), although DARTree accepts more tokens per round (9.97 against 9.67 and 10.09 against 9.76
at $T{=}0$): its per-depth correction takes 2.6--2.7\,ms per round to build the tree, against 0.35\,ms for \mfd{}. With 64 nodes, \mfd{} fine-tuned on the 100K data is 4.4\% slower than DARTree
at $T{=}0$ and 0.4\% faster at $T{=}1$ on Qwen3-4B, and 4.8\% slower and on par on Qwen3-8B; at its best budget ($\B{=}256$ and 128), it is 3.7\% / 8.7\% and 1.4\% / 6.4\% faster in the same runs, the range quoted in \cref{sec:main}. The factor of 14 in \cref{sec:main} 
counts fine-tuning data only: the released checkpoints that \mfd{} starts from were trained on about 800K samples. We report DARTree's pruned tree, which is 5\% faster than
its fixed one on Qwen3-4B; the Domino chain reaches 5.00$\times$ / 4.06$\times$. JetSpec's released Qwen3-8B head, run as its default one-pass
draft (an optional extension in its released code, which extends the best chain past the block by conditioning on drafted tokens, is off by default) and decoded with the same tree,
reaches 4.86$\times$ / 4.17$\times$ at $\B{=}64$, the best of the budgets we timed (16, 64 and 256), below the released
\dflash{} drafter at that budget (5.32$\times$ / 4.66$\times$); \mfd{} is 24\% / 20\% faster at $\B{=}64$.

\FloatBarrier
\section{Serving at larger batch sizes}
\label{app:batch}

\begin{figure}[h]
  \centering
  \includegraphics[width=0.55\linewidth]{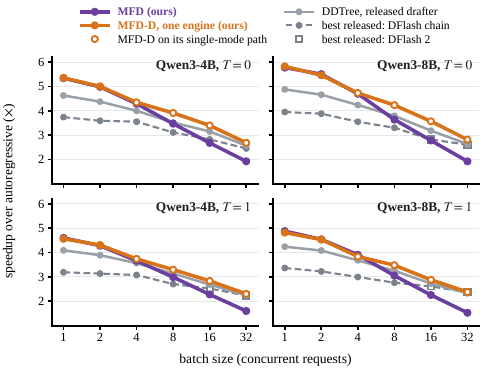}
  \caption{Speedup over SGLang's autoregressive decoding at the same batch size (top: $T{=}0$, bottom: $T{=}1$).
  Closed loop: 320 requests from the five tasks with as many in flight as the batch size, at most 2048 new tokens,
  steady-state output throughput (prefills of new requests included, so batch-1 values lie below those of
  \cref{tab:sglang}), one RTX PRO 6000 GPU. \mfd{} and \mfd{}-D run their drafter's linear layers in FP8 (SGLang's
  online option: per-channel weights, per-token activations, no new checkpoint; the acceptance length is unchanged).
  Every released option is tuned for each batch size and shown at the better of its bf16 and FP8 timings: trees over
  two or three budgets in $\{7,15,31,63\}$, the \dflash{} chain over 16-token and, from batch 4, 8-token blocks,
  EAGLE-3 over 6--10 tree shapes in bf16 and the best of them in FP8. Dashed: the fastest released option on
  SGLang's own engine paths (\dflash{} chain, \dflash{}~2, EAGLE-3) at each batch size, which is the \dflash{} chain
  except where \dflash{}~2 is faster (open squares). \ddtree{}: the released \dflash{} drafter in our tree engine.
  \mfd{}-D is one engine with a fixed schedule: its eight-mode mixture for one or two requests, its single-mode
  default path (open circles) from three, with the budget falling from 63 to 7 as the batch grows. Every
  configuration is timed against the autoregressive reference of its own session; the references of different
  sessions agree within 0.8\%, and a configuration re-timed in bf16 within 2.8\%.}
  \label{fig:engine}
\end{figure}

\textbf{Engine.} Our SGLang implementation verifies trees inside the target's CUDA graphs with the overlap scheduler,
builds the tree on the host as in \cref{sec:method}, runs the mode attention as a Triton kernel (a fused single-pass
variant from batch 12, where the split kernel's partial sums dominate), and commits with a
kernel that moves only the KV entries of the accepted path that change, which at batch 16 cuts the commit from 1.75 to
0.32\,ms per round with identical outputs (Qwen3-4B, $\B{=}16$). Both trees draft over the 64K tokens most frequent
in the target's training responses, which hold 99.8\% of the tokens of its responses on the five tasks (Qwen3-4B),
cost about 0.6\% of $\tau$ and make the drafter's output layer 32--60\% cheaper; verification uses the full vocabulary,
so decoding stays lossless (at $T{=}1$, 800 sampled six-token continuations of one prompt equal those of the full vocabulary token for token).
The timed engine configurations use budgets one below
a power of two, $\B\in\{7,15,31,63\}$, so that the verified tokens fill the kernels' tiles; in the harness of
\cref{tab:main}, this rounding shortens a round by at most 1.7\% (\cref{app:details}).

\textbf{Released drafters.} The released drafters run as distributed, each at its best setting: the \dflash{} drafters
\citep{chen2026dflash} as 16-token chains, community \dflash{}~2 releases \citep{inco2026dflash2} with their candidate
selectors (9-token blocks for Qwen3-4B, 8 for Qwen3-8B; configuration converted to SGLang's layout, weights
unchanged), and the fastest of the released EAGLE-3 heads \citep{li2025eagle3} we screened (two for Qwen3-4B, four for
Qwen3-8B) at the best of ten tree shapes at batch size 1 (7 steps, top-8, 64 draft tokens; 8 / 10 / 96 on Qwen3-8B at
$T{=}0$). At batch size 1, EAGLE-3 is slower than the \dflash{} chain, and so are the \dflash{}~2 releases, whose 9-
and 8-token blocks cap a round at eight and seven drafted tokens; the trees of \cref{tab:sglang} run at the faster of
$\B\in\{31,63\}$ ($\B{=}63$ throughout). \mfd{}-D, the full-data \dpace{} tree and the \dflash{}~2 releases were timed in later sessions, in which \mfd{}
reproduced its speedups within 0.3\%.

\textbf{Batch size.} With as many requests in flight as the batch size (\cref{fig:engine}), the lead of \mfd{} is a
small-batch one. The figure runs our drafters with their linear layers in FP8 (SGLang's online option: per-channel
weights, per-token activations, no new checkpoint), which leaves the acceptance length unchanged ($-0.2$\% on average
at a fixed budget, within 3\% per run) and batch 1--2 within noise, and speeds \mfd{} up by 4--8\% at batch 4--8 and
8--15\% at batch 16--32, where its eight mode blocks make the draft stage the bottleneck. The single-mode drafters
gain little: \ddtree{} with the released drafter 3--6\% at batch 32 and nothing below, SGLang's own paths nothing at any batch size
(within 4\%; SGLang also disables its fused KV materialization for a quantized \dflash{} drafter, on our tree path as
well). The figure therefore shows every released option at the better of its bf16 and FP8 timings. \mfd{} is then
10--18\% faster than every released option at batch 1--2 and 3--11\% faster at batch 4, 1--6\% behind \ddtree{} with the released drafter at
batch 8, behind SGLang's own paths from batch 16, and 25--36\% slower than the best released option at batch
32; in bf16 the same picture shifts by one batch size (9--17\% faster at batch 1--2, 0.7--4.0\% faster than that tree
at batch 4, behind it from batch 8, 29--42\% slower than the best released option at batch 32). Its acceptance length
does not change with the batch size (within 2\% at a fixed budget at $T{=}0$ and 3.1\% at $T{=}1$); its draft cost
does. The eight mode blocks add 0.2\,ms per round at batch 1, but at batch 32 the draft stage of \mfd{} takes
18.6\,ms per round against 3.9\,ms for a single-mode drafter (Qwen3-4B, $\B{=}16$, bf16). Large batches favour cheap
rounds: from batch 16 the short \dflash{}~2 chains overtake the \dflash{} chain in most settings. \mfd{}-D keeps both
regimes in one drafter. Fine-tuned like \mfd{} with a ninth, default mode (the drafter without a mode embedding,
trained with the \dpace{} objective; about twice the training compute of \mfd{}), it shares every other weight
between its two paths, so one engine switches by batch size: the mixture for one or two requests, the default path
from three. It matches \mfd{} up to batch 4 ($-1.8$\% to $+2.2$\% in \cref{tab:sglang} and \cref{fig:engine}) and is the fastest
configuration we timed at every batch size: 4--19\% faster than every released option up to batch 16 and 0.4--7\%
faster at batch 32, where its default path at $\B{=}7$ meets the \dflash{}~2 release (Qwen3-8B, $T{=}1$).

\textbf{Limitations.} \mfd{} speeds up latency-bound decoding. In SGLang it is faster than the released options and \ddtree{} with the released drafter at batch sizes 1--4, but its lead shrinks as the batch grows: it is 3--11\% ahead of that tree at batch 4 and falls
behind from batch 8, because its $\K$ mode blocks make drafting compute-bound. A drafter trained with an additional default mode, at about twice the training
compute, avoids the loss: one engine switches to its single-mode path as the batch grows and is the fastest
configuration we timed at every batch size (\cref{sec:engine}), but from three requests on it no longer
uses the mixture. Keeping the mixture's gain at large batch sizes would need a cheaper mixture, for example drafting
the modes for only some requests or a smaller drafter head. The hybrid Gated-DeltaNet targets are timed only in our
harness.

\section{Modes on a drafter that already conditions}
\label{sec:combine}

The two routes of \cref{sec:theory} close the same gap. To test whether they add up, we put 8 input modes on the
released Qwen3-4B Domino drafter, whose correction head conditions every position on the branch's own tokens, and
fine-tune it with \dpace{} on the corrected logits inside each mode (winner-take-all), against the same fine-tune with
one mode. The mixture DARTree grows one corrected beam per mode and merges them by prefix under the max rule; with one
mode it reproduces DARTree's trees exactly. On three tasks (16 prompts each, $T{=}0$), the modes change $\tau$ by
$-1.3\%$ at $\B{=}64$ and $+1.3\%$ at $\B{=}128$: once a drafter conditions on the realized prefix, a latent future has
little left to explain. During teacher-forced training the correction head already sees which future the data takes,
so the modes have nothing to specialize on. In this small test, modes did not help a drafter that
already conditions on its branch; \mfd{} reaches its gains without sequential correction.

%% file: sections/pertask_tables.tex

\Cref{tab:pertask-q3-4b,tab:pertask-q3-8b,tab:pertask-q35-4b,tab:pertask-q36-27b} break every cell of \cref{tab:main} down by task, with the same protocol, runs and budgets. \mfd{} is the fastest method on every task, for every target and temperature.

\begin{table}[!htbp]
  \centering
  \caption{Per-task speedup over autoregressive decoding and $\tau$ (in parentheses) on Qwen3-4B: the cells of \cref{tab:main} by task, from the same runs (every tree at $\B{=}256$ except EAGLE-3's, tuned as in \cref{app:details}; DARTree verifies 64 nodes). Bold: fastest in each column within each block.}
  \label{tab:pertask-q3-4b}
  \small
  \setlength{\tabcolsep}{3pt}
  \begin{tabular}{l l ccccc}
    \toprule
    Method & Training & GSM8K & MATH-500 & HumanEval & MBPP & MT-Bench \\
    \midrule
    \multicolumn{7}{l}{\emph{$T{=}0$}} \\
    EAGLE-3 & released & 4.67$\times$ (7.11) & 3.88$\times$ (6.00) & 4.25$\times$ (6.51) & 4.31$\times$ (6.59) & 3.91$\times$ (6.10) \\
    \dflash{} chain & released & 4.62$\times$ (6.34) & 5.58$\times$ (7.59) & 4.81$\times$ (6.56) & 4.35$\times$ (5.97) & 2.54$\times$ (3.44) \\
    \dflash{}~2 selector chain & FT 1 ep & 5.07$\times$ (7.11) & 5.97$\times$ (8.35) & 5.17$\times$ (7.26) & 4.73$\times$ (6.65) & 2.70$\times$ (3.77) \\
    \ddtree{} & released & 6.47$\times$ (9.01) & 7.19$\times$ (10.29) & 6.58$\times$ (9.28) & 6.30$\times$ (8.82) & 3.85$\times$ (5.54) \\
    \dflash{}~2 + \ddtree{} & FT 1 ep & 6.59$\times$ (9.38) & 7.33$\times$ (10.68) & 6.66$\times$ (9.60) & 6.19$\times$ (8.86) & 3.88$\times$ (5.67) \\
    \dpace{} + \ddtree{} & FT 1 ep & 6.82$\times$ (9.49) & 7.47$\times$ (10.70) & 6.87$\times$ (9.69) & 6.42$\times$ (8.98) & 3.97$\times$ (5.70) \\
    \dpace{} + \ddtree{} & FT 3 ep & 6.72$\times$ (9.33) & 7.47$\times$ (10.67) & 6.67$\times$ (9.40) & 6.31$\times$ (8.81) & 3.94$\times$ (5.66) \\
    \cmidrule{1-7}
    \mfd{} ($\K{=}8$) & FT 1 ep & 7.38$\times$ (10.49) & 7.88$\times$ (11.53) & \textbf{7.22$\times$} (10.43) & 6.77$\times$ (9.68) & 4.11$\times$ (6.04) \\
    \mfd{} ($\K{=}8$) & FT 3 ep & \textbf{7.39$\times$} (10.49) & \textbf{7.98$\times$} (11.67) & 7.21$\times$ (10.38) & \textbf{6.89$\times$} (9.84) & \textbf{4.19$\times$} (6.16) \\
    \midrule
    \multicolumn{7}{l}{\emph{$T{=}0$, full data: 1.42M target-regenerated prompts}} \\
    DARTree & released & 8.26$\times$ (12.58) & 7.46$\times$ (11.40) & 6.46$\times$ (9.87) & 6.48$\times$ (9.90) & 4.00$\times$ (6.11) \\
    \dpace{} + \ddtree{} & FT 3 ep & 8.06$\times$ (11.18) & 7.85$\times$ (11.09) & 6.81$\times$ (9.55) & 6.38$\times$ (8.94) & 4.20$\times$ (5.96) \\
    \mfd{} ($\K{=}8$) & FT 3 ep & \textbf{9.05$\times$} (12.83) & \textbf{8.52$\times$} (12.33) & \textbf{7.44$\times$} (10.68) & \textbf{7.19$\times$} (10.27) & \textbf{4.37$\times$} (6.36) \\
    \midrule
    \multicolumn{7}{l}{\emph{$T{=}1$}} \\
    EAGLE-3 & released & 4.39$\times$ (6.78) & 3.52$\times$ (5.52) & 3.93$\times$ (6.11) & 4.04$\times$ (6.21) & 3.47$\times$ (5.45) \\
    \dflash{} chain & released & 4.11$\times$ (5.62) & 4.39$\times$ (5.92) & 4.14$\times$ (5.62) & 3.78$\times$ (5.17) & 2.25$\times$ (3.04) \\
    \dflash{}~2 selector chain & FT 1 ep & 4.59$\times$ (6.47) & 4.93$\times$ (6.90) & 4.54$\times$ (6.39) & 4.10$\times$ (5.75) & 2.46$\times$ (3.43) \\
    \ddtree{} & released & 5.91$\times$ (8.28) & 6.01$\times$ (8.61) & 6.05$\times$ (8.53) & 5.73$\times$ (8.03) & 3.48$\times$ (5.04) \\
    \dflash{}~2 + \ddtree{} & FT 1 ep & 6.05$\times$ (8.64) & 5.87$\times$ (8.54) & 5.96$\times$ (8.59) & 5.58$\times$ (7.99) & 3.48$\times$ (5.12) \\
    \dpace{} + \ddtree{} & FT 1 ep & 6.21$\times$ (8.68) & 6.02$\times$ (8.62) & 6.15$\times$ (8.68) & 5.81$\times$ (8.14) & 3.52$\times$ (5.09) \\
    \dpace{} + \ddtree{} & FT 3 ep & 6.03$\times$ (8.42) & 5.93$\times$ (8.46) & 6.03$\times$ (8.51) & 5.52$\times$ (7.74) & 3.47$\times$ (5.03) \\
    \cmidrule{1-7}
    \mfd{} ($\K{=}8$) & FT 1 ep & 6.67$\times$ (9.54) & 6.36$\times$ (9.32) & \textbf{6.60$\times$} (9.54) & 6.22$\times$ (8.90) & 3.66$\times$ (5.42) \\
    \mfd{} ($\K{=}8$) & FT 3 ep & \textbf{6.80$\times$} (9.70) & \textbf{6.55$\times$} (9.60) & 6.52$\times$ (9.45) & \textbf{6.23$\times$} (8.92) & \textbf{3.70$\times$} (5.49) \\
    \midrule
    \multicolumn{7}{l}{\emph{$T{=}1$, full data: 1.42M target-regenerated prompts}} \\
    DARTree & released & 6.87$\times$ (10.42) & 5.66$\times$ (8.58) & 5.82$\times$ (8.83) & 5.62$\times$ (8.58) & 3.51$\times$ (5.31) \\
    \dpace{} + \ddtree{} & FT 3 ep & 6.81$\times$ (9.49) & 6.01$\times$ (8.53) & 6.09$\times$ (8.56) & 5.70$\times$ (8.01) & 3.69$\times$ (5.22) \\
    \mfd{} ($\K{=}8$) & FT 3 ep & \textbf{8.06$\times$} (11.45) & \textbf{6.77$\times$} (9.84) & \textbf{6.67$\times$} (9.60) & \textbf{6.30$\times$} (9.04) & \textbf{3.94$\times$} (5.71) \\
    \bottomrule
  \end{tabular}
\end{table}

\begin{table}[!htbp]
  \centering
  \caption{Per-task speedup over autoregressive decoding and $\tau$ (in parentheses) on Qwen3-8B: the cells of \cref{tab:main} by task, from the same runs (every tree at $\B{=}128$ except EAGLE-3's, tuned as in \cref{app:details}; DARTree verifies 64 nodes). Bold: fastest in each column within each block.}
  \label{tab:pertask-q3-8b}
  \small
  \setlength{\tabcolsep}{3pt}
  \begin{tabular}{l l ccccc}
    \toprule
    Method & Training & GSM8K & MATH-500 & HumanEval & MBPP & MT-Bench \\
    \midrule
    \multicolumn{7}{l}{\emph{$T{=}0$}} \\
    EAGLE-3 & released & 4.22$\times$ (6.77) & 3.56$\times$ (5.86) & 3.84$\times$ (6.09) & 3.82$\times$ (6.09) & 3.29$\times$ (5.55) \\
    \dflash{} chain & released & 4.55$\times$ (6.38) & 5.57$\times$ (7.85) & 4.57$\times$ (6.40) & 4.17$\times$ (5.89) & 2.42$\times$ (3.42) \\
    \dflash{}~2 selector chain & FT 1 ep & 5.09$\times$ (7.24) & 6.18$\times$ (8.82) & 5.06$\times$ (7.21) & 4.71$\times$ (6.69) & 2.62$\times$ (3.73) \\
    \ddtree{} & released & 6.21$\times$ (8.69) & 6.97$\times$ (10.14) & 6.28$\times$ (8.88) & 5.78$\times$ (8.10) & 3.43$\times$ (5.10) \\
    \dflash{}~2 + \ddtree{} & FT 1 ep & 6.35$\times$ (9.18) & 6.99$\times$ (10.63) & 6.28$\times$ (9.34) & 5.88$\times$ (8.49) & 3.51$\times$ (5.36) \\
    \dpace{} + \ddtree{} & FT 1 ep & 6.60$\times$ (9.21) & 7.38$\times$ (10.69) & 6.63$\times$ (9.34) & 6.08$\times$ (8.50) & 3.61$\times$ (5.37) \\
    \dpace{} + \ddtree{} & FT 3 ep & 6.43$\times$ (8.97) & 7.39$\times$ (10.70) & 6.65$\times$ (9.37) & 6.16$\times$ (8.64) & 3.57$\times$ (5.32) \\
    \cmidrule{1-7}
    \mfd{} ($\K{=}8$) & FT 1 ep & 7.06$\times$ (10.10) & 7.72$\times$ (11.50) & 6.95$\times$ (10.02) & 6.60$\times$ (9.45) & 3.73$\times$ (5.69) \\
    \mfd{} ($\K{=}8$) & FT 3 ep & \textbf{7.14$\times$} (10.17) & \textbf{7.90$\times$} (11.71) & \textbf{7.13$\times$} (10.29) & \textbf{6.62$\times$} (9.44) & \textbf{3.79$\times$} (5.76) \\
    \midrule
    \multicolumn{7}{l}{\emph{$T{=}0$, full data: 1.42M target-regenerated prompts}} \\
    DARTree & released & 8.00$\times$ (12.51) & 7.46$\times$ (11.75) & 6.53$\times$ (10.24) & 6.26$\times$ (9.77) & 3.87$\times$ (6.17) \\
    \dpace{} + \ddtree{} & FT 3 ep & 7.80$\times$ (10.93) & 7.75$\times$ (11.17) & 6.61$\times$ (9.40) & 6.23$\times$ (8.77) & 3.76$\times$ (5.61) \\
    \mfd{} ($\K{=}8$) & FT 3 ep & \textbf{8.72$\times$} (12.47) & \textbf{8.24$\times$} (12.18) & \textbf{7.12$\times$} (10.36) & \textbf{6.82$\times$} (9.78) & \textbf{4.01$\times$} (6.13) \\
    \midrule
    \multicolumn{7}{l}{\emph{$T{=}1$}} \\
    EAGLE-3 & released & 3.98$\times$ (6.41) & 3.15$\times$ (5.25) & 3.55$\times$ (5.70) & 3.53$\times$ (5.66) & 2.86$\times$ (4.87) \\
    \dflash{} chain & released & 4.00$\times$ (5.61) & 4.00$\times$ (5.61) & 3.84$\times$ (5.37) & 3.46$\times$ (4.86) & 2.09$\times$ (2.94) \\
    \dflash{}~2 selector chain & FT 1 ep & 4.56$\times$ (6.56) & 4.79$\times$ (6.88) & 4.41$\times$ (6.34) & 3.95$\times$ (5.66) & 2.29$\times$ (3.30) \\
    \ddtree{} & released & 5.66$\times$ (7.94) & 5.55$\times$ (8.09) & 5.43$\times$ (7.68) & 5.24$\times$ (7.35) & 3.14$\times$ (4.59) \\
    \dflash{}~2 + \ddtree{} & FT 1 ep & 5.68$\times$ (8.26) & 5.41$\times$ (8.27) & 5.29$\times$ (7.88) & 5.06$\times$ (7.38) & 2.96$\times$ (4.59) \\
    \dpace{} + \ddtree{} & FT 1 ep & 5.93$\times$ (8.29) & 5.61$\times$ (8.17) & 5.52$\times$ (7.79) & 5.24$\times$ (7.34) & 3.23$\times$ (4.71) \\
    \dpace{} + \ddtree{} & FT 3 ep & 5.73$\times$ (8.01) & 5.62$\times$ (8.16) & 5.50$\times$ (7.75) & 5.17$\times$ (7.25) & 3.15$\times$ (4.59) \\
    \cmidrule{1-7}
    \mfd{} ($\K{=}8$) & FT 1 ep & 6.30$\times$ (9.02) & \textbf{5.98$\times$} (8.92) & 5.90$\times$ (8.52) & \textbf{5.62$\times$} (8.06) & 3.24$\times$ (4.86) \\
    \mfd{} ($\K{=}8$) & FT 3 ep & \textbf{6.45$\times$} (9.22) & 5.91$\times$ (8.80) & \textbf{5.93$\times$} (8.54) & 5.59$\times$ (8.01) & \textbf{3.28$\times$} (4.90) \\
    \midrule
    \multicolumn{7}{l}{\emph{$T{=}1$, full data: 1.42M target-regenerated prompts}} \\
    DARTree & released & 6.92$\times$ (10.67) & 5.37$\times$ (8.40) & 5.31$\times$ (8.24) & 5.29$\times$ (8.19) & 3.25$\times$ (5.06) \\
    \dpace{} + \ddtree{} & FT 3 ep & 6.55$\times$ (9.15) & 5.65$\times$ (8.20) & 5.48$\times$ (7.77) & 5.32$\times$ (7.48) & 3.30$\times$ (4.79) \\
    \mfd{} ($\K{=}8$) & FT 3 ep & \textbf{7.71$\times$} (10.97) & \textbf{6.25$\times$} (9.26) & \textbf{6.16$\times$} (8.93) & \textbf{5.81$\times$} (8.32) & \textbf{3.44$\times$} (5.12) \\
    \bottomrule
  \end{tabular}
\end{table}

\begin{table}[!htbp]
  \centering
  \caption{Per-task speedup over autoregressive decoding and $\tau$ (in parentheses) on Qwen3.5-4B (thinking): the cells of \cref{tab:main} by task, from the same runs (every tree at $\B{=}256$). Bold: fastest in each column within each block.}
  \label{tab:pertask-q35-4b}
  \small
  \setlength{\tabcolsep}{3pt}
  \begin{tabular}{l l ccccc}
    \toprule
    Method & Training & GSM8K & MATH-500 & HumanEval & MBPP & MT-Bench \\
    \midrule
    \multicolumn{7}{l}{\emph{$T{=}0$}} \\
    EAGLE-3 & released & -- & -- & -- & -- & -- \\
    \dflash{} chain & released & 4.52$\times$ (5.82) & 5.41$\times$ (6.97) & 5.43$\times$ (7.01) & 4.98$\times$ (6.43) & 3.88$\times$ (5.00) \\
    \dflash{}~2 selector chain & FT 1 ep & 5.05$\times$ (6.59) & 5.93$\times$ (7.72) & 6.03$\times$ (7.86) & 5.54$\times$ (7.21) & 4.34$\times$ (5.66) \\
    \ddtree{} & released & 6.14$\times$ (8.25) & 7.10$\times$ (9.57) & 7.22$\times$ (9.74) & 6.76$\times$ (9.07) & 5.40$\times$ (7.28) \\
    \dflash{}~2 + \ddtree{} & FT 1 ep & 6.45$\times$ (8.85) & 7.34$\times$ (10.04) & 7.61$\times$ (10.41) & 6.93$\times$ (9.39) & 5.67$\times$ (7.76) \\
    \dpace{} + \ddtree{} & FT 1 ep & 6.61$\times$ (8.87) & 7.41$\times$ (9.98) & 7.73$\times$ (10.43) & 7.32$\times$ (9.82) & 5.75$\times$ (7.74) \\
    \dpace{} + \ddtree{} & FT 3 ep & 6.70$\times$ (9.00) & 7.50$\times$ (10.10) & 7.58$\times$ (10.23) & 7.27$\times$ (9.75) & 5.79$\times$ (7.81) \\
    \cmidrule{1-7}
    \mfd{} ($\K{=}8$) & FT 1 ep & 6.95$\times$ (9.56) & 7.78$\times$ (10.75) & 7.89$\times$ (10.93) & 7.28$\times$ (10.00) & 5.92$\times$ (8.17) \\
    \mfd{} ($\K{=}8$) & FT 3 ep & \textbf{7.03$\times$} (9.68) & \textbf{7.88$\times$} (10.89) & \textbf{8.03$\times$} (11.09) & \textbf{7.52$\times$} (10.33) & \textbf{6.07$\times$} (8.39) \\
    \midrule
    \multicolumn{7}{l}{\emph{$T{=}0$, full data: 1.42M target-regenerated prompts}} \\
    DARTree & released & -- & -- & -- & -- & -- \\
    \dpace{} + \ddtree{} & FT 3 ep & 7.36$\times$ (9.90) & 7.76$\times$ (10.46) & 7.97$\times$ (10.70) & 7.37$\times$ (9.88) & 6.06$\times$ (8.19) \\
    \mfd{} ($\K{=}8$) & FT 3 ep & \textbf{7.88$\times$} (10.85) & \textbf{8.40$\times$} (11.60) & \textbf{8.31$\times$} (11.44) & \textbf{7.94$\times$} (10.90) & \textbf{6.36$\times$} (8.80) \\
    \midrule
    \multicolumn{7}{l}{\emph{$T{=}1$}} \\
    EAGLE-3 & released & -- & -- & -- & -- & -- \\
    \dflash{} chain & released & 3.09$\times$ (3.96) & 3.82$\times$ (4.91) & 3.89$\times$ (4.99) & 3.29$\times$ (4.23) & 2.36$\times$ (3.03) \\
    \dflash{}~2 selector chain & FT 1 ep & 3.76$\times$ (4.90) & 4.54$\times$ (5.93) & 4.60$\times$ (6.01) & 4.03$\times$ (5.28) & 2.90$\times$ (3.80) \\
    \ddtree{} & released & 4.58$\times$ (6.23) & 5.42$\times$ (7.37) & 5.50$\times$ (7.49) & 4.82$\times$ (6.57) & 3.56$\times$ (4.87) \\
    \dflash{}~2 + \ddtree{} & FT 1 ep & 4.40$\times$ (6.09) & 5.19$\times$ (7.18) & 5.31$\times$ (7.37) & 4.74$\times$ (6.56) & 3.46$\times$ (4.81) \\
    \dpace{} + \ddtree{} & FT 1 ep & 4.40$\times$ (5.99) & 5.30$\times$ (7.22) & 5.44$\times$ (7.39) & 4.70$\times$ (6.41) & 3.54$\times$ (4.83) \\
    \dpace{} + \ddtree{} & FT 3 ep & 4.48$\times$ (6.10) & 5.26$\times$ (7.16) & 5.33$\times$ (7.24) & 4.73$\times$ (6.45) & 3.51$\times$ (4.79) \\
    \cmidrule{1-7}
    \mfd{} ($\K{=}8$) & FT 1 ep & 4.84$\times$ (6.76) & \textbf{5.77$\times$} (8.06) & \textbf{5.81$\times$} (8.10) & 5.01$\times$ (7.01) & \textbf{3.70$\times$} (5.17) \\
    \mfd{} ($\K{=}8$) & FT 3 ep & \textbf{4.88$\times$} (6.81) & 5.72$\times$ (7.98) & 5.69$\times$ (7.92) & \textbf{5.19$\times$} (7.25) & \textbf{3.70$\times$} (5.18) \\
    \midrule
    \multicolumn{7}{l}{\emph{$T{=}1$, full data: 1.42M target-regenerated prompts}} \\
    DARTree & released & -- & -- & -- & -- & -- \\
    \dpace{} + \ddtree{} & FT 3 ep & 4.61$\times$ (6.29) & 5.32$\times$ (7.24) & 5.44$\times$ (7.40) & 4.72$\times$ (6.43) & 3.60$\times$ (4.93) \\
    \mfd{} ($\K{=}8$) & FT 3 ep & \textbf{5.10$\times$} (7.14) & \textbf{5.91$\times$} (8.25) & \textbf{5.84$\times$} (8.14) & \textbf{5.19$\times$} (7.25) & \textbf{3.85$\times$} (5.38) \\
    \bottomrule
  \end{tabular}
\end{table}

\begin{table}[!htbp]
  \centering
  \caption{Per-task speedup over autoregressive decoding and $\tau$ (in parentheses) on Qwen3.6-27B (thinking): the cells of \cref{tab:main} by task, from the same runs (every tree at $\B{=}64$). Bold: fastest in each column within each block.}
  \label{tab:pertask-q36-27b}
  \small
  \setlength{\tabcolsep}{3pt}
  \begin{tabular}{l l ccccc}
    \toprule
    Method & Training & GSM8K & MATH-500 & HumanEval & MBPP & MT-Bench \\
    \midrule
    \multicolumn{7}{l}{\emph{$T{=}0$}} \\
    EAGLE-3 & released & -- & -- & -- & -- & -- \\
    \dflash{} chain & released & 4.86$\times$ (6.07) & 5.63$\times$ (6.99) & 4.94$\times$ (6.18) & 4.51$\times$ (5.61) & 3.37$\times$ (4.21) \\
    \dflash{}~2 selector chain & FT 1 ep & 5.11$\times$ (6.38) & 6.12$\times$ (7.62) & 5.07$\times$ (6.32) & 4.58$\times$ (5.70) & 3.55$\times$ (4.44) \\
    \ddtree{} & released & 6.13$\times$ (7.88) & 6.82$\times$ (8.83) & 6.20$\times$ (8.06) & 5.79$\times$ (7.44) & 4.52$\times$ (5.90) \\
    \dflash{}~2 + \ddtree{} & FT 1 ep & 6.03$\times$ (7.78) & 6.93$\times$ (8.94) & 5.86$\times$ (7.55) & 5.42$\times$ (6.94) & 4.44$\times$ (5.76) \\
    \dpace{} + \ddtree{} & FT 1 ep & 6.69$\times$ (8.60) & 7.45$\times$ (9.76) & 6.79$\times$ (8.76) & 6.36$\times$ (8.13) & 4.85$\times$ (6.31) \\
    \dpace{} + \ddtree{} & FT 3 ep & 6.78$\times$ (8.70) & 7.53$\times$ (9.74) & 6.81$\times$ (8.85) & 6.34$\times$ (8.14) & 4.82$\times$ (6.30) \\
    \cmidrule{1-7}
    \mfd{} ($\K{=}8$) & FT 1 ep & 6.83$\times$ (8.94) & 7.54$\times$ (10.07) & 6.99$\times$ (9.18) & 6.46$\times$ (8.41) & 4.98$\times$ (6.59) \\
    \mfd{} ($\K{=}8$) & FT 3 ep & \textbf{7.04$\times$} (9.20) & \textbf{7.93$\times$} (10.43) & \textbf{7.19$\times$} (9.52) & \textbf{6.67$\times$} (8.72) & \textbf{5.04$\times$} (6.69) \\
    \midrule
    \multicolumn{7}{l}{\emph{$T{=}0$, full data: 1.42M target-regenerated prompts}} \\
    DARTree & released & -- & -- & -- & -- & -- \\
    \dpace{} + \ddtree{} & FT 3 ep & 7.32$\times$ (9.50) & 7.83$\times$ (10.19) & 6.91$\times$ (8.95) & 6.63$\times$ (8.43) & 5.03$\times$ (6.52) \\
    \mfd{} ($\K{=}8$) & FT 3 ep & \textbf{7.78$\times$} (10.28) & \textbf{8.29$\times$} (11.00) & \textbf{7.38$\times$} (9.72) & \textbf{7.07$\times$} (9.15) & \textbf{5.26$\times$} (6.93) \\
    \midrule
    \multicolumn{7}{l}{\emph{$T{=}1$}} \\
    EAGLE-3 & released & -- & -- & -- & -- & -- \\
    \dflash{} chain & released & 4.13$\times$ (5.15) & 4.81$\times$ (5.97) & 4.29$\times$ (5.36) & 4.05$\times$ (5.04) & 2.95$\times$ (3.69) \\
    \dflash{}~2 selector chain & FT 1 ep & 4.52$\times$ (5.65) & 5.33$\times$ (6.66) & 4.56$\times$ (5.72) & 4.16$\times$ (5.18) & 3.16$\times$ (3.96) \\
    \ddtree{} & released & 5.47$\times$ (7.10) & 6.06$\times$ (7.89) & 5.55$\times$ (7.26) & 5.29$\times$ (6.84) & 3.96$\times$ (5.20) \\
    \dflash{}~2 + \ddtree{} & FT 1 ep & 5.36$\times$ (6.89) & 5.96$\times$ (7.78) & 5.24$\times$ (6.82) & 4.90$\times$ (6.29) & 3.89$\times$ (5.08) \\
    \dpace{} + \ddtree{} & FT 1 ep & 5.77$\times$ (7.48) & 6.32$\times$ (8.27) & 5.81$\times$ (7.58) & 5.56$\times$ (7.17) & 4.13$\times$ (5.44) \\
    \dpace{} + \ddtree{} & FT 3 ep & 5.68$\times$ (7.38) & 6.26$\times$ (8.15) & 5.84$\times$ (7.65) & 5.59$\times$ (7.23) & 4.14$\times$ (5.44) \\
    \cmidrule{1-7}
    \mfd{} ($\K{=}8$) & FT 1 ep & 6.06$\times$ (8.00) & 6.67$\times$ (8.88) & 6.20$\times$ (8.23) & 5.89$\times$ (7.73) & 4.28$\times$ (5.71) \\
    \mfd{} ($\K{=}8$) & FT 3 ep & \textbf{6.33$\times$} (8.35) & \textbf{6.79$\times$} (9.00) & \textbf{6.28$\times$} (8.36) & \textbf{5.92$\times$} (7.79) & \textbf{4.37$\times$} (5.83) \\
    \midrule
    \multicolumn{7}{l}{\emph{$T{=}1$, full data: 1.42M target-regenerated prompts}} \\
    DARTree & released & -- & -- & -- & -- & -- \\
    \dpace{} + \ddtree{} & FT 3 ep & 6.13$\times$ (7.95) & 6.47$\times$ (8.46) & 5.83$\times$ (7.64) & 5.75$\times$ (7.41) & 4.24$\times$ (5.55) \\
    \mfd{} ($\K{=}8$) & FT 3 ep & \textbf{6.73$\times$} (8.88) & \textbf{7.02$\times$} (9.33) & \textbf{6.45$\times$} (8.58) & \textbf{6.17$\times$} (8.09) & \textbf{4.44$\times$} (5.91) \\
    \bottomrule
  \end{tabular}
\end{table}

\FloatBarrier

%% file: refs.bib
@inproceedings{leviathan2023fast,
  title={Fast Inference from Transformers via Speculative Decoding},
  author={Leviathan, Yaniv and Kalman, Matan and Matias, Yossi},
  booktitle={International Conference on Machine Learning},
  pages={19274--19286},
  year={2023},
  organization={PMLR}
}

@article{chen2023accelerating,
  title={Accelerating Large Language Model Decoding with Speculative Sampling},
  author={Chen, Charlie and Borgeaud, Sebastian and Irving, Geoffrey and Lespiau, Jean-Baptiste and Sifre, Laurent and Jumper, John},
  journal={arXiv preprint arXiv:2302.01318},
  year={2023}
}

@inproceedings{chen2026dflash,
  title={{DFlash}: Block Diffusion for Flash Speculative Decoding},
  author={Chen, Jian and Liang, Yesheng and Liu, Zhijian},
  booktitle={International Conference on Machine Learning},
  pages={15343--15355},
  year={2026},
  organization={PMLR}
}

@inproceedings{ringel2026ddtree,
  title={Accelerating Speculative Decoding with Block Diffusion Draft Trees},
  author={Ringel, Liran and Romano, Yaniv},
  booktitle={Conference on Language Modeling},
  year={2026}
}

@article{hu2026jetspec,
  title={{JetSpec}: Breaking the Scaling Ceiling of Speculative Decoding with Parallel Tree Drafting},
  author={Hu, Lanxiang and Feng, Zhaoxiang and Wu, Yulun and Yuan, Haoran and Zhao, Yujie and Qian, Yu-Yang and Wang, Bojun and Zhao, Peng and Jiang, Daxin and Zhu, Yibo and others},
  journal={arXiv preprint arXiv:2606.18394},
  year={2026}
}

@article{oda2026weaver,
  title={Trees from Marginals: Autoregressive Drafting with Factorized Priors},
  author={Oda, Yuma and Mathieu, Ryan and Knyazhitskiy, Roman and Chakhvadze, Artur},
  journal={arXiv preprint arXiv:2607.06763},
  year={2026}
}

@article{li2026dartree,
  title={{DARTree}: Speculative Diffusion Decoding with Autoregressive Draft Trees},
  author={Li, Tianyi and Luo, Yaxin and Shang, Xinyi and Shen, Zhiqiang},
  journal={arXiv preprint arXiv:2608.13524},
  year={2026}
}

@article{liang2026ubtree,
  title={{UBTree}: Parallel Tree Drafting via Unigram and Bigram Models for Speculative Decoding},
  author={Liang, Chumeng and Wang, Linxuan and Peng, Xinyu and Liu, Huabin and Guo, Xiaoxu and Chen, Yuxin and Liu, Ge and Lin, Guang and Song, Qifan and Li, Jianguo},
  journal={arXiv preprint arXiv:2609.39972},
  year={2026}
}

@inproceedings{dpace2026,
  title={{D-PACE}: Dynamic Position-Aware Cross-Entropy for Parallel Speculative Drafting},
  author={Wu, Tianyu and Yao, Yu and Qi, Zhenting and Zheng, Han and Zhang, Chengxi and Wang, Zhuohan and Ma, Haoran and Liao, Zichun and Lakkaraju, Himabindu and Li, Ju and others},
  booktitle={Advances in Neural Information Processing Systems},
  year={2026}
}

@inproceedings{miao2024specinfer,
  title={{SpecInfer}: Accelerating Large Language Model Serving with Tree-based Speculative Inference and Verification},
  author={Miao, Xupeng and Oliaro, Gabriele and Zhang, Zhihao and Cheng, Xinhao and Wang, Zeyu and Zhang, Zhengxin and Wong, Rae Ying Yee and Zhu, Alan and Yang, Lijie and Shi, Xiaoxiang and others},
  booktitle={Proceedings of the 29th ACM International Conference on Architectural Support for Programming Languages and Operating Systems, Volume 3},
  pages={932--949},
  year={2024}
}

@inproceedings{cai2024medusa,
  title={{Medusa}: Simple {LLM} Inference Acceleration Framework with Multiple Decoding Heads},
  author={Cai, Tianle and Li, Yuhong and Geng, Zhengyang and Peng, Hongwu and Lee, Jason D. and Chen, Deming and Dao, Tri},
  booktitle={International Conference on Machine Learning},
  pages={5209--5235},
  year={2024},
  organization={PMLR}
}

@inproceedings{li2024eagle,
  title={{EAGLE}: Speculative Sampling Requires Rethinking Feature Uncertainty},
  author={Li, Yuhui and Wei, Fangyun and Zhang, Chao and Zhang, Hongyang},
  booktitle={International Conference on Machine Learning},
  pages={28935--28948},
  year={2024},
  organization={PMLR}
}

@inproceedings{li2024eagle2,
  title={{EAGLE-2}: Faster Inference of Language Models with Dynamic Draft Trees},
  author={Li, Yuhui and Wei, Fangyun and Zhang, Chao and Zhang, Hongyang},
  booktitle={Proceedings of the 2024 Conference on Empirical Methods in Natural Language Processing},
  pages={7421--7432},
  year={2024}
}

@article{li2025eagle3,
  title={{EAGLE-3}: Scaling up Inference Acceleration of Large Language Models via Training-Time Test},
  author={Li, Yuhui and Wei, Fangyun and Zhang, Chao and Zhang, Hongyang},
  journal={Advances in Neural Information Processing Systems},
  volume={38},
  pages={151568--151587},
  year={2025}
}

@article{chen2024sequoia,
  title={{Sequoia}: Scalable and Robust Speculative Decoding},
  author={Chen, Zhuoming and May, Avner and Svirschevski, Ruslan and Huang, Yuhsun and Ryabinin, Max and Jia, Zhihao and Chen, Beidi},
  journal={Advances in Neural Information Processing Systems},
  volume={37},
  pages={129531--129563},
  year={2024}
}

@inproceedings{domino2026,
  title={{Domino}: Decoupling Causal Modeling from Autoregressive Drafting in Speculative Decoding},
  author={Huang, Jianuo and Zhang, Yaojie and Zhang, Qituan and Lin, Hao and Xu, Hanlin and Zhang, Linfeng},
  booktitle={Proceedings of the 2026 Conference on Empirical Methods in Natural Language Processing},
  pages={23346--23356},
  year={2026}
}

@article{dspark2026,
  title={{DSpark}: Confidence-Scheduled Speculative Decoding with Semi-Autoregressive Generation},
  author={Cheng, Xin and Yu, Xingkai and Shao, Chenze and Li, Jiashi and Xiong, Yunfan and Qian, Yi and Zhu, Jiaqi and Ma, Shirong and Zhang, Xiaokang and Ye, Jiasheng and others},
  journal={arXiv preprint arXiv:2607.05147},
  year={2026}
}

@article{zhao2026treespark,
  title={{TreeSpark}: Learning to Draft the Trees You Verify for Speculative Decoding},
  author={Zhao, Jihao and Jiang, Qianyu and Zu, Shuaishuai and Li, Pengfei and Fang, Guanghuan and Feng, Ben and Qin, Biao and Wang, Zhibin and Yang, Baochuan},
  journal={arXiv preprint arXiv:2610.20175},
  year={2026}
}

@article{basharin2024faster,
  title={Faster Language Models with Better Multi-Token Prediction Using Tensor Decomposition},
  author={Basharin, Artem and Chertkov, Andrei and Oseledets, Ivan},
  journal={arXiv preprint arXiv:2410.17765},
  year={2024}
}

@inproceedings{mtpc2025,
  title={Fast and Expressive Multi-Byte Prediction with Probabilistic Circuits},
  author={Grivas, Andreas and Loconte, Lorenzo and van Krieken, Emile and Nawrot, Piotr and Zhao, Yu and Wielewski, Euan and Minervini, Pasquale and Ponti, Edoardo and Vergari, Antonio},
  booktitle={International Conference on Machine Learning},
  pages={36671--36704},
  year={2026},
  organization={PMLR}
}

@article{dblast2026,
  title={{DBLAST}: Dependent Block Drafting for Stochastic Speculative Decoding},
  author={Karimi, Amirmohammad and Gao, Chao and Hassanpour, Negar},
  journal={arXiv preprint arXiv:2608.05448},
  year={2026}
}

@inproceedings{ptp2025,
  title={Parallel Token Prediction for Language Models},
  author={Draxler, Felix and Will, Justus and Sofian, Farrin Marouf and Karaletsos, Theofanis and Singh, Sameer and Mandt, Stephan},
  booktitle={International Conference on Learning Representations},
  year={2026}
}

@article{flashmap2026,
  title={{FlashMap}: Flow Map Speculative Decoding},
  author={Shen, Jeff and Zhou, Zhejian and Lee, Ryan and May, Jonathan and Ho, Shirley},
  journal={arXiv preprint arXiv:2610.24365},
  year={2026}
}

@article{bole2026,
  title={{Twig}: Parallel and Memory-Efficient Tree Speculative Decoding for Hybrid-Attention {LLMs}},
  author={Wang, Li and Su, Yi and Wu, Xiabao and You, Chiran and Liu, Yongchao and Qiu, Zhan and Zhang, Juelu and Zheng, Jiajun and Liu, Fangxin and Zhang, Jie and others},
  journal={arXiv preprint arXiv:2608.01651},
  year={2026}
}

@article{treewy2026,
  title={{TreeWY}: Speculative Verification for {Gated DeltaNet} Hybrids},
  author={Ghantasala, Sneha Murthy},
  journal={arXiv preprint arXiv:2608.20961},
  year={2026}
}

@misc{inco2026dflash2,
  title={{DFlash 2}: Keep Drafting Parallel},
  author={{Inco AI}},
  howpublished={Blog post, \url{https://inco.ai/blog/dflash2/}},
  year={2026}
}

@inproceedings{gu2018nat,
  title={Non-Autoregressive Neural Machine Translation},
  author={Gu, Jiatao and Bradbury, James and Xiong, Caiming and Li, Victor O.K. and Socher, Richard},
  booktitle={International Conference on Learning Representations},
  year={2018}
}

@inproceedings{zhou2020understanding,
  title={Understanding Knowledge Distillation in Non-Autoregressive Machine Translation},
  author={Zhou, Chunting and Gu, Jiatao and Neubig, Graham},
  booktitle={International Conference on Learning Representations},
  year={2020}
}

@inproceedings{shen2019mixture,
  title={Mixture Models for Diverse Machine Translation: Tricks of the Trade},
  author={Shen, Tianxiao and Ott, Myle and Auli, Michael and Ranzato, Marc'Aurelio},
  booktitle={International Conference on Machine Learning},
  pages={5719--5728},
  year={2019},
  organization={PMLR}
}

@article{lee2016smcl,
  title={Stochastic Multiple Choice Learning for Training Diverse Deep Ensembles},
  author={Lee, Stefan and Purushwalkam, Senthil and Cogswell, Michael and Ranjan, Viresh and Crandall, David and Batra, Dhruv},
  journal={Advances in Neural Information Processing Systems},
  volume={29},
  year={2016}
}

@inproceedings{rupprecht2017uncertain,
  title={Learning in an Uncertain World: Representing Ambiguity Through Multiple Hypotheses},
  author={Rupprecht, Christian and Laina, Iro and DiPietro, Robert and Baust, Maximilian and Tombari, Federico and Navab, Nassir and Hager, Gregory D.},
  booktitle={Proceedings of the IEEE International Conference on Computer Vision},
  pages={3611--3620},
  year={2017}
}

@article{versteeg2014corex,
  title={Discovering Structure in High-Dimensional Data Through Correlation Explanation},
  author={Ver Steeg, Greg and Galstyan, Aram},
  journal={Advances in Neural Information Processing Systems},
  volume={27},
  year={2014}
}

@inproceedings{liu2024copula,
  title={Discrete Copula Diffusion},
  author={Liu, Anji and Broadrick, Oliver and Niepert, Mathias and Van den Broeck, Guy},
  booktitle={International Conference on Learning Representations},
  year={2025}
}

@inproceedings{wu2025fastdllm,
  title={{Fast-dLLM}: Training-free Acceleration of Diffusion {LLM} by Enabling {KV} Cache and Parallel Decoding},
  author={Wu, Chengyue and Zhang, Hao and Xue, Shuchen and Liu, Zhijian and Diao, Shizhe and Zhu, Ligeng and Luo, Ping and Han, Song and Xie, Enze},
  booktitle={International Conference on Learning Representations},
  year={2026}
}

@article{benhamu2025entropy,
  title={Accelerated Sampling from Masked Diffusion Models via Entropy Bounded Unmasking},
  author={Ben-Hamu, Heli and Gat, Itai and Severo, Daniel and Nolte, Niklas and Karrer, Brian},
  journal={Advances in Neural Information Processing Systems},
  volume={38},
  pages={62365--62391},
  year={2025}
}

@article{stern2018blockwise,
  title={Blockwise Parallel Decoding for Deep Autoregressive Models},
  author={Stern, Mitchell and Shazeer, Noam and Uszkoreit, Jakob},
  journal={Advances in Neural Information Processing Systems},
  volume={31},
  year={2018}
}

@inproceedings{ankner2024hydra,
  title={{Hydra}: Sequentially-Dependent Draft Heads for {Medusa} Decoding},
  author={Ankner, Zachary and Parthasarathy, Rishab and Nrusimha, Aniruddha and Rinard, Christopher and Ragan-Kelley, Jonathan and Brandon, William},
  booktitle={Conference on Language Modeling},
  year={2024}
}

@inproceedings{gloeckle2024mtp,
  title={Better \& Faster Large Language Models via Multi-Token Prediction},
  author={Gloeckle, Fabian and Idrissi, Badr Youbi and Rozi{\`e}re, Baptiste and Lopez-Paz, David and Synnaeve, Gabriel},
  booktitle={International Conference on Machine Learning},
  pages={15706--15734},
  year={2024},
  organization={PMLR}
}

@inproceedings{huang2022dat,
  title={Directed Acyclic Transformer for Non-Autoregressive Machine Translation},
  author={Huang, Fei and Zhou, Hao and Liu, Yang and Li, Hang and Huang, Minlie},
  booktitle={International Conference on Machine Learning},
  pages={9410--9428},
  year={2022},
  organization={PMLR}
}

@article{sun2023spectr,
  title={{SpecTr}: Fast Speculative Decoding via Optimal Transport},
  author={Sun, Ziteng and Suresh, Ananda Theertha and Ro, Jae Hun and Beirami, Ahmad and Jain, Himanshu and Yu, Felix},
  journal={Advances in Neural Information Processing Systems},
  volume={36},
  pages={30222--30242},
  year={2023}
}

@article{guzmanrivera2012mcl,
  title={Multiple Choice Learning: Learning to Produce Multiple Structured Outputs},
  author={Guzman-Rivera, Abner and Batra, Dhruv and Kohli, Pushmeet},
  journal={Advances in Neural Information Processing Systems},
  volume={25},
  year={2012}
}

@article{watanabe1960information,
  title={Information Theoretical Analysis of Multivariate Correlation},
  author={Watanabe, Satosi},
  journal={IBM Journal of Research and Development},
  volume={4},
  number={1},
  pages={66--82},
  year={1960}
}

@article{zheng2024sglang,
  title={{SGLang}: Efficient Execution of Structured Language Model Programs},
  author={Zheng, Lianmin and Yin, Liangsheng and Xie, Zhiqiang and Sun, Chuyue and Huang, Jeff and Yu, Cody Hao and Cao, Shiyi and Kozyrakis, Christos and Stoica, Ion and Gonzalez, Joseph E and others},
  journal={Advances in Neural Information Processing Systems},
  volume={37},
  pages={62557--62583},
  year={2024}
}

@inproceedings{yang2025gated,
  title={Gated Delta Networks: Improving {Mamba2} with Delta Rule},
  author={Yang, Songlin and Kautz, Jan and Hatamizadeh, Ali},
  booktitle={International Conference on Learning Representations},
  year={2025}
}

@inproceedings{he2018sequence,
  title={Sequence to Sequence Mixture Model for Diverse Machine Translation},
  author={He, Xuanli and Haffari, Gholamreza and Norouzi, Mohammad},
  booktitle={Proceedings of the 22nd Conference on Computational Natural Language Learning},
  pages={583--592},
  year={2018}
}

@inproceedings{khisti2025multidraft,
  title={Multi-Draft Speculative Sampling: Canonical Decomposition and Theoretical Limits},
  author={Khisti, Ashish and Ebrahimi, M. Reza and Dbouk, Hassan and Behboodi, Arash and Memisevic, Roland and Louizos, Christos},
  booktitle={International Conference on Learning Representations},
  year={2025}
}

@article{wang2025opttree,
  title={{OPT-Tree}: Speculative Decoding with Adaptive Draft Tree Structure},
  author={Wang, Jikai and Su, Yi and Li, Juntao and Xia, Qingrong and Ye, Zi and Duan, Xinyu and Wang, Zhefeng and Zhang, Min},
  journal={Transactions of the Association for Computational Linguistics},
  volume={13},
  pages={188--199},
  year={2025}
}

@inproceedings{kaiser2018fast,
  title={Fast Decoding in Sequence Models Using Discrete Latent Variables},
  author={Kaiser, Lukasz and Bengio, Samy and Roy, Aurko and Vaswani, Ashish and Parmar, Niki and Uszkoreit, Jakob and Shazeer, Noam},
  booktitle={International Conference on Machine Learning},
  pages={2390--2399},
  year={2018}
}

@inproceedings{ma2019flowseq,
  title={{FlowSeq}: Non-Autoregressive Conditional Sequence Generation with Generative Flow},
  author={Ma, Xuezhe and Zhou, Chunting and Li, Xian and Neubig, Graham and Hovy, Eduard},
  booktitle={Proceedings of the 2019 Conference on Empirical Methods in Natural Language Processing and the 9th International Joint Conference on Natural Language Processing},
  pages={4282--4292},
  year={2019}
}

@inproceedings{shu2020latent,
  title={Latent-Variable Non-Autoregressive Neural Machine Translation with Deterministic Inference Using a Delta Posterior},
  author={Shu, Raphael and Lee, Jason and Nakayama, Hideki and Cho, Kyunghyun},
  booktitle={Proceedings of the AAAI Conference on Artificial Intelligence},
  volume={34},
  number={05},
  pages={8846--8853},
  year={2020}
}

@inproceedings{wu2025stree,
  title={{STree}: Speculative Tree Decoding for Hybrid State Space Models},
  author={Wu, Yangchao and Qin, Zongyue and Wong, Alex and Soatto, Stefano},
  booktitle={Advances in Neural Information Processing Systems},
  year={2025}
}

@inproceedings{rheinboldt2026treeflash,
  title={{TreeFlash}: Parallel {AR}-Approximation for Faster Speculative Decoding},
  author={Rheinboldt, Peer and Berdoz, Fr{\'e}d{\'e}ric and Wattenhofer, Roger},
  booktitle={Findings of the Association for Computational Linguistics: EMNLP 2026},
  pages={45297--45310},
  year={2026}
}
